%% file: submission-template/main.tex
\documentclass[sigconf, nonacm, balance=false]{acmart}
\usepackage{submission-template/popets}
\usepackage[many]{tcolorbox}
\usepackage{pifont}
\usepackage{enumitem} 
\begin{document}

\title{Explainable AI for Analyzing Person-Specific Patterns in Facial Recognition Tasks
}


\author{Paweł Jakub Borsukiewicz}
\email{pawel.borsukiewicz@uni.lu}
\orcid{0000-0002-2934-6115}
\affiliation{
  \institution{University of Luxembourg}
  \city{Luxembourg}
  \country{Luxembourg}
}

\author{Jordan Samhi}
\email{jordan.samhi@uni.lu}
\orcid{0000-0001-6052-6184}
\affiliation{
  \institution{University of Luxembourg}
  \city{Luxembourg}
  \country{Luxembourg}
}

\author{Jacques Klein}
\email{jacques.klein@uni.lu}
\orcid{0000-0003-4052-475X}
\affiliation{
  \institution{University of Luxembourg}
  \city{Luxembourg}
  \country{Luxembourg}
}

\author{Tegawendé F. Bissyandé }
\email{tegawende.bissyande@uni.lu}
\orcid{0000-0001-7270-9869}
\affiliation{
  \institution{University of Luxembourg}
  \city{Luxembourg}
  \country{Luxembourg}
}


\renewcommand{\shortauthors}{Borsukiewicz et al.}


\input{./Sections/abstract.tex}

\keywords{facial recognition prevention, explainable AI, privacy, CAM}

\maketitle

\newcommand{\find}[1]{
\begin{tcolorbox}[leftrule=0.4mm,rightrule=0mm,toprule=0mm,bottomrule=0mm,left=0.0pt,right=0.0pt,top=0pt,bottom=0pt]
\em #1
\end{tcolorbox}
}

\input{./Sections/introduction.tex}

\input{./Sections/related.tex}

\input{./Sections/methodology.tex}
\input{./Sections/experiments.tex}
\input{./Sections/discussions.tex}
\input{./Sections/conclusion.tex}

\section*{Acknowledgments}
\begin{itemize}[leftmargin=*]
    \item The authors acknowledge the use of AI-based tools in this work. Their usage was limited to correcting typos and grammatical errors as well as to rephrasing and refining the final text for readability and clarity.
    \item The experiments presented in this paper were partially carried
out using the HPC\footnote{https://hpc.uni.lu} facilities of the University of Luxembourg.
    \item This research was supported by the Luxembourg Army.
\end{itemize}


\input{submission-template/main.bbl}



\input{./Sections/appendix.tex}
\end{document}

%% file: Sections/abstract.tex
\begin{abstract}
The proliferation of facial recognition systems presents major privacy risks, driving the need for effective countermeasures. Current adversarial techniques apply generalized methods rather than adapting to individual facial characteristics, limiting their effectiveness and inconspicuousness. In this work, we introduce Layer Embedding Activation Mapping (LEAM), a novel technique that identifies which facial areas contribute most to recognition at an individual level.
Unlike adversarial attack methods that aim to fool recognition systems, LEAM is an explainability technique designed to understand how these systems work, providing insights that could inform future privacy protection research. 
We integrate LEAM with a face parser to analyze data from 1000 individuals across 9 pre-trained facial recognition models.

Our analysis reveals that while different layers within facial recognition models vary significantly in their focus areas, these models generally prioritize similar facial regions across architectures when considering their overall activation patterns
, which show significantly higher similarity between images of the same individual (Bhattacharyya Coefficient: 0.32-0.57) vs. different individuals (0.04-0.13), validating the existence of person-specific recognition patterns. Our results show that facial recognition models prioritize the central region of face images (with nose areas accounting for 18.9-29.7\% of critical recognition regions), while still distributing attention across multiple facial fragments.
Proper selection of relevant facial areas was confirmed using validation occlusions, based on just 1\% of the most relevant, LEAM-identified, image pixels, which proved to be transferable across different models (decrease in cosine similarity scores of 0.1148 vs. 0.0286 for random positions).
Gender analysis shows that occlusions are more effective for female images (0.1248 cosine similarity drop vs. 0.1038 for males), suggesting that protection strategies may need to be optimized based on gender.
Our findings establish the foundation for future individually tailored privacy protection systems centered around LEAM's choice of areas to be perturbed. 
\end{abstract}

%% file: Sections/introduction.tex
\section{Introduction}
\begin{figure}[!t] 
    \centering
    \includegraphics[width=.4\textwidth]{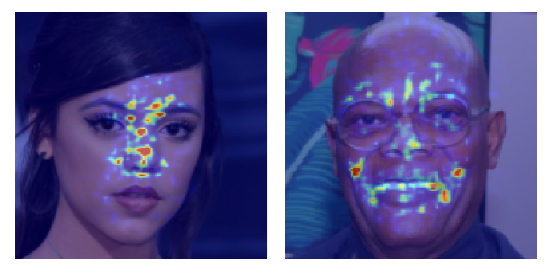}
    \caption{{Depending on the individuals, 
    neural networks look at different regions for facial recognition}}
    \label{fig:Intro_LEAM}
\end{figure}

Facial recognition systems have become ubiquitous, deployed across various security applications from border control to law enforcement and commercial surveillance. While these systems serve legitimate security functions, they simultaneously pose significant privacy risks as the extent of data collection and processing is often opaque to the public. The dual-use nature of this technology has raised considerable ethical and legal concerns, particularly when deployed for mass surveillance or in military contexts. These concerns have led to a growing demand for effective facial recognition prevention techniques that can protect individual privacy.

The problematic balance between security applications and privacy violations has recently been acknowledged within the academic community. Multiple prominent datasets specifically designed for facial recognition research have been retracted due to ethical considerations around consent and potential misuse~\cite{BoutrosEtAl2023}. This underscores the urgent need for research that addresses both the technical aspects of facial recognition and its privacy implications.

Research on human facial recognition processes has established that central facial features—face outline, nose, eyes, and mouth—are frequently the most distinctive areas for human perception~\cite{ZhaoEtAl2003}. However, in the context of machine learning-based recognition, there is a significant gap in our understanding of which areas are most important. While recent studies have explored attention mechanisms~\cite{soydaner2022attention,GuettaEtAl2021} to guide neural networks toward relevant facial regions, they primarily aim to improve or decrease recognition accuracy rather than explain the key areas for recognition.

In the domain of adversarial attacks, subtle occlusions in critical facial areas can lead to successful recognition evasion. However, current attack methods generally apply a one-size-fits-all approach rather than targeting the specific facial features most important for recognizing a particular individual. 
While adversarial attack research focuses on fooling recognition systems, our work takes a fundamentally different approach: we aim to understand and explain how these systems identify individuals through visualization techniques.

{\bf This paper.} Our work addresses this research gap by introducing Layer Embedding Activation Mapping (LEAM), a novel approach to class activation mapping visualization for facial recognition models. LEAM leverages activation maps to compute the impact of specific image pixels on facial recognition predictions, highlighting the most influential areas. This allows us to analyze which facial regions predominate at various layers of neural networks and determine their importance in the recognition process.

Our main contributions include: (1) introducing LEAM targeting embedding-based facial recognition models; (2) demonstrating that while individual layers focus differently, models exhibit similar overall attention patterns; (3) showing that recognition attention is distributed across multiple facial fragments rather than concentrated on specific features; (4) providing evidence that activation patterns are significantly more similar between images of the same person than different individuals, indicating the existence of individual-specific recognition patterns illustrated in Figure \ref{fig:Intro_LEAM}; (5) empirically demonstrating relevance of regions highlighted by showing that LEAM-guided validation occlusions significantly outperforming random approaches; (6) and demonstrating that the most relevant areas are transferable across model architectures.

The empirical results of our study have significant security and privacy implications, particularly for the design of privacy-enhancing technologies in the face of increasingly sophisticated facial recognition systems. Our work proves the potential and necessity for further research on developing individually tailored recognition prevention solutions that can more effectively protect privacy while remaining inconspicuous. Our research artifacts are made publicly available for open science: 
\begin{center}
\url{https://anonymous.4open.science/r/LEAM-C01B/README.md}
\end{center}

The remainder of this paper is structured as follows. Section II explores related works with emphasis on state of the art CAM-related techniques. Section III explains the methodology and design choices. Section IV focuses on the results of the experiments and their interpretation. Section V is dedicated to results discussion. Finally, section VI concludes the findings and suggests future work.

%% file: Sections/related.tex
\section{Related work}
Our research builds on and extends work in three key areas: privacy protection against facial recognition, neural network explainability through activation mapping, and facial representation learning. Privacy protection research provides context for the future practical applications of our findings, explainability techniques form the methodological foundation for our approach, and facial representation learning offers the tools to interpret our results at a granular level. This section reviews significant developments in each area to position our contributions within the broader research landscape.

\subsection{Privacy protection}
Research on facial recognition prevention spans both digital and physical domains. Digital attacks achieve high success rates with minimal perturbations~\cite{VakhshitehEtAl2021}, but require pre-processing images before analysis. This limitation makes them impractical for real-world scenarios where images are captured without knowledge or consent~\cite{DongEtAl2019}.

Physical countermeasures employ various occlusion techniques including adversarial stickers~\cite{Komkov2019}, makeup~\cite{GuettaEtAl2021}, specially designed glasses~\cite{sharif2019general}, 3D printed masks~\cite{YangEtAl2023}, and light-based approaches~\cite{shen2019vla}. These methods are designed to transfer effectively from digital simulations to physical applications across diverse individuals~\cite{YangEtAl2022}. 

The challenge with physical privacy protection lies in balancing effectiveness with inconspicuousness—the modified area should be minimal to avoid raising suspicion while maximizing privacy protection. This motivates our research to identify which specific facial landmarks contribute most significantly to recognition across models, enabling targeted privacy-preserving interventions.

While adversarial methods focus on actively evading recognition, our work takes an orthogonal approach by seeking to understand and explain recognition patterns, without proposing new attack methods.

\subsection{Neural Network Explainability with Class Activation Mapping}
Neural network explainability research~\cite{RibeiroEtAl2016} has historically overlooked facial recognition models despite their widespread deployment and privacy implications. Class activation mapping (CAM) has emerged as a valuable technique for visualizing how image elements influence model predictions~\cite{ZhouEtAl2015}. The field has evolved rapidly with significant advancements: Grad-CAM~\cite{SelvarajuEtAl2016} introduced gradient weighting for improved localization; Grad-CAM++~\cite{ChattopadhyayEtAl2017} refined this approach with pixel-wise gradient calculation, enhancing performance particularly for images with multiple instances of predicted classes; and LayerCAM~\cite{JiangEtAL2021} extended these capabilities to any convolutional layer or combination of layers.

However, facial recognition presents unique challenges for activation mapping because these models produce embeddings rather than classification outputs. Recognition decisions rely on similarity metrics between embeddings, making standard CAM approaches suboptimal. Recent adaptations for embedding networks have followed two main approaches: Chen et al.~\cite{ChenEtAl2020} substituted triplet loss for class scores, particularly effective for models using pair-based training methods~\cite{SchroffEtAl15}, while Bachhawat~\cite{bachhawat2024} developed EmbeddingCAM using proxy classes for gradient calculation, demonstrating improved performance on non-facial datasets. Despite these advances, no technique has been specifically designed for facial recognition and its associated privacy concerns—a gap our work addresses.

\subsection{Facial Representation Learning}
The Facial Representation Learning (FaRL) framework developed by Zheng et al.~\cite{ZhengEtAl2022} represents a significant advancement in face analysis by correlating images with textual descriptions. This approach has improved performance across multiple tasks, including face parsing, alignment, and attribute recognition, compared to previous pre-trained models. Their FACER toolkit~\cite{facer} implements these capabilities, allowing ease of use.

The face parser identifies 19 distinct classes that include primary facial features (\textit{face}, \textit{nose}, \textit{inner mouth}, and 2 variants -- \textit{left} or \textit{right} -- per \textit{eyes}, \textit{eyebrows}, \textit{lips} and \textit{ears}), contextual elements (\textit{hair}, \textit{neck}, \textit{background}), and accessories (\textit{clothes}, \textit{eyeglasses}, \textit{hat}, \textit{earrings}, \textit{necklace}). This comprehensive labeling enables precise correlation between activation patterns and specific facial regions, making it particularly valuable for our investigation into which facial areas contribute most significantly to recognition.

%% file: Sections/methodology.tex
\section{Methodology}

\begin{figure*}[!ht] 
    \centering
    \includegraphics[width=1.0\textwidth]{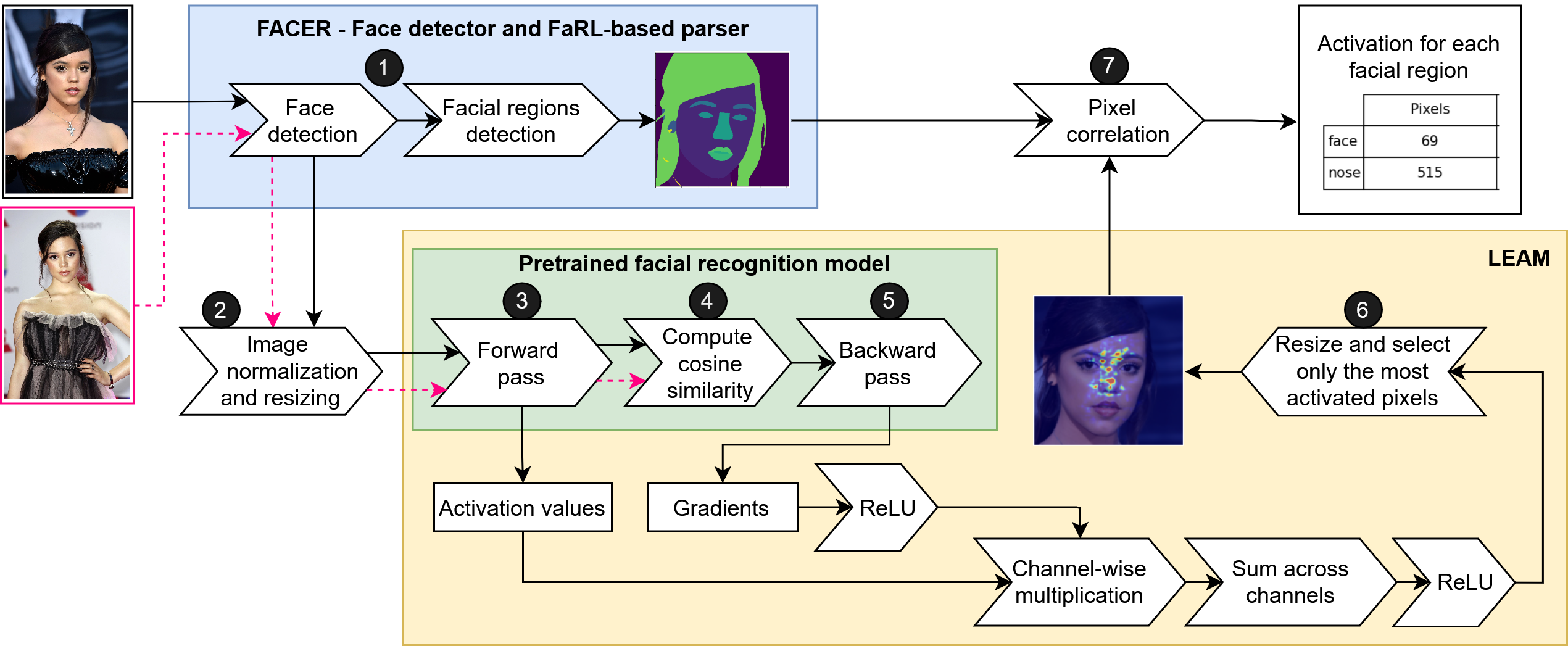}
    \caption{Workflow overview - interpreting facial recognition models through activation visualization}
    \label{fig:leam_exp}
\end{figure*}

This section introduces LEAM, our novel technique for interpreting facial recognition models through activation visualization. We first present the mathematical foundation of LEAM, then explain how we correlate these activation maps with facial regions using the FaRL framework, and finally describe our approaches for quantifying similarity between activation patterns. This methodology provides the foundation for uncovering individual-specific patterns in facial recognition that can inform more effective privacy protection.

It is important to note that LEAM is an explainable AI method for understanding facial recognition models, not an adversarial attack or a privacy protection system itself. Unlike methods such as Fawkes~\cite{shan2020fawkes} or Lowkey~\cite{cherepanova2021lowkey} that actively perturb images to evade recognition, LEAM visualizes and analyzes existing model behavior without attempting to fool or retrain these systems.

\subsection{Layer Embedding Activation
Mapping}
After analyzing various Class Activation Mapping (CAM)
approa\-ches and considering the specific requirements for facial recognition privacy protection, we developed LEAM—a novel extension of LayerCAM designed specifically for networks that produce embeddings rather than classification outputs. This approach uniquely addresses the gap between existing CAM techniques and facial recognition models, enabling visualization of the most discriminative facial features across multiple network layers. For consistency, we adopt mathematical notation from~\cite{JiangEtAL2021}.

LayerCAM relies on two primary components: weights and activation values. The weights of spatial locations \( w_{ij}^{kc} \)
(where (\textit{i},\textit{j}) are pixel coordinates of the \textit{k}-th feature map, and \textit{c} corresponds to the target class) are derived from pixel-specific gradients \(g_{ij}^{kc}\) obtained during backpropagation. Since facial recognition uses embeddings rather than class labels, we adapt this notation to \(w_{ij}^{ke}\) and \(g_{ij}^{ke}\) respectively:

\vspace{-1.0em}
\begin{equation}
\label{LEAM_eq1}
w_{ij}^{ke} = \text{ReLU} \left( g_{ij}^{ke} \right)
\end{equation}

Activation values \( A_{ij}^{k} \), obtained during the forward pass, are independent of both class and embedding. These values are multiplied with corresponding weights to create channel-specific embedding activation maps:

\vspace{-1.0em}
\begin{equation}
\label{LEAM_eq2}
\hat{A}_{ij}^{k} = w_{ij}^{ke} \cdot A_{ij}^{k}
\end{equation}

To generate a complete activation map for a specific layer, we aggregate results across all channels:

\vspace{-1.0em}
\begin{equation}
\label{LEAM_eq3}
M^{e} = \text{ReLU} \left( \sum_{k} \hat{A}^{k} \right)
\end{equation}

By using ReLU to extract only positive values, we focus on areas that positively contribute to recognition, i.e., have impact on the final classification.

The key innovation of LEAM lies in our adaptation of gradient calculation for facial recognition networks, which use similarity metrics rather than classification scores. Due to the subsequent choice of models producing L2-normalized embeddings, the majority of which are optimized for angular relationships on the hypersphere~\cite{deng2018arcface}, we select cosine similarity~\cite{XiaEtAl2015} as our foundational metric, defined for two embeddings \(x_1, x_2\) as:

\begin{equation} \label{cosine_eq}
CosSim(x_1,x_2) = \frac{x_1 \cdot x_2}{|x_1||x_2|}
\end{equation}

This metric produces values in the range [-1,1], where 1 indicates perfect similarity, 0 indicates no similarity, and negative values indicate dissimilarity. In facial recognition, embeddings from the same individual should yield values close to 1, while embeddings from different individuals should produce values below a predefined threshold.
While cosine similarity has faced criticism regarding gradient behavior during training~\cite{DraganovEtAl2024}—specifically that gradients for high-magnitude points tend toward zero—and potential meaningless similarities in certain contexts~\cite{SteckEtAl2024}, proper model training mitigates these concerns for our application.

For LEAM, we calculate the cosine similarity loss function for same-identity comparisons:

\begin{equation} \label{pos_cos_loss_eq}
\mathcal{L}(x_1, x_2) = 1 - \cos(x_1, x_2)
\end{equation}

It enables us to calculate pixel-specific gradients \(g_{ij}^{ke}\) during backpropagation, which form the foundation of our LEAM method.

\subsection{LEAM-FaRL correlation}

To associate LEAM activation patterns with specific facial regions, we leverage the FACER face parser implementation based on the FaRL framework. Figure \ref{fig:leam_exp} illustrates our processing pipeline.

Our process iterates through each identity in the dataset as follows:

\begin{enumerate}[leftmargin=*]
\item[\ding{182}] For each image, we extract the facial area using the face parser, storing information about 19 predefined facial regions. Images with multiple faces or no detected face are discarded.

\item[\ding{183}] We resize images according to the target model's specifications and normalize them to optimize embedding calculation.

\item[\ding{184}] We perform a forward pass through the model to obtain embeddings and activation values.

\item[\ding{185}] We generate all possible ordered pairs of images to calculate cosine similarities, ensuring each image serves as an anchor (black continuous line) and a positive sample (pink dashed line).

\item[\ding{186}] Using these similarities as our foundation, we calculate gradients during backpropagation and generate layer activation maps according to equations (\ref{LEAM_eq1}-\ref{LEAM_eq3}).

\item[\ding{187}] We correlate LEAM results with the face parser by resizing activation maps to a common size, allowing for pixel-perfect quantification of relevant pixels for each facial region class. Pixel relevance is visualized using a ``warmness'' threshold.

\item[\ding{188}] Finally, for each iteration, we save the yielded data for further analysis, including pixel counts per facial class, cosine similarity scores, image identifiers, model information, and analyzed layer.
\end{enumerate}

\begin{figure}[h] 
    \centering
    \includegraphics[width=1.0\linewidth]{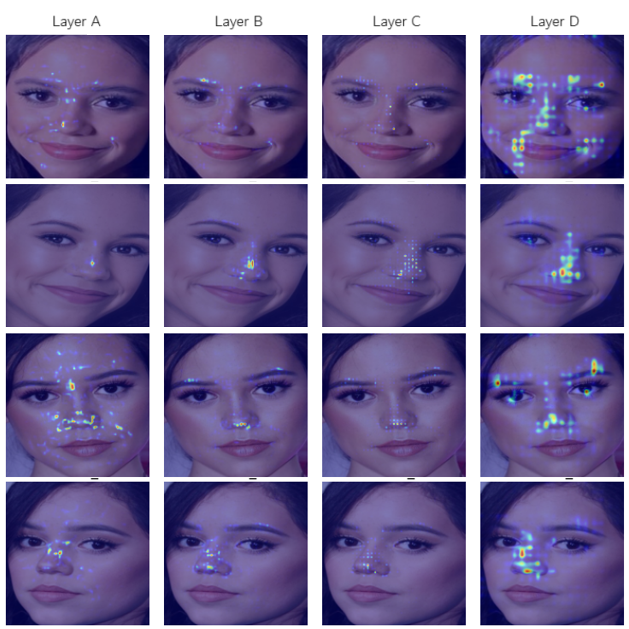}
    \caption{LEAM activation comparison for multiple images across selected layers}
    \label{fig:LEAM_dif}
    \vspace{-1.0em}
\end{figure}

This systematic approach enables us to precisely quantify which facial regions contribute most to recognition for different individuals, models, and network layers (Figure~\ref{fig:LEAM_dif}).

\subsection{LEAM similarity}
To assess consistency among activation mappings, we employ two complementary approaches:

First, we conduct statistical analysis of activated classes per individual, which reveals underlying patterns and potential biases in how different models or layers attend to facial regions.

Second, we treat embedding activation maps as 2D probability distributions by normalizing pixel values. To measure similarity between two normalized maps P and Q across coordinates \(\mathcal{X}\), we calculate the Bhattacharyya Coefficient~\cite{Bhattacharyya}:

\begin{equation}
BC(P, Q) = \sum_{x \in \mathcal{X}} \sqrt{P(x) Q(x)}
\end{equation}

This produces values between 0 (no similarity) and 1 (perfect similarity). However, since facial images vary in angle and expression (e.g. open vs. closed mouth), achieving pixel-perfect alignment is challenging. Therefore, to reduce the impact of this issue and achieve the highest possible precision, an approach of rotating and repositioning pictures to the common position was designed. To that end, we implement a multi-step face alignment process:

\begin{enumerate}
    \item We randomly select a baseline reference image for each identity.
    \item We detect facial landmarks using the \textit{dlib} library~\cite{dlib}.
    \item  We rotate images based on eye corner positions to achieve vertical facial symmetry.
    \item We appropriately adjust other images of the same individual by comparing landmark positions to the reference, towards achieving the best possible overlap of facial landmarks.
\end{enumerate} 

As a complementary similarity metric, we use Earth Mover's Distance (EMD)~\cite{rubner2000earth}, which quantifies the minimum ``work'' required to transform one distribution into another. Unlike the Bhattacharyya Coefficient, EMD can detect similarity even when activations are not perfectly aligned. Despite its high computational complexity \(O(N^3 \log N)\), its robustness makes it valuable for our analysis. We implement EMD using the Python Optimal Transport (POT) library~\cite{flamary2021pot}.

\subsection{Pretrained models}
Facial recognition models are complex structures, often coming in multiple variants~\cite{Elharrouss_2024} that balance two key metrics -- accuracy and processing speed. In general, the more layers/building blocks in the model, the slower and the more accurate the model is. Additionally, due to the usage of particular filters, various layers in the model may have various hidden properties that make them focus on particular facial elements. Uncovering such patterns could be helpful in designing more efficient adversarial attacks.

As facial recognition models work with embeddings similarity, there is no need for an individual to be in the training dataset. It is a huge advantage, as any properly pre-trained model should be able to recognize almost every individual given a pair of their images. 
For that reason, a potential threat actor can resort to publicly available repositories and, as such, does not need to use additional computational resources to train their own model.  Therefore, to ensure the generalizability of our findings, we test LEAM across diverse, publicly available pre-trained facial recognition models. This approach reflects real-world scenarios where potential privacy threats come from existing models rather than custom-trained ones.
We evaluate three families of models:

\begin{itemize}[leftmargin=*]
    \item  InceptionResnetV1~\cite{SzegedyEtAl2015} implementations from the facenet-py\-torch library~\cite{facenetpytorch} trained on two different datasets: VGG-Face2~\cite{CaoEtAl2018} and CASIA-WebFace~\cite{YiEtAl2014}.
    \item VGG16~\cite{CaoEtAl2018} reconstructed from official PyTorch weights\cite{vggweights}, originally trained on the VGG-Face dataset.
    \item Invertible Residual Networks (IResNet)~\cite{behrmann2019invertible} from InsightFace Model Zoo~\cite{insightface} in various configurations (R18, R50, R100) with different numbers of layers. These models were trained using different loss functions (ArcFace~\cite{deng2018arcface}, CosFace~\cite{wang2018cosfacelargemargincosine}) and on different datasets (GLINT-360K~\cite{an2021partialfctraining10}, MS1MV2~\cite{deng2018arcface}, MS1MV3~\cite{DengEtAl2019}).
\end{itemize}

%% file: Sections/experiments.tex
\newtcolorbox{takeawaybox}[1]{%
  colback=blue!5!white,
  colframe=blue!75!black,
  title=#1
}

\section{Experiments}

This section presents an empirical evaluation of LEAM.
 Our experiments are designed to answer seven key research questions: 
 \begin{itemize}
     \item[\ding{172}] What visual patterns are highlighted by LEAM? 
     \item[\ding{173}] Where and how intensely do models focus their attention?
     \item[\ding{174}] What activation patterns emerge at the layer level? 
     \item[\ding{175}] Can we identify key facial landmarks for individual recognition?
     \item[\ding{176}] What is the impact of the most relevant pixel thresholds on key facial landmarks?
     \item[\ding{177}] Are individual-specific patterns transferable across different models?
     \item[\ding{178}] Is LEAM consistent across demographic variables, including gender, ethnicity, and age?
 \end{itemize}

Manual inspection was our starting point for pattern discovery (RQ1), with all subsequent analyses (RQ2-RQ7) being fully automated and quantitative. Our structure follows a logical progression from visual pattern identification (RQ1) to model-level analysis (RQ2-RQ3) and finally individual-level analysis (RQ4-RQ7). Each research question builds on previous findings to assess the personalization potential for privacy applications.

Through systematic analysis, we demonstrate how LEAM provides insights into recognition patterns that can inform privacy-preserving techniques. We begin by describing our goals and experimental setup, followed by detailed findings for each research question.

\subsection{Experimental setup}
For our evaluation dataset, we acquired images of celebrities using links collected as part of the IMDb-Face \cite{wang2018} dataset. We consider this approach ethically appropriate as these are publicly available images of public figures, and our research aims to contribute towards privacy protection. Due to changes in online content, only about 300k images of over 20k identities (from the original 1.7M pictures of 59k individuals) were available. These data are sufficient for drawing meaningful conclusions.

We selected nine pretrained facial recognition models for our evaluation: two variants of InceptionResnetV1, VGG16, and three pairs of IResNets (IResNet\_R18, IResNet\_R50, and IResNet\_R100), each with both CosFace (trained on GLINT-360K) and ArcFace (trained on MS1MV2 or MS1MV3) versions. For clarity in presenting results, we use the following abbreviations: ArcFace-based IResNet (AF), CosFace-based IResNet (CF), and InceptionResnetV1 (IR).

Since LEAM is a visual technique and to avoid prior assumptions, we dedicated our first RQ to a deliberately unstructured manual assessment.

Structured approach and scalable automation were our goals for subsequent RQs. Following the initial assessment, we performed automated analysis on a larger scale. Due to computational constraints, we focused on 1000 individuals with 10-20 images each. 
For each model, we analyzed nine convolutional layers, selecting those closest to the input to maximize resolution and precision in mapping activations to facial regions. To ensure comparability, we used identical layers across models with the same backbone architecture.

Our automated analysis examined both full layer activation maps (using threshold=0.01 to exclude only marginally activated pixels) and the most activated pixels. Based on analyzing the relationship between cosine similarity drop and occlusion size, we primarily focused on the top 1\% of most relevant pixels, as this threshold significantly affects recognition accuracy and gives potential for visually inconspicuous adversarial attacks -- an important consideration for privacy protection applications.

\subsection{[RQ1] Visual patterns highlighted by LEAM}

\textbf{Goal:} 
We examine visually how activation is distributed across facial areas to compare neural networks with human perception. We assess pattern consistency and look for outliers causing unexpected behavior.

\noindent \textbf{Experiment:} 
We manually assessed LEAM outputs across all layers and models in our study on selected images. We have chosen an unstructured approach to avoid any prior assumptions.

\begin{figure*}[!ht]
    \centering
    \includegraphics[width=0.9\textwidth]{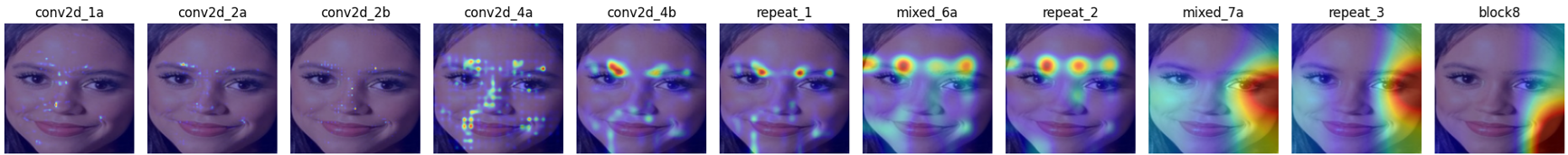}
    \caption{LEAM layer-by-layer comparison for IR\_CASIA}
    \label{fig:LEAM_overlayed}
\end{figure*}

\begin{figure*}[!ht] 
    \centering
    \includegraphics[width=0.9\textwidth]{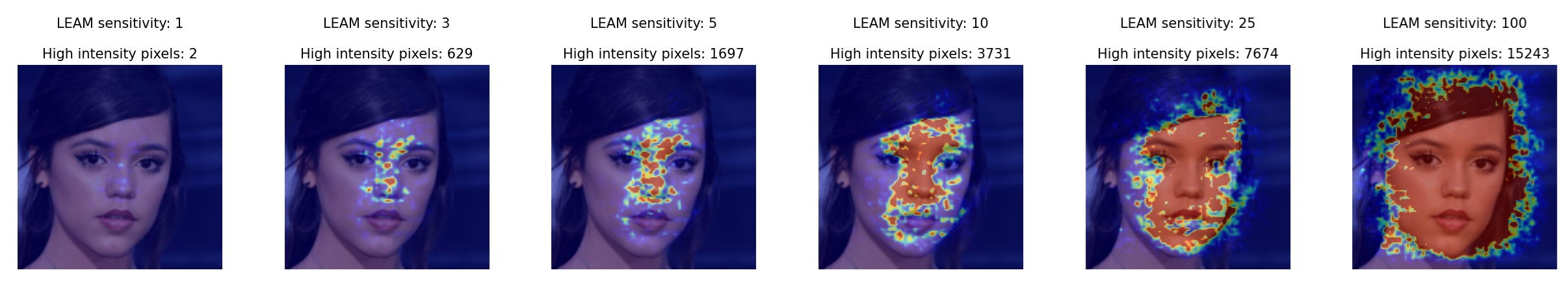}
    \caption{LEAM sensitivity adjustment reveals an abundance of regions with lower intensity positive contribution}
    \label{fig:LEAM_sensitivity}
\end{figure*}

\noindent \textbf{Results:} 
Our layer-by-layer analysis (Figure~\ref{fig:LEAM_overlayed}) shows that different network layers focus on different facial regions. Early layers highlight very specific pixels,
while layers closer to output often produce ``blob-like'' activations that extend beyond facial boundaries due to their lower resolution feature maps, which cannot precisely be correlated with specific facial landmarks. The primary reason for such behavior is related to the structure of neural networks. Usually, the further the layer, the lower its dimensions. For example, 3x3 feature map of a convolution layer in \textit{block8} of InceptionResnetV1 will produce just 9 values that cannot be reliably correlated with any facial landmark. 
These findings guided our decision to focus on layers closer to the input in our automated analysis, since their dimensions are sufficient to draw meaningful conclusions.

\begin{figure}[!h] 
    \centering
    \includegraphics[width=0.40\textwidth]{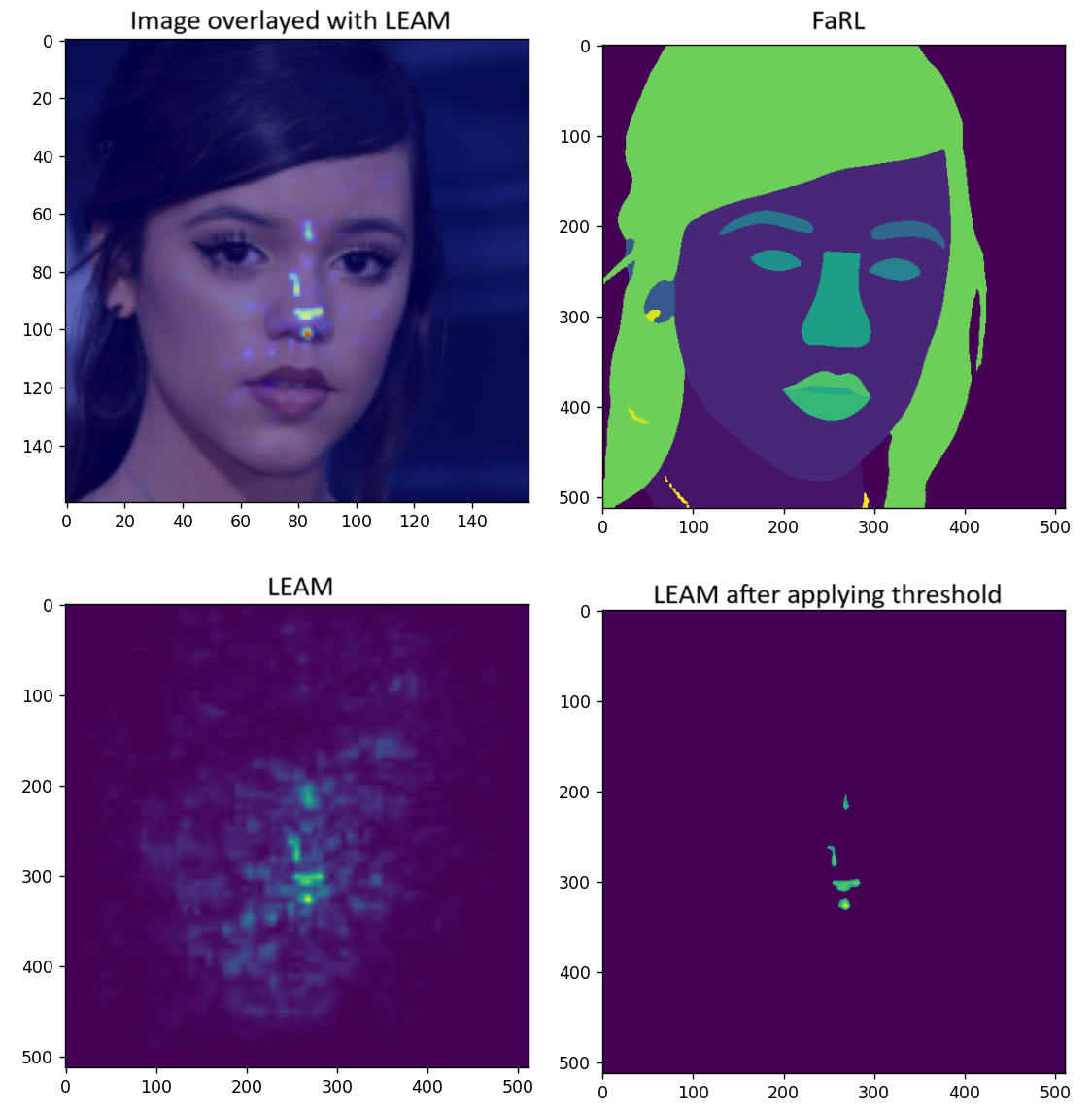}
    {\footnotesize
    \begin{tabular}{|c|c|c|c|}
        \hline
        & Pixels & Absolute \% & Relative \% \\
        \hline
        face & 69 & 0.03 & 11.82 \\
        \hline
        nose & 515 & 0.20 & 88.18 \\
        \hline
    \end{tabular}}
    \caption{Visualization of LEAM-FaRL correlation}
    \label{fig:Leam_farl}
\end{figure}

\begin{figure}[!h] 
    \centering
    \includegraphics[width=0.4\textwidth]{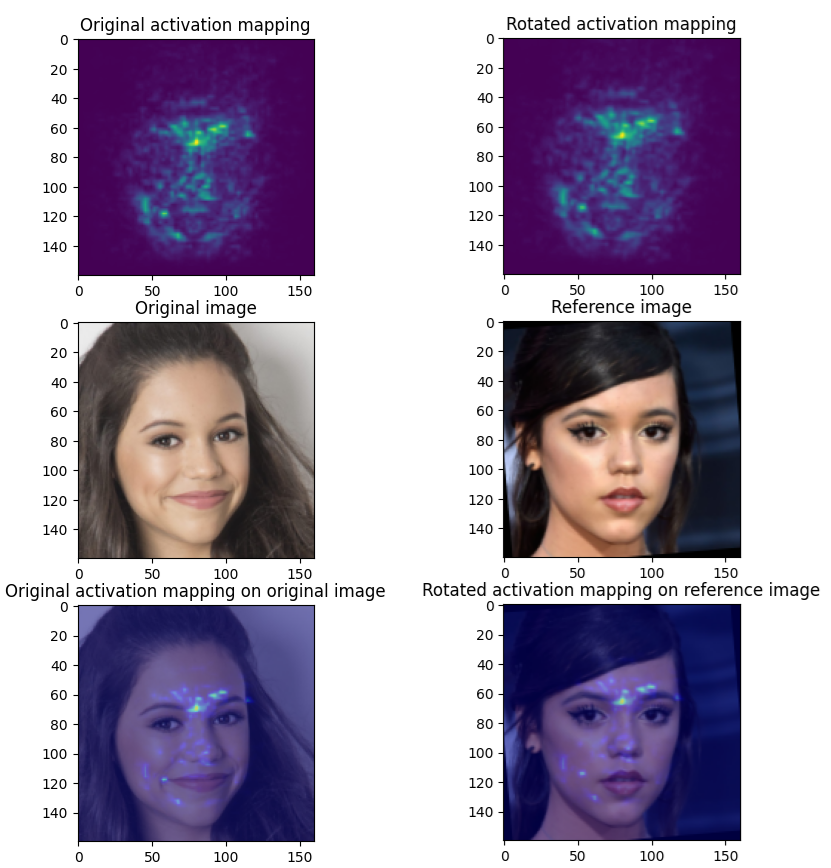}
    \caption{Impact of facial expressions on activation map transfer}
    \label{fig:Rotation}
\end{figure}

Our observations also indicate that visual representation of activation maps can be misleading about how models recognize faces. When analyzing such maps, one typically focus on the highest activation areas, which represent a small fraction of the image. Figure \ref{fig:Leam_farl} demonstrates this with the LEAM-FaRL correlation process. Using \textit{threshold=0.5} for LEAM preserves only the most intensely activated regions, with subsequent data indicating that over 88\% of relevant pixels belong to the nose area. 
This selective attention approach mirrors human visual processing, where neural systems~\cite{desimone1995neural} and retinal structure~\cite{retina2013} limit focus to a small portion of the visual field. 
However, Figure~\ref{fig:LEAM_sensitivity} reveals that in reality facial recognition models distribute attention more broadly, with thousands of less intensely activated pixels contributing at approximately 1/10$^{th}$ of the strength of the most relevant pixels. This distribution presents challenges for privacy protection, as targeting only the most activated regions may be insufficient.

When testing face alignment for activation mapping transferability, we found that facial expressions significantly impact landmark locations. Figure~\ref{fig:Rotation} shows that despite similar camera angles, different expressions change the mouth area's position, causing activations to map to different regions: activation at the left edge of the lips in the original image corresponds with the cheek area in the second image. Conversely, eyebrow regions maintained consistent alignment across images.

\find{
{\bf\ding{45}  Key take-aways of [RQ-1]}
\begin{itemize}[leftmargin=*]
    \item Convolutional layers located close to the input have higher resolution activation maps, hence, are more precise in underlining key facial regions.
    \item Similarly to humans perceiving a small part of the field of view at high resolution ~\cite{desimone1995neural,retina2013}, facial recognition models strongly focus only on a tiny fraction of all pixels. However, contrary to people, focus is not limited to one area, but distributed across the whole image. 
    \item During facial recognition the majority of the face is less intensely but still positively activated. It is a challenge for privacy preservation-oriented approaches.
\end{itemize}
}

\subsection{[RQ2] Focus areas of recognition models}

\noindent \textbf{Goal:}
We measure how strongly models look into various areas and which of them are the most relevant for facial recognition. We also assess the uniqueness of the findings across various models. 

\noindent \textbf{Experiment:}
We have performed two LEAM-FaRL correlation assessments. At first, we analyzed the results for the full activation to understand the models as a whole.
Then, we focused on a more inconspicuous scenario, where only the positions of the top 1\% of activated pixels are considered for potential attack.

\noindent \textbf{Results:}
Due to the volume of collected data, we provide detailed values on activated facial regions, discussed in this subsection, in Appendix \ref{Appendix_RQ2}.
When the attention of whole models is concerned, we could observe that they are consistent about the activated regions. Models tend to focus on almost the entirety of the facial area for positive detection, with the highest fraction of attention paid to classes such as \textit{face}, \textit{nose}, \textit{hair} and \textit{background}. 
While background can be eliminated during preprocessing, it is worth noting that a small portion of the attention is also paid to accessories that are not so easily removable, such as earrings or eyeglasses, and which can be leveraged to protect one's privacy. 
 
While the top 1\% of activated pixels were analyzed, we could observe a sharp increase in the focus on the classes belonging to the center of the face area -- encompassing eyes, nose, and mouth -- across all tested models. At the same time, there was a significant drop in the quantity of background and hair pixels. 
However, even for such a small threshold, activations do not appear to be strictly bound by particular physical class boundaries, but are still distributed across the face.
All facial areas remain at least slightly activated, being a very similar pattern when compared with total activation. This result suggests that the most relevant pixels are dispersed across different image areas, a finding that serves as proof of models' robustness and constitutes a challenge for privacy-preserving techniques.
Even though there are some differences between particular models, it can be observed that, as in the case of full activation, they all tend to yield similar results in terms of the order of the most relevant areas. 
Further, patterns are not more uniform when the same backbone or loss function is used. 
Overall, these findings suggest the potential for designing model-independent privacy-preserving adversarial techniques, but at the same time underline the challenging nature of creating inconspicuous occlusions in real-life scenarios.

\find{
{\bf\ding{45}  Key take-aways of [RQ-2]}
\begin{itemize}[leftmargin=*]
    \item For similar activation thresholds, tested models focus the most intensely on similar facial areas.
    \item When only 1\% of the most significant pixels are concerned, more attention is paid to the landmarks encompassing the center of the face
    and the impact of occlusions is reduced when compared with full activation.
    \item Most strongly activated facial areas are not bound by facial landmarks, a serious complication one has to overcome while designing real-life inconspicuous privacy-preservation techniques.
\end{itemize}
}

\subsection{[RQ3] Activation patterns at layer level}

\textbf{Goal:}
We assess how similar the layers are in their attention. 
We investigate the focus areas across the same and different models. 
We analyze if activation mappings are individual-dependent.

\noindent \textbf{Experiment:}
We have extended our analysis of the outcomes of the experiments from the previous research question to the layers. Further, we have compared activation mappings across the same and different individuals by calculating Bhattacharyya Coefficients and EMDs.

\noindent \textbf{Results:}
While analyzing models' total activations in the previous subsection, one could observe quite similar results. On the contrary, layer-by-layer patterns were quite diverse. While the biggest classes \textit{face}, \textit{hair}, \textit{background} and \textit{nose} still remained on top, there were a lot of changes in the prevalence of other facial areas. As shown in Table~\ref{tab:layer_diversity}, some layers can put much more attention to certain areas, such as in the case of the first convolutional layer \textit{conv1}, where one can notice significantly elevated focus on \textit{hair} and \textit{hat} classes.  
Overall, observed results imply that each layer has its unique function in FR process as they prioritize distinct facial areas.

\begin{table}[!h]
    \centering
    \caption{Percentage of CosFace\_R18 positive activation by layer}
    \vspace{-0.7em}
    \label{tab:layer_diversity}
    \resizebox{\columnwidth}{!}{
    \begin{tabular}{|l|c|c|c|c|c|c|}
     \hline
        Facial area & conv1 & l1\_0.conv1 & l1\_0.conv2 & l2\_0.conv1 & l2\_0.conv2 \\ \hline
        face & 40.73 & 47.57 & 52.41 & 58.76 & 51.88 \\ \hline
        hair & 21.06 & 9.52 & 10.41 & 7.12 & 12.24 \\ \hline
        background & 12.75 & 9.68 & 9.07 & 10.21 & 11.56 \\ \hline
        nose & 8.86 & 10.94 & 11.17 & 10.38 & 7.96 \\ \hline
        neck & 1.58 & 1.34 & 2.14 & 2.52 & 4.37 \\ \hline
        clothes & 2.20 & 1.68 & 1.71 & 1.14 & 2.92 \\ \hline
        right eyebrow & 2.12 & 2.88 & 2.03 & 1.57 & 0.96 \\ \hline
        left eyebrow & 1.70 & 2.50 & 1.74 & 1.55 & 0.94 \\ \hline
        lower lip & 1.17 & 2.28 & 1.73 & 1.35 & 1.56 \\ \hline
        upper lip & 1.20 & 2.31 & 1.82 & 1.10 & 1.20 \\ \hline
        right eye & 1.11 & 2.28 & 0.98 & 1.07 & 0.58 \\ \hline
        eyeglasses & 1.23 & 1.60 & 1.25 & 0.92 & 0.81 \\ \hline
        left eye & 0.96 & 2.32 & 0.90 & 1.02 & 0.55 \\ \hline
        hat & 1.79 & 0.93 & 1.13 & 0.49 & 0.98 \\ \hline
        inner mouth & 0.75 & 1.31 & 0.98 & 0.43 & 0.60 \\ \hline
        right ear & 0.34 & 0.40 & 0.29 & 0.16 & 0.43 \\ \hline
        left ear & 0.33 & 0.36 & 0.18 & 0.18 & 0.37 \\ \hline
        earrings & 0.11 & 0.09 & 0.05 & 0.02 & 0.08 \\ \hline
        necklace & 0.01 & 0.01 & 0.01 & 0.00 & 0.01 \\ \hline
    \end{tabular}}
\end{table}

In Table \ref{tab:layer_diversity_2} we have presented our results for the first convolutional layer of each of the IResNet models. Interestingly, when analyzing the very same layer, we have observed that different model variants focus on distinct areas. It suggests that both loss function and model size have a significant impact on the model's attention. 

\begin{table}[!h]
    \centering
    \caption{Percentage of \textit{conv1} layer positive activation by IResNet model}
    \vspace{-0.7em}
    \label{tab:layer_diversity_2}
    \resizebox{\columnwidth}{!}{
    \begin{tabular}{|l|c|c|c|c|c|c|}
    \hline
        Facial area & AF\_R18 & AF\_R50 & AF\_R100 & CF\_R18 & CF\_R50 & CF\_R100 \\ \hline
        face & 55.89 & 48.67 & 66.73 & 40.73 & 51.09 & 34.97 \\ \hline
        nose & 13.31 & 11.19 & 16.02 & 21.06 & 11.79 & 10.43 \\ \hline
        background & 6.94 & 15.22 & 7.14 & 12.75 & 13.62 & 22.74 \\ \hline
        hair & 6.90 & 14.04 & 2.72 & 8.86 & 13.49 & 12.81 \\ \hline
        neck & 1.39 & 1.34 & 1.07 & 1.58 & 0.89 & 2.32 \\ \hline
        lower lip & 2.73 & 0.85 & 0.65 & 2.20 & 0.61 & 2.89 \\ \hline
        clothes & 2.81 & 1.97 & 0.64 & 2.12 & 0.77 & 1.29 \\ \hline
        right eyebrow & 1.40 & 0.76 & 0.35 & 1.70 & 1.01 & 1.56 \\ \hline
        left eyebrow & 1.50 & 0.79 & 0.82 & 1.17 & 1.75 & 2.13 \\ \hline
        upper lip & 2.33 & 0.64 & 0.55 & 1.20 & 0.49 & 1.64 \\ \hline
        eyeglasses & 1.01 & 0.82 & 0.75 & 1.11 & 0.80 & 1.09 \\ \hline
        hat & 0.74 & 1.51 & 0.66 & 1.23 & 0.61 & 1.68 \\ \hline
        right eye & 0.71 & 0.58 & 0.44 & 0.96 & 0.58 & 1.56 \\ \hline
        left eye & 0.98 & 0.59 & 0.44 & 1.79 & 1.48 & 1.38 \\ \hline
        inner mouth & 0.75 & 0.66 & 0.70 & 0.75 & 0.53 & 0.88 \\ \hline
        right ear & 0.28 & 0.17 & 0.16 & 0.34 & 0.21 & 0.30 \\ \hline
        left ear & 0.26 & 0.16 & 0.15 & 0.33 & 0.22 & 0.23 \\ \hline
        earrings & 0.05 & 0.03 & 0.02 & 0.11 & 0.06 & 0.10 \\ \hline
        necklace & 0.00 & 0.00 & 0.00 & 0.01 & 0.00 & 0.01 \\ \hline
    \end{tabular}}
\end{table}

By comparing results from Tables \ref{tab:bh1} and \ref{tab:bh2}, that were obtained for the top 1\% activation scenario, one can notice that there is very high dissimilarity between Bhattacharyya Coefficients for images of the same and different identities. This state implies that models are capable of changing focus regions based on the assessed individual. While the results for InceptionResnets and VGG16 models were not shown in the table due to structural differences and layer name mismatch, the obtained values were within the same range of magnitude as the presented values. 
At the same time, activation maps obtained for images of the same identity are quite similar, allowing for the hypothetical existence of individual-oriented perturbations that could successfully target any picture.
Overall, this implies that regardless of the used model, we can expect quite diverse activation maps for different individuals, leading to the conclusion that a personalized approach can greatly benefit adversarial attacks. 

\begin{table}[!h]
    \centering
    \vspace{-0.3em}
    \caption{Average Bhattacharyya Coefficient per layer -- different images of the same identities}
    \vspace{-0.7em}
    \label{tab:bh1}
    \resizebox{\columnwidth}{!}{
    \begin{tabular}{|l|c|c|c|c|c|c|}
    \hline
        Layer & AF\_R18 & AF\_R50 & AF\_R100 & CF\_R18 & CF\_R50 & CF\_R100 \\ \hline
        conv1 & 0.32 & 0.28 & 0.32 & 0.21 & 0.28 & 0.20 \\ \hline
        l1\_0.conv1 & 0.49 & 0.37 & 0.40 & 0.42 & 0.47 & 0.44 \\ \hline
        l1\_0.conv2 & 0.44 & 0.48 & 0.45 & 0.41 & 0.48 & 0.49 \\ \hline
        l1\_1.conv1 & 0.32 & 0.52 & 0.51 & 0.35 & 0.46 & 0.37 \\ \hline
        l1\_1.conv2 & 0.47 & 0.40 & 0.47 & 0.45 & 0.45 & 0.21 \\ \hline
        l2\_0.conv1 & 0.55 & 0.54 & 0.54 & 0.54 & 0.54 & 0.53 \\ \hline
        l2\_0.conv2 & 0.44 & 0.51 & 0.46 & 0.45 & 0.45 & 0.49 \\ \hline
        l2\_1.conv1 & 0.57 & 0.47 & 0.32 & 0.54 & 0.54 & 0.45 \\ \hline
        l2\_1.conv2 & 0.43 & 0.40 & 0.35 & 0.38 & 0.48 & 0.38 \\ \hline
    \end{tabular}}
\end{table}

\begin{table}[!h]
    \centering
    \caption{Average Bhattacharyya Coefficient per layer -- different identities}
    \vspace{-0.7em}
    \label{tab:bh2}
    \resizebox{\columnwidth}{!}{
    \begin{tabular}{|l|c|c|c|c|c|c|}
    \hline
        Layer & AF\_R18 & AF\_R50 & AF\_R100 & CF\_R18 & CF\_R50 & CF\_R100 \\ \hline
        conv1 & 0.09 & 0.07 & 0.09 & 0.06 & 0.07 & 0.05 \\ \hline
        l1\_0.conv1 & 0.12 & 0.08 & 0.09 & 0.10 & 0.11 & 0.10 \\ \hline
        l1\_0.conv2 & 0.11 & 0.11 & 0.11 & 0.09 & 0.11 & 0.11 \\ \hline
        l1\_1.conv1 & 0.05 & 0.09 & 0.09 & 0.06 & 0.10 & 0.06 \\ \hline
        l1\_1.conv2 & 0.12 & 0.09 & 0.11 & 0.11 & 0.11 & 0.05 \\ \hline
        l2\_0.conv1 & 0.13 & 0.12 & 0.11 & 0.13 & 0.12 & 0.11 \\ \hline
        l2\_0.conv2 & 0.09 & 0.12 & 0.10 & 0.11 & 0.10 & 0.12 \\ \hline
        l2\_1.conv1 & 0.13 & 0.07 & 0.04 & 0.11 & 0.11 & 0.08 \\ \hline
        l2\_1.conv2 & 0.09 & 0.06 & 0.05 & 0.09 & 0.10 & 0.09 \\ \hline
    \end{tabular}}
\end{table}

\begin{table}[!h]
    \centering
    \caption{Average EMD per layer -- different images of the same identities}
    \vspace{-0.7em}
    \label{tab:emd1}
    \resizebox{\columnwidth}{!}{
    \begin{tabular}{|l|c|c|c|c|c|c|}
    \hline
        Layer & AF\_R18 & AF\_R50 & AF\_R100 & CF\_R18 & CF\_R50 & CF\_R100 \\ \hline
        conv1 & 7.73 & 10.22 & 8.62 & 10.56 & 10.18 & 12.81 \\ \hline
        l1\_0.conv1 & 5.99 & 9.19 & 8.17 & 7.78 & 6.20 & 7.08 \\ \hline
        l1\_0.conv2 & 8.52 & 7.55 & 6.40 & 8.70 & 7.32 & 7.50 \\ \hline
        l1\_1.conv1 & 13.51 & 7.14 & 7.58 & 10.72 & 8.85 & 9.85 \\ \hline
        l1\_1.conv2 & 7.97 & 8.96 & 6.51 & 7.87 & 8.30 & 11.60 \\ \hline
        l2\_0.conv1 & 7.45 & 6.52 & 6.06 & 7.29 & 6.62 & 6.40 \\ \hline
        l2\_0.conv2 & 8.90 & 7.36 & 7.83 & 10.27 & 8.06 & 7.62 \\ \hline
        l2\_1.conv1 & 6.91 & 9.56 & 14.77 & 7.40 & 8.22 & 10.31 \\ \hline
        l2\_1.conv2 & 10.56 & 10.78 & 11.15 & 11.08 & 8.34 & 11.07 \\ \hline
    \end{tabular}}
\end{table}

Further, analyzing EMD values across Tables \ref{tab:emd1} and \ref{tab:emd2}, it is possible to observe a major increase in the average transformation cost across all layers when different identities were compared. Contrary to the Bhattacharyya Coefficient, EMD values are very layer and model-dependent. 
As previously highlighted, for each of the tested models, certain layers may tend to put more focus on particular areas, leading to relatively small values of distances between activations. While others could consider the image more holistically, resulting in activations in various parts of the picture. The disparity between EMD values for the same and different identities can serve as another proof of the existence of individual-based patterns that could be leveraged in adversarial attacks.

\begin{table}[h]
    \centering
    \vspace{-0.5em}
    \caption{Average EMD per layer - different identities}
    \vspace{-0.7em}
    \label{tab:emd2}
    \resizebox{\columnwidth}{!}{
    \begin{tabular}{|l|c|c|c|c|c|c|}
    \hline
        Layer & AF\_R18 & AF\_R50 & AF\_R100 & CF\_R18 & CF\_R50 & CF\_R100 \\ \hline
        conv1 & 13.89 & 18.92 & 14.82 & 18.65 & 18.10 & 20.34 \\ \hline
        l1\_0.conv1 & 11.54 & 16.33 & 15.77 & 14.03 & 11.75 & 12.92 \\ \hline
        l1\_0.conv2 & 14.71 & 14.92 & 12.99 & 14.68 & 14.49 & 14.48 \\ \hline
        l1\_1.conv1 & 28.12 & 14.46 & 16.47 & 23.07 & 17.24 & 20.17 \\ \hline
        l1\_1.conv2 & 14.71 & 15.26 & 13.07 & 13.43 & 14.00 & 20.82 \\ \hline
        l2\_0.conv1 & 14.47 & 13.36 & 13.11 & 13.83 & 13.71 & 13.20 \\ \hline
        l2\_0.conv2 & 15.60 & 14.70 & 15.41 & 15.94 & 15.46 & 14.42 \\ \hline
        l2\_1.conv1 & 15.03 & 22.10 & 29.31 & 16.19 & 16.90 & 19.92 \\ \hline
        l2\_1.conv2 & 19.15 & 21.77 & 21.75 & 18.14 & 17.03 & 19.38 \\ \hline
    \end{tabular}}
    \vspace{-1.0em}
\end{table}

\find{
{\bf\ding{45}  Key take-aways of [RQ-3]}\begin{itemize}[leftmargin=*]
    \item Regardless of whether the layers of the same or different models are compared, their focus varies substantially.
    \item Model's internal structure and training procedure have a significant impact on the position of the most relevant pixels at the layer level.
    \item Disparities between values of Bhattacharyya Coefficient and EMD for activation mappings of the same and different people's images suggest the existence of individual-based patterns that can be leveraged for privacy preservation.
\end{itemize}
}

\subsection{[RQ4] Key facial landmarks for an individual}

\textbf{Goal:}
We investigate whether we can identify individual-specific patterns that could inform the development of personalized privacy-preserving techniques.

\noindent \textbf{Experiment:}
We analyzed the LEAM-FaRL correlation results for the top 1\% of activated pixels at the individual level. To preserve privacy while illustrating our findings, we present results for ten sample individuals, denoted as \textit{I\_1} through \textit{I\_10}.

\noindent \textbf{Results:}  
As shown in Table \ref{tab:10_individuals}, activation patterns exhibit substantial variation across individuals. This variability supports a hypothesis that person-specific adversarial techniques could be more effective than generic approaches.

\begin{table}[h]
    \centering
    \caption{Activation per individual across tested models}
    \vspace{-0.7em}
    \label{tab:10_individuals}
    \resizebox{\columnwidth}{!}{
    \begin{tabular}{|l|c|c|c|c|c|c|c|c|c|c|}
    \hline
        Facial area & I\_1 & I\_2 & I\_3 & I\_4 & I\_5 & I\_6 & I\_7 & I\_8 & I\_9 & I\_10 \\ \hline
        face & 56.70 & 50.34 & 54.54 & 52.23 & 55.34 & 55.87 & 51.46 & 56.23 & 42.34 & 49.54 \\ \hline
        nose & 17.62 & 19.16 & 19.15 & 25.13 & 22.16 & 19.55 & 18.56 & 20.41 & 22.94 & 20.95 \\ \hline
        background & 5.66 & 8.91 & 4.99 & 1.35 & 5.40 & 9.59 & 12.84 & 5.27 & 1.76 & 1.08 \\ \hline
        hair & 1.02 & 6.79 & 2.45 & 4.08 & 2.73 & 3.91 & 4.21 & 5.35 & 9.06 & 5.86 \\ \hline
        eyeglasses & 0.00 & 1.58 & 0.00 & 6.70 & 2.86 & 0.00 & 1.33 & 0.00 & 14.96 & 6.84 \\ \hline
        right eyebrow & 2.62 & 2.97 & 2.86 & 1.73 & 2.34 & 2.49 & 3.20 & 1.80 & 3.54 & 3.81 \\ \hline
        left eyebrow & 2.59 & 1.73 & 3.17 & 1.85 & 2.84 & 1.56 & 1.07 & 5.09 & 2.30 & 3.42 \\ \hline
        right eye & 2.08 & 1.34 & 1.88 & 1.34 & 1.04 & 1.53 & 1.69 & 0.60 & 0.10 & 1.44 \\ \hline
        left eye & 2.33 & 0.80 & 1.99 & 1.48 & 0.96 & 1.12 & 0.98 & 1.28 & 0.13 & 1.30 \\ \hline
        upper lip & 1.20 & 1.37 & 1.80 & 0.79 & 1.21 & 1.13 & 1.31 & 1.18 & 0.87 & 0.81 \\ \hline
        clothes & 3.27 & 1.89 & 1.95 & 0.81 & 0.65 & 0.43 & 0.71 & 0.21 & 0.07 & 1.65 \\ \hline
        lower lip & 1.72 & 1.36 & 1.42 & 0.86 & 0.81 & 1.15 & 1.37 & 0.91 & 0.76 & 0.93 \\ \hline
        neck & 0.91 & 0.87 & 0.39 & 0.87 & 0.73 & 0.59 & 0.79 & 0.41 & 0.71 & 1.12 \\ \hline
        hat & 1.69 & 0.54 & 2.56 & 0.00 & 0.31 & 0.39 & 0.00 & 0.00 & 0.00 & 0.00 \\ \hline
        inner mouth & 0.01 & 0.17 & 0.45 & 0.58 & 0.21 & 0.26 & 0.24 & 0.84 & 0.29 & 0.89 \\ \hline
        right ear & 0.37 & 0.13 & 0.17 & 0.13 & 0.21 & 0.32 & 0.20 & 0.01 & 0.11 & 0.21 \\ \hline
        left ear & 0.23 & 0.06 & 0.24 & 0.07 & 0.18 & 0.12 & 0.05 & 0.40 & 0.08 & 0.17 \\ \hline
        necklace & 0.00 & 0.00 & 0.00 & 0.00 & 0.00 & 0.00 & 0.00 & 0.00 & 0.00 & 0.00 \\ \hline
        earrings & 0.00 & 0.00 & 0.00 & 0.00 & 0.00 & 0.00 & 0.00 & 0.00 & 0.00 & 0.00 \\ \hline
    \end{tabular}}
    \vspace{-1.0em}
\end{table}

Looking more closely at within-individual variation, we conducted a detailed statistical analysis for subject \textit{I\_1}, shown in Table~\ref{tab:individual}. The coefficient of variation (CV) reveals substantial variability in how models attend to different features across images of the same person. For instance, features such as inner mouth (CV=137.55), neck (CV=97.67), and background (CV=81.16) show high variability, while the overall face region remains more consistent (CV=10.14). This suggests that 
the specific sub-regions that drive recognition may shift significantly across different images of the same person, creating a challenge for a personalized adversarial approach. 

\begin{table}[!h]
    \centering
    \small
    \caption{Statistics of a tested individual}
    \vspace{-0.7em}
    \label{tab:individual}
    \begin{tabular}{|l|c|c|c|c|c|}
    \hline
        Feature & Std & Mean & Min & Max & CV \\ \hline
        background & 4.59 & 5.66 & 0.31 & 28.24 & 81.16 \\ \hline
        neck & 0.89 & 0.91 & 0.07 & 4.19 & 97.67 \\ \hline
        face & 5.75 & 56.70 & 37.58 & 67.87 & 10.14 \\ \hline
        clothes & 2.51 & 3.28 & 0.01 & 12.73 & 76.68 \\ \hline
        right ear & 0.33 & 0.37 & 0.00 & 1.19 & 88.92 \\ \hline
        left ear & 0.22 & 0.23 & 0.00 & 1.06 & 94.84 \\ \hline
        right eyebrow & 1.22 & 2.62 & 0.01 & 5.05 & 46.42 \\ \hline
        left eyebrow & 1.32 & 2.59 & 0.01 & 5.02 & 50.92 \\ \hline
        right eye & 1.08 & 2.08 & 0.00 & 4.97 & 52.08 \\ \hline
        left eye & 1.33 & 2.33 & 0.08 & 5.73 & 57.01 \\ \hline
        nose & 7.28 & 17.62 & 4.31 & 43.95 & 41.35 \\ \hline
        inner mouth & 0.02 & 0.01 & 0.00 & 0.09 & 137.55 \\ \hline
        lower lip & 0.71 & 1.72 & 0.42 & 3.60 & 41.17 \\ \hline
        upper lip & 0.48 & 1.20 & 0.14 & 2.52 & 40.27 \\ \hline
        hair & 0.88 & 1.02 & 0.00 & 3.99 & 86.91 \\ \hline
        hat & 1.43 & 1.69 & 0.00 & 5.94 & 84.63 \\ \hline
    \end{tabular}
    \vspace{-0.0em}
\end{table}

One limitation of our analysis framework is that the \textit{face} category, which accounts for approximately 50-56\% of the activated pixels across individuals, is quite general. This makes it challenging to identify precisely which facial microfeatures are most discriminative. However, this finding itself provides valuable insight: facial recognition models do not rely exclusively on discrete landmarks like eyes or nose, but rather integrate information from across the entire facial region, with different patterns of integration for different individuals, but also with a significant variability within an individual.

These results have significant implications for privacy protection. The high variability across individuals suggests that generic adversarial approaches may be less effective than personalized techniques that target the specific patterns most relevant to an individual's recognition. Furthermore, the within-individual variability that can be caused by plenty of factors -- those include but are not limited to lighting, expression, angle or appearance change (plastic surgery, new hairstyle, different make-up) -- indicates that robust privacy protection may need to adapt to these contextual factors.

\find{
{\bf\ding{45}  Key take-aways of [RQ-4]}\begin{itemize}[leftmargin=*]
    \item LEAM-FaRL correlation patterns vary substantially between individuals, providing strong evidence for the potential effectiveness of personalized privacy-protection approaches.
    \item Image-specific factors significantly impact model focus for the same identity, suggesting a need for adaptive privacy protection strategies.
\end{itemize}
}

\subsection{[RQ5] Impact of the most relevant pixels thresholds on key facial landmarks}

\textbf{Goal:}
We research the priority of facial landmarks. By changing the threshold of assessed pixels, we want to get greater insight into the relevance of each of the 19 FaRL classes.

\noindent \textbf{Experiment:}
We have analyzed average results of LEAM-FaRL correlation performed on the following pixel relevance thresholds: 0.01, 0.05, 0.1, 0.25, 0.5, 0.75, 1, 1.25, 1.5, 1.75, 2, 3, 5, 10 and 100 percent of the most relevant pixels.

\noindent \textbf{Results:} Detailed results of the experiments, described in this section, can be found in Appendix~\ref {Appendix_RQ5}.
We have observed a few important patterns across all the assessed models. The lower the threshold the lower the ratio of pixels belonging to non-face classes such as \textit{background}, \textit{hair}, \textit{earings}, \textit{neck}, \textit{clothes} and \textit{hat}. It suggests that even if models focus on occlusions and face surroundings, they put relatively low importance on them. Further, we have also noticed the same characteristic for \textit{left ear}, \textit{right ear} and \textit{face} classes. 
At the same time, there is an increase in the share of \textit{nose}, \textit{left eye}, \textit{right eye}, \textit{eyeglasses}, \textit{left eyebrow}, \textit{right eyebrow} classes.
Those two findings combined suggest that the most relevant pixels tend to be more concentrated in the very central area of a face. 
Interestingly, patterns of mouth-related classes are unique in its diversity. Frequently one can observe the decrease of their importance with the lower threshold values. Nonetheless, there are cases when it firstly increases to peak around 1\%. It suggests that the mouth is important, however, it does not contribute towards recognition as significantly as central facial area. Hence, it may be valid, but not necessarily optimal target for privacy-oriented adversarial attacks.

Another relevant pattern was observed when symmetrical facial parts were analyzed. Elements such as eyes, ears and eyebrows were characterized by the results of the left and right organs being almost identical across all tests. It suggests that the image alignment step, that at the same time strives for face symmetry, is critical in activation transfer across photos of the same identity.  

\find{
{\bf\ding{45}  Key take-aways of [RQ-5]}\begin{itemize}[leftmargin=*]
    \item Nose, eyes, and eyebrows regions tend to be the most strongly activated pixels. Non-face-related classes have low activation levels.
    \item Mouth area is fairly relevant. Yet, it is not as impactful as the aforementioned facial landmarks.
    \item Models tend to pay similar levels of attention to vertically symmetric face parts.  
\end{itemize}
}

\subsection{[RQ6] Transferability of individual-specific patterns across different models}
\label{(1)and(2)}

\noindent \textbf{Goal:}
To validate that LEAM correctly identifies important facial regions (rather than to develop an attack method), we conduct a proof-of-concept experiment using simple black pixel occlusions.
We assess and compare the relation between the perturbed area and the cosine similarity scores to the random approach. Finally, we investigate whether occlusion positions targeting one model transfer effectively to other models. 

We emphasize that this experiment is designed solely for validating LEAM's identification of important regions, not for developing adversarial perturbations.

\noindent \textbf{Experiment:} 
Contrary to privacy protection tools, which focus on absolute effectiveness against facial recognition models, with LEAM we only aim to guide occlusions, not to decide their contents. Hence, we have decided to compare the cosine similarity score drops of pictures perturbed with validation occlusions, based on LEAM-identified regions, in the form of black pixels. For that purpose, we have generated two types of masks. The first was created based on the positions indicated as most relevant by LEAM for IR\_CASIA. In the second approach positions of pixels to be perturbed were chosen randomly. We have tested various occlusion strengths by changing the quantity of masked pixels. Our experiments target the first convolutional layers of each model to enable transferability assessment.

\noindent \textbf{Results:}
Firstly, we have compared the LEAM-guided approach with a random mask to showcase the benefits of individual-oriented privacy protection techniques. As shown in Table \ref{tab:Similarity_drop}, LEAM allowed to significantly increase cosine similarity drop regardless of the size of concealed area.

\begin{table}[!h]
    \centering
    \small
    \vspace{-0.2em}
    \caption{Average cosine similarity drop across all models}
    \vspace{-0.2em}
    \label{tab:Similarity_drop}
    \begin{tabular}{|c|c|c|}
    \hline
        Occlusion (\%) & LEAM Occluded & Randomly Occluded \\ \hline
        0.01 & 0.0042 & 0.0003 \\ \hline
        0.05 & 0.0162 & 0.0016 \\ \hline
        0.10 & 0.0281 & 0.0031 \\ \hline
        0.25 & 0.0520 & 0.0074 \\ \hline
        0.50 & 0.0783 & 0.0153 \\ \hline
        0.75 & 0.0978 & 0.0223 \\ \hline
        1.00 & 0.1148 & 0.0286 \\ \hline
        1.25 & 0.1293 & 0.0343 \\ \hline
        1.50 & 0.1429 & 0.0407 \\ \hline
        2.00 & 0.1668 & 0.0511 \\ \hline
        5.00 & 0.2640 & 0.1016 \\ \hline
        10.00 & 0.3371 & 0.1673 \\ \hline
    \end{tabular}
    \vspace{-0.7em}
\end{table}

\begin{figure}[h] 
    \centering
    \includegraphics[width=0.47\textwidth]{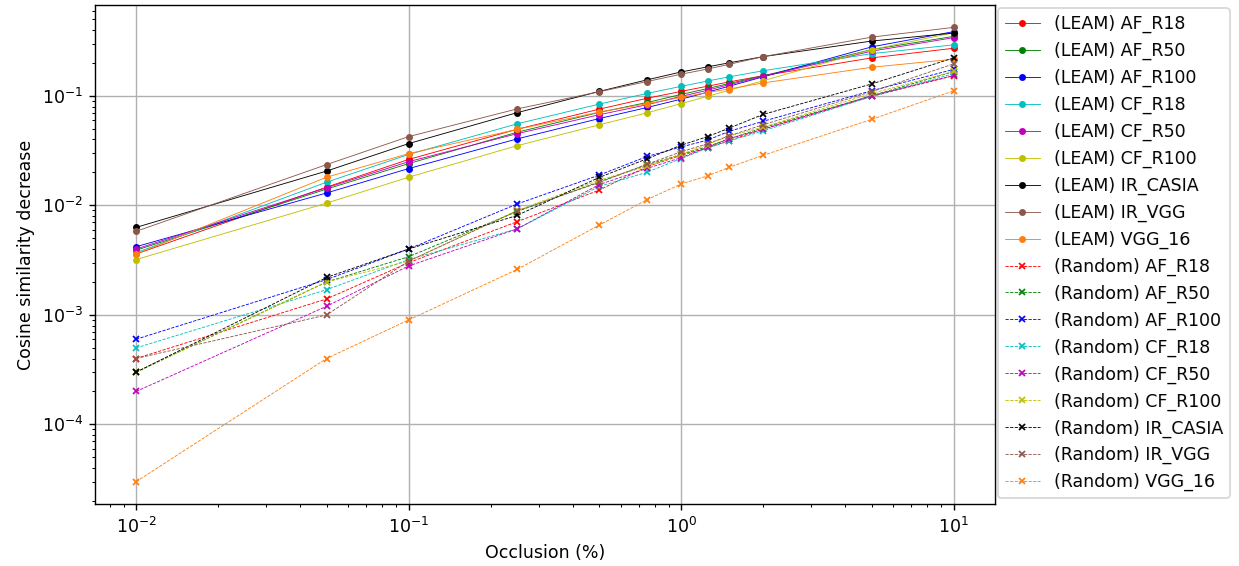}
    \vspace{-1.2em}
    \caption{Cosine similarity drop with occlusion per model}
    \vspace{-0.4em}
    \label{fig:Similarity_drop_model}
\end{figure}

Further, we have split our results to showcase cosine similarity drop per model, but due to the quantity of collected data, precise numerical results were presented in Appendix~\ref{Appendix_RQ6}.
In Figure \ref{fig:Similarity_drop_model}, one can observe that LEAM consistently and considerably outperforms the random approach, proving that the activation mapping generated for one model is transferable across all the tested models. Unsurprisingly, we uncovered that the greater the magnitude of the obstruction, the more the models' performance decreases. 
Furthermore, we have noticed that the similarity drop was the largest for IR\_CASIA, which was used to generate LEAM, followed by IR\_VGG, which shares the same backbone with the aforementioned model.
It is also worth pointing out that IResNet model sizes also had a clear impact on their performance. 
For larger, one can observe a smaller similarity drop when occlusion size is below 2\%. On the contrary, the decrease is much more significant for 5\% and 10\% occlusion. It implies that the larger neural networks are more robust against the transferability of inconspicuous privacy-protecting adversarial attacks, as they typically affect a small fraction of the face.

\find{
{\bf\ding{45}  Key take-aways of [RQ-6]}
\begin{itemize}[leftmargin=*]
    \item Simple LEAM-guided validation occlusions significantly decrease facial similarity during recognition, proving that LEAM's choice of the relevant pixels is valid.
    \item Occlusions generated for one model are transferable to other models. There are indications that models sharing the same backbone or small in size could be more susceptible to transfer attacks. 
\end{itemize}
}

\subsection{[RQ7] Gender, ethnicity and age variation ablation study}
\textbf{Goal:}
Imbalances of gender and ethnicity in the dataset often deteriorate the facial recognition model's accuracy for the underrepresented groups. For that reason, we would like to check LEAM consistency across various demographic groups. 

\noindent \textbf{Experiment:}
We have assigned demographic attributes to our identities -- \textit{Male} or \textit{Female} for gender and \textit{Caucasian}, \textit{African}, \textit{East Asian} or \textit{South Asian} for ethnicity.
To assess the influence of age variation between pictures, we have acquired access to FG-NET \cite{FG-NET2015} dataset, composed of 1002 images of 82 identities with a maximal age difference of 54 years. 
We have subsequently generated random and validation black pixel occlusions based on LEAM-identified regions with multiple thresholds of the most relevant pixels to assess their impacts on different demographic groups.

\noindent \textbf{Results:}
Due to the volume of collected data, we provide detailed values on cosine similarity decrease within ablation study in Appendix \ref{Appendix_RQ7}.

Analysing results split by gender
we have two main observations. Firstly, LEAM-guided occlusions consistently outperform the random method, again proving LEAM's ability to select relevant pixels. Secondly, while models tend to provide slightly lower average unoccluded similarity scores for women than for men -- respectively 0.4643 and 0.4817 -- similarity drop is almost always greater for females with the average LEAM-guided decrease of 0.1248 vs. 0.1038 for males, for 1\% occlusion threshold. 
Except for random occlusions smaller than 0.1\% threshold, the Mann–Whitney U test returned p-values below 10e-4, implying a statistically significant difference. 
Our finding suggests that due to underlying biases, which existence is supported by the results of both random and LEAM-guided occlusions, facial recognition privacy-protecting techniques could be more effective for women. One can hypothesize that this phenomenon stems from models struggling to address the inter-class variance caused by make-up, diverse hairstyle choice, or plastic surgeries observed more frequently across pictures of female identities.  

\begin{figure}[h] 
    \centering
    \includegraphics[width=0.48\textwidth]{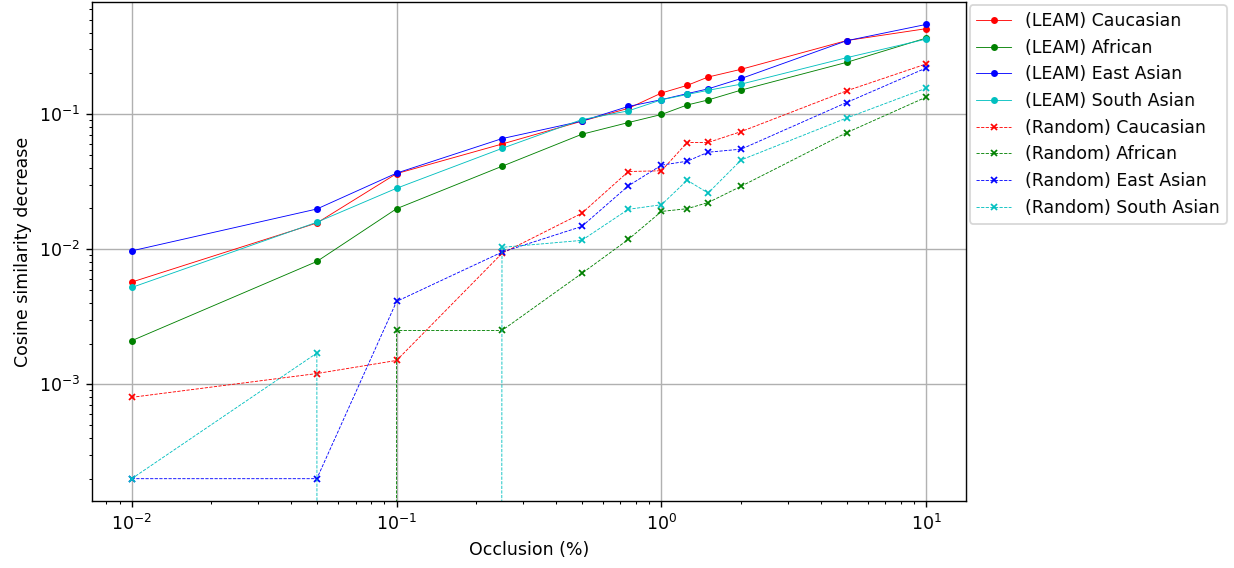}
    \vspace{-2.5em}
    \caption{Average cosine similarity drop per ethnicity}
    \label{fig:Ethnic_study}
    \vspace{-0.5em}
\end{figure}

Figure \ref{fig:Ethnic_study} presents a comparison of LEAM efficiency across ethnic groups. It can be observed that the cosine similarity decrease is the highest for East Asian and Caucasian individuals. The same pattern can be observed for random occlusions, an outcome that can be attributed to the combination of underlying biases and to the usage of black pixels as a mask that would be expected to have a higher impact on individuals of lighter skin tones. Regardless of the differences in results, one can conclude that LEAM-guided methods can be effective in reducing cosine similarity values across all ethnicities.

\begin{figure}[h] 
    \centering
    \vspace{-0.5em}
    \includegraphics[width=.45\textwidth]{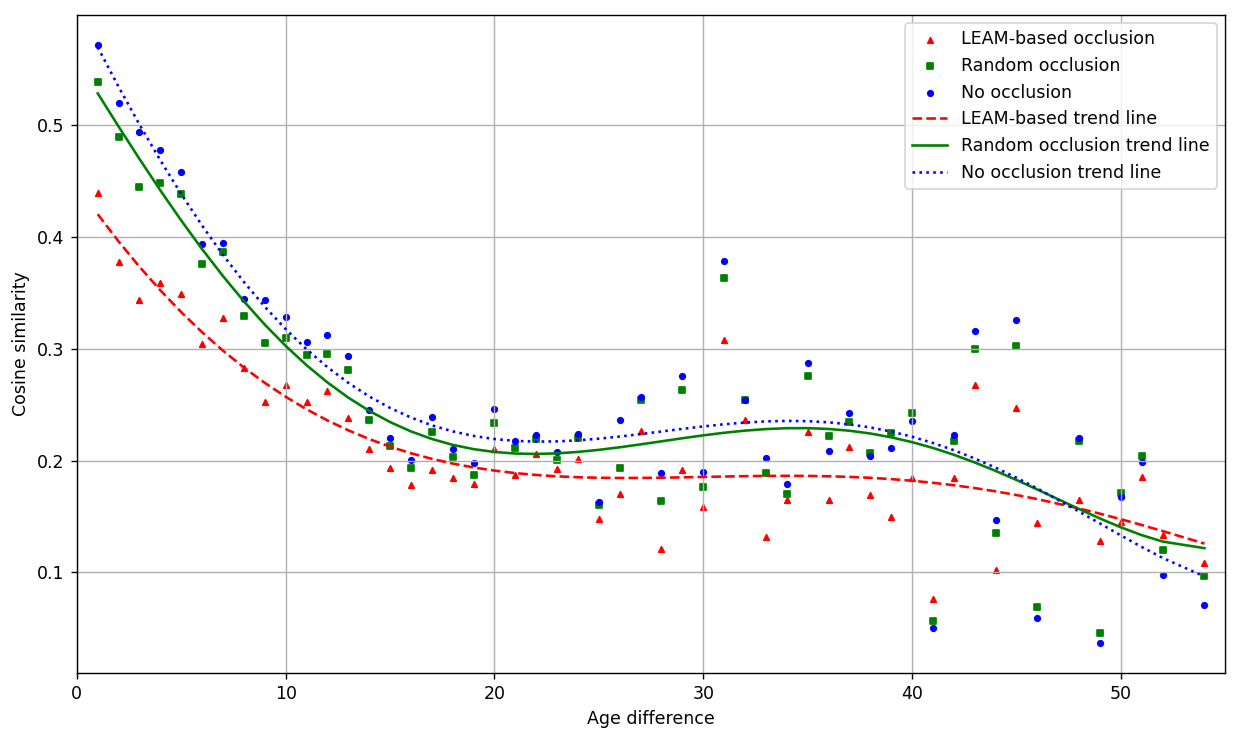}
    \vspace{-0.5em}
    \caption{Influence of age variance on cosine similarity -- 1\% occlusion}
    \vspace{-0.5em}
    \label{fig:Age_occlusion}
\end{figure}

\begin{figure}[h]
    \centering
    \vspace{-0.3em}
    \includegraphics[width=.45\textwidth]{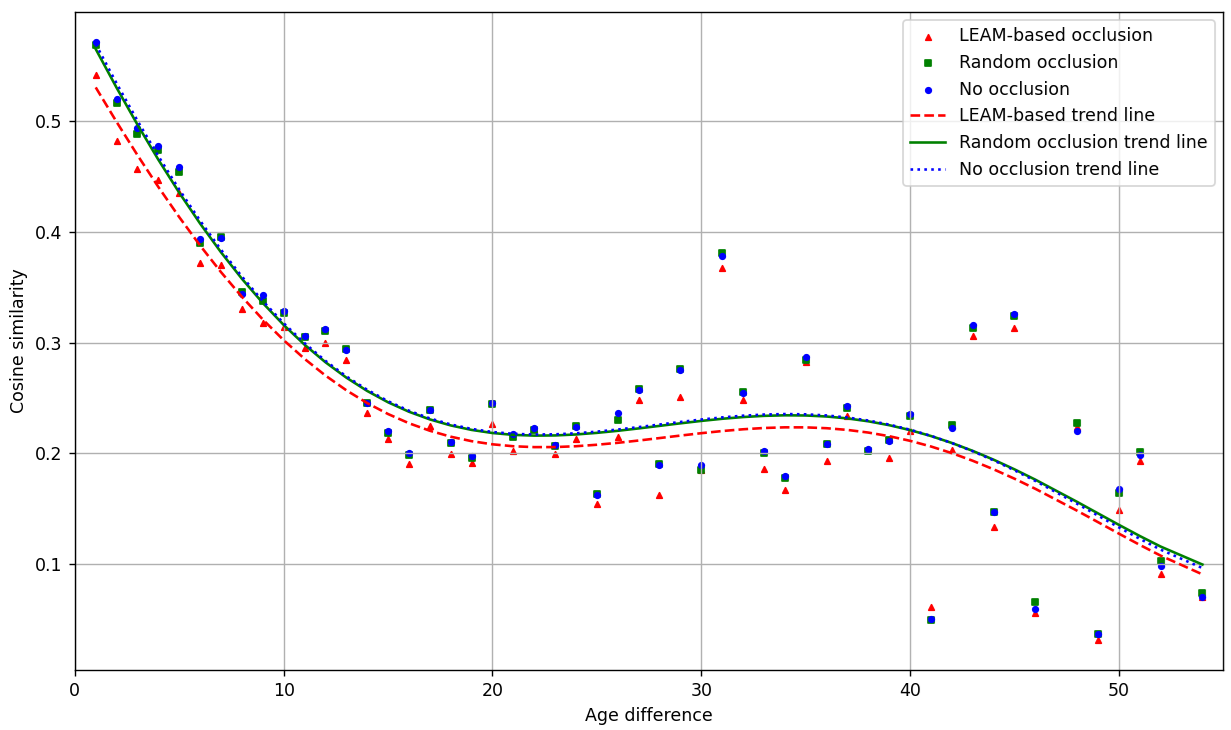}
    \vspace{-0.5em}
    \caption{Influence of age variance on cosine similarity -- 0.1\% occlusion}
    \label{fig:Age_occlusion_low}
    \vspace{-0.5em}
\end{figure}

Figure \ref{fig:Age_occlusion} presents the average impact of the age variance on the cosine similarity score across all tested models. It can be observed that the LEAM-guided occlusions are the most efficient when the age difference is below 15 years. Around this time results start to become quite inconsistent. This phenomenon is expected as the greater the age variance, the less data can be assessed. 
On the other hand, the models themselves are not robust enough to handle such notable time differences with visible similarity drops even if occlusions are absent. 

Further, we have observed that the lower occlusion percentages lead not only to the already presented lower similarity drop, but also LEAM-guided occlusions seem to be significantly less efficient for outliers with greater similarity scores, as shown in Figure \ref{fig:Age_occlusion_low}.

\find{
{\bf\ding{45}  Key take-aways of [RQ-7]}\begin{itemize}[leftmargin=*]
    \item Both LEAM-guided and random perturbations result in greater recognition performance drop for images of females. This finding implies a higher potential for privacy preservation techniques to be successful when concealing an identity of a woman.
    \item LEAM-guided occlusions successfully reduce recognition similarity scores across different ethnicities at a comparable rate.
    \item LEAM proves to be effective in highlighting key facial areas even if images used for recognition are taken far apart in time.
\end{itemize}
}

%% file: Sections/discussions.tex
\section{Discussion}
Our findings provide important insights into facial recognition mechanisms and their privacy implications. Here, we discuss key interpretations and validity considerations.
\subsection{Distinction from Adversarial Methods}
Our work fundamentally differs from adversarial attack research in both goals and methods. While adversarial attacks aim to fool recognition systems through carefully crafted perturbations, LEAM is an explainability tool that visualizes how models make recognition decisions. We do not generate adversarial examples, train attack models, or propose evasion techniques. Instead, we provide insights into model behavior that could inform various applications, including improving model robustness and understanding privacy vulnerabilities.
\subsection{Individual-Specific Recognition Patterns}
The substantial variation in activation patterns across individuals challenges the notion that facial recognition relies on standard features. Instead, models develop unique recognition signatures for each person, mirroring findings in cognitive psychology about human face recognition~\cite{desimone1995neural}. This individual-specificity explains the effectiveness gap between LEAM-guided validation and random occlusions (0.1148 vs. 0.0286 similarity drop with just 1\% occlusion) and suggests previous uniform adversarial approaches~\cite{DongEtAl2019} may be fundamentally suboptimal.
\subsection{Model Attention Distribution}
Facial recognition models distribute attention across facial regions rather than focusing exclusively on specific landmarks, contrasting with prior work suggesting primary attention to high-contrast regions \cite{wang2018}. This distributed attention explains model robustness to partial occlusions. However, the concentration on the nose region when examining only the most activated pixels is noteworthy, aligning with studies identifying the nasal region's high discriminative power in 3D facial recognition \cite{DongEtAl2019}.
\subsection{Cross-Model Transferability}
LEAM-guided occlusions effectively transfer across different model architectures, suggesting commonalities in facial information processing despite architectural differences. This contrasts with prior research showing limited cross-model transferability of adversarial examples \cite{sharif2019general}. The implications are dual: privacy protections may generalize well across models, while system developers face a vulnerability that spans architectures.
\subsection{Demographic Considerations}
The higher efficacy of occlusions on female images raises important questions about fairness in privacy protection. This finding could stem from training data imbalances, differences in photography practices, or genuine differences in feature distinctiveness. These demographic variations underscore the importance of testing privacy protection techniques across diverse populations to avoid disparate outcomes.
\subsection{Limitations}
Facial recognition is subject to multiple challenges, such as variance in face position, lighting level or temporal changes. These factors influence how models process images and prevent the generation of a universal activation map for any individual. Consequently, potential LEAM-guided adversarial attack applied to a specific image creates a unique output optimized for that particular image. 
When an averaged activation mapping is transferred to other photos of the same person, it cannot guarantee equal protection against facial recognition across all images, but rather increases the overall probability of detection evasion. Finally, LEAM itself does not determine the type of the occlusion, but can only specify its location.

\subsection{Threats to Validity}

While our study provides valuable insights into facial recognition mechanisms, several factors could potentially affect validity. However, our experimental design choices mitigate many of these concerns:

\subsubsection{Dataset Considerations} 
Our use of celebrity images from IMDb may introduce certain biases due to professional photography conditions. However, this choice offers important advantages: the high diversity of individuals (1000 subjects with 10-20 images each), significant variation in capture conditions across each person's images, and the availability of multiple angles and expressions. Our filtering process, which excluded images where faces couldn't be detected, further ensured data quality.
97.2\% of images (13445/13838) were retained across all 1000 identities, and most importantly, no identity was eliminated in the process. Our approach was not biased toward high-quality, easily detectable faces, but rather helped to reduce the dataset noise.

\subsubsection{Model Selection}
Our deliberate selection of nine models spanning different architectures, loss functions, and training data\-sets, with consistent findings across all, suggests results extend beyond specific architectural choices.

\subsubsection{Activation Mapping Limitations}
LEAM, like all visualization techniques, approximates rather than perfectly captures network attention. However, our validation through multiple quantitative metrics and comparison between full and top 1\% activation provides strong evidence for the patterns identified.

\subsubsection{Occlusion Effects}
While we evaluated straightforward black pixel occlusions as our LEAM validation method, future work should explore state of the art occlusion methods. Nonetheless, our findings confirm LEAM's effectiveness in finding relevant facial areas and establish fundamental principles that would likely extend to other perturbations.

\subsubsection{Temporal Stability}
Our analysis of images with substantial time differences (using the FG-NET dataset) demonstrates that LEAM-guided occlusions remain effective despite temporal changes. While very long-term aging effects warrant further study, our experiments show that the individual-specific patterns LEAM identifies are robust across the range of appearance variations typically encountered in practical facial recognition scenarios. This suggests that personalized privacy interventions based on LEAM would maintain effectiveness for reasonable time periods.

%% file: Sections/conclusion.tex
\section{Conclusions}
This paper presents Layer Embedding Activation Mapping (LEAM), a novel technique that reveals which facial areas contribute most to recognition by adapting the concept of LayerCAM for embedding-based facial recognition models. By correlating LEAM with FaRL-based face parsing, we found that while models activate almost the entire facial area, they concentrate attention on the central regions, primarily the nose, when considering only the most activated pixels.
Our analysis revealed that different network layers focus on distinct facial areas, yet overall attention patterns remain consistent across models. Significantly, the dissimilarity of Bhattacharyya Coefficients and EMDs between the same and different individuals indicates that activation patterns are individual-specific rather than model-dependent, suggesting unique recognition signatures for each person. LEAM-guided validation occlusions transferred effectively across all tested models, significantly reducing similarity scores compared to random occlusions.
Our findings demonstrate that facial recognition models prioritize different facial regions for different individuals, with female images showing higher vulnerability to privacy-protecting interventions. 

These results have significant privacy implications, highlighting the need for personalized rather than generic approaches to privacy protection.
Future work should focus on developing individual-specific privacy protection systems based on LEAM guidance, developing more precise facial region classifications, and designing computationally efficient methods for comparing activation maps to facilitate more effective privacy-preserving techniques.

%% file: submission-template/main.bbl

%% file: Sections/appendix.tex
\begin{table*}[]
    \centering
    \small
    \caption{Percentage distribution of facial areas positively activated by model}
    \label{tab:all_features}
    \begin{tabular}{|l|c|c|c|c|c|c|c|c|c|}
    \hline
        Facial area & AF\_R18 & AF\_R50 & AF\_R100 & CF\_R18 & CF\_R50 & CF\_R100 & IR\_CASIA  & IR\_VGG & VGG\_16 \\ \hline
         face & 58.32 & 59.50 & 59.61 & 51.43 & 57.79 & 58.42 & 55.32 & 55.46 & 50.79 \\ \hline
        hair & 8.42 & 8.39 & 7.50 & 11.84 & 8.76 & 10.21 & 12.09 & 11.50 & 12.75 \\ \hline
        background & 7.47 & 8.74 & 9.05 & 9.03 & 8.77 & 7.74 & 10.95 & 10.76 & 12.70 \\ \hline
        nose & 10.76 & 10.79 & 12.02 & 10.24 & 10.95 & 10.58 & 8.28 & 7.95 & 11.16 \\ \hline
        neck & 2.10 & 1.57 & 1.15 & 2.42 & 1.38 & 0.96 & 2.24 & 2.70 & 0.51 \\ \hline
        clothes & 1.46 & 1.34 & 1.07 & 1.86 & 1.28 & 1.12 & 1.74 & 2.08 & 1.37 \\ \hline
        lower lip & 1.63 & 1.56 & 1.36 & 1.67 & 1.54 & 1.47 & 1.45 & 1.45 & 1.16 \\ \hline
        right eybrow & 1.73 & 1.28 & 1.35 & 2.03 & 1.50 & 1.71 & 1.35 & 1.34 & 1.55 \\ \hline
        left eyebrow & 1.75 & 1.22 & 1.29 & 1.78 & 1.62 & 1.66 & 1.32 & 1.32 & 1.50 \\ \hline
        upper lip & 1.53 & 1.12 & 1.10 & 1.57 & 1.34 & 1.27 & 1.11 & 1.13 & 1.02 \\ \hline
        hat & 0.70 & 0.82 & 0.91 & 0.96 & 0.82 & 0.91 & 0.95 & 0.90 & 1.53 \\ \hline
        eyeglasses & 0.99 & 0.90 & 0.95 & 1.12 & 1.02 & 1.05 & 0.85 & 0.88 & 1.15 \\ \hline
        right eye & 0.94 & 0.82 & 0.79 & 1.31 & 0.96 & 0.84 & 0.56 & 0.59 & 0.80 \\ \hline
        left eye & 0.92 & 0.81 & 0.78 & 1.24 & 0.99 & 0.84 & 0.56 & 0.59 & 0.80 \\ \hline
        inner mouth & 0.68 & 0.65 & 0.68 & 0.83 & 0.71 & 0.68 & 0.58 & 0.59 & 0.62 \\ \hline
        right ear & 0.30 & 0.24 & 0.20 & 0.32 & 0.25 & 0.27 & 0.31 & 0.36 & 0.28 \\ \hline
        left ear & 0.26 & 0.21 & 0.16 & 0.28 & 0.26 & 0.21 & 0.29 & 0.32 & 0.26 \\ \hline
        earrings & 0.05 & 0.03 & 0.03 & 0.07 & 0.04 & 0.05 & 0.06 & 0.07 & 0.04 \\ \hline
        necklace & 0.01 & 0.00 & 0.00 & 0.01 & 0.00 & 0.00 & 0.01 & 0.01 & 0.00 \\ \hline
    \end{tabular}
\end{table*}

\begin{table*}[]
    \centering
    \small
    \caption{Percentage distribution of top 1\% of the most positively activated pixels by model}
    \label{tab:all_features_1_percent}
    \begin{tabular}{|l|c|c|c|c|c|c|c|c|c|}
    \hline
        Facial area & AF\_R18 & AF\_R50 & AF\_R100 & CF\_R18 & CF\_R50 & CF\_R100 & IR\_CASIA & IR\_VGG & VGG\_16 \\ \hline
       face & 55.00 $\downarrow$ & 54.30 $\downarrow$& 56.57 $\downarrow$& 50.45 $\downarrow$& 55.68 $\downarrow$& 53.63 $\downarrow$& 50.52 $\downarrow$& 53.66 $\downarrow$& 48.51 $\downarrow$\\ \hline
        nose & 20.49 $\uparrow$& 20.07 $\uparrow$& 20.00 $\uparrow$& 18.57 $\uparrow$& 20.17 $\uparrow$& 18.87 $\uparrow$& 29.72 $\uparrow$& 26.19 $\uparrow$& 20.95 $\uparrow$\\ \hline
        background & 3.78 $\downarrow$& 5.49 $\downarrow$& 5.36 $\downarrow$& 5.34 $\downarrow$& 4.67 $\downarrow$& 5.48 $\downarrow$& 4.17 $\downarrow$& 4.38 $\downarrow$& 8.39 $\downarrow$\\ \hline
        hair & 3.15 $\downarrow$& 4.64 $\downarrow$& 4.11 $\downarrow$& 6.89 $\downarrow$& 4.08 $\downarrow$& 7.71 $\downarrow$& 4.43 $\downarrow$& 3.42 $\downarrow$& 8.25 $\downarrow$\\ \hline
        right eyebrow & 2.45 $\uparrow$& 1.85 $\uparrow$& 1.99 $\uparrow$& 2.71 $\uparrow$& 2.09 $\uparrow$& 2.47 $\uparrow$& 2.32 $\uparrow$& 2.05 $\uparrow$& 2.66 $\uparrow$\\ \hline
        left eyebrow & 2.51 $\uparrow$& 1.71 $\uparrow$& 1.93 $\uparrow$& 2.30 $\uparrow$& 2.38 $\uparrow$& 2.27 $\uparrow$& 2.09 $\uparrow$& 1.93 $\uparrow$& 2.55 $\uparrow$\\ \hline
        upper lip & 1.99 $\uparrow$& 1.57 $\uparrow$& 1.51 $\uparrow$& 1.94 $\uparrow$& 1.81 $\uparrow$& 1.46 $\uparrow$& 1.31 $\uparrow$& 1.83 $\uparrow$& 0.97 $\downarrow$\\ \hline
        right eye & 2.26 $\uparrow$& 1.78 $\uparrow$& 1.52 $\uparrow$& 2.27 $\uparrow$& 1.78 $\uparrow$& 1.26 $\uparrow$& 0.75 $\uparrow$& 0.73 $\uparrow$& 1.32 $\uparrow$\\ \hline
        left eye & 2.20 $\uparrow$& 1.78 $\uparrow$& 1.47 $\uparrow$& 2.11 $\uparrow$& 1.87 $\uparrow$& 1.23 $\uparrow$& 0.67 $\uparrow$& 0.66 $\uparrow$& 1.31 $\uparrow$\\ \hline
        lower lip & 1.69 $\uparrow$& 2.16 $\uparrow$& 1.35 $\downarrow$& 1.55 $\downarrow$& 1.51 $\downarrow$& 1.23 $\downarrow$& 1.08 $\downarrow$& 1.34 $\downarrow$& 0.77 $\downarrow$\\ \hline
        eyeglasses & 1.25 $\uparrow$& 1.19 $\uparrow$& 1.24 $\uparrow$& 1.29 $\uparrow$& 1.23 $\uparrow$& 1.24 $\uparrow$& 0.75 $\downarrow$& 0.73 $\downarrow$& 1.40 $\uparrow$\\ \hline
        clothes & 0.80 $\downarrow$& 0.88 $\downarrow$& 0.66 $\downarrow$& 1.16 $\downarrow$& 0.65 $\downarrow$& 0.83 $\downarrow$& 0.65 $\downarrow$& 0.81 $\downarrow$& 0.72 $\downarrow$\\ \hline
        inner mouth & 0.82 $\uparrow$& 1.00 $\uparrow$& 0.88 $\uparrow$& 0.91 $\uparrow$& 0.89 $\uparrow$& 0.76 $\uparrow$& 0.56 $\downarrow$& 0.74 $\uparrow$& 0.65 $\uparrow$\\ \hline
        neck & 1.05 $\downarrow$& 0.71 $\downarrow$& 0.59 $\downarrow$& 1.53 $\downarrow$& 0.41 $\downarrow$& 0.43 $\downarrow$& 0.59 $\downarrow$& 1.02 $\downarrow$& 0.16 $\downarrow$\\ \hline
        hat & 0.30 $\downarrow$& 0.54 $\downarrow$& 0.59 $\downarrow$& 0.59 $\downarrow$& 0.48 $\downarrow$& 0.75 $\downarrow$& 0.30 $\downarrow$& 0.35 $\downarrow$& 1.12 $\downarrow$\\ \hline
        right ear & 0.13 $\downarrow$& 0.16 $\downarrow$& 0.12 $\downarrow$& 0.18 $\downarrow$& 0.14 $\downarrow$& 0.19 $\downarrow$& 0.05 $\downarrow$& 0.07 $\downarrow$& 0.13 $\downarrow$\\ \hline
        left ear & 0.10 $\downarrow$& 0.13 $\downarrow$& 0.09 $\downarrow$& 0.15 $\downarrow$& 0.13 $\downarrow$& 0.15 $\downarrow$& 0.04 $\downarrow$& 0.07 $\downarrow$& 0.12 $\downarrow$\\ \hline
        earrings & 0.02 $\downarrow$& 0.03 $\leftrightarrow$& 0.02 $\downarrow$& 0.05 $\downarrow$& 0.03 $\downarrow$& 0.04 $\downarrow$& 0.02 $\downarrow$& 0.02 $\downarrow$& 0.02 $\downarrow$\\ \hline
        necklace & 0.00 $\downarrow$& 0.00 $\leftrightarrow$& 0.00 $\leftrightarrow$& 0.00 $\downarrow$& 0.00 $\leftrightarrow$& 0.00 $\leftrightarrow$& 0.00 $\downarrow$& 0.00 $\downarrow$& 0.00 $\leftrightarrow$\\ \hline
    \end{tabular}
\end{table*}

\appendix
\vspace{2em} To ensure transparency and replicability of the study, our research artifacts are made publicly available for open science: 
\begin{center}
\url{https://anonymous.4open.science/r/LEAM-C01B/README.md}
\end{center}

\section{[RQ2] Focus areas of recognition models - numerical data}
\label{Appendix_RQ2}

As presented in Table \ref{tab:all_features}, models focus on almost the entirety of the facial area for positive detection. For all of them the most relevant class, due to its general area, is \textit{face}, followed by \textit{hair}, \textit{background}, and \textit{nose}. Each of the remaining classes represents on its own less than 3\% of the total positively activated area, yet many of them span across a relatively small percentage of the image.
Usage of different versions of the same model does not guarantee similar results as one can observe fluctuations in IResNet models' attention based on their backbone and loss function used.

While looking at the averages for the top 1\% of activated pixels, presented in Table \ref{tab:all_features_1_percent}, we could observe a sharp increase in the focus on the areas that lie in the center of the face -  nose, eyes and eyebrows. The most notable improvement was observed for the class \textit{nose}. It accounts for more than 20\% of the most activated area for the majority of the models. Its relevance rose at the expense of \textit{face} and \textit{background}.

Taking both tables into consideration together, one can observe that the values obtained for symmetric face parts are quite close to one another. The pattern is more visible across eyes or ears and less across eyebrows or lips. Considering accessories, \textit{earrings} and \textit{necklace} pixels are not strongly represented across the most relevant areas and are too infrequent to have a significant impact on prediction. \textit{Hat} pixels, even though higher in quantity, also do not seem to be very highly activated. Conversely, \textit{eyeglasses} have a much stronger effect on the prediction.

\section{[RQ5] Impact of the most relevant pixels
thresholds on key facial landmarks}
\label{Appendix_RQ5}

We have computed the distributions of the most relevant pixels at various thresholds, presented in
Tables~\ref{tab:thresholds_af18}-\ref{tab:thresholds_vgg16} to assess the importance of facial landmarks. It can be observed that the classes such as \textit{face}, \textit{background} and \textit{hair} are the most prevalent at high threshold values, while \textit{nose}, eyebrows and eyes (with exception of IR\_CASIA and IR\_VGG models) become more dominant at low thresholds. The change is very apparent in the case of class \textit{nose}, which usually accounts for around 10\% of total activation, but it reaches 20-30\% at the lowest thresholds, depending on the assessed model. This finding suggests that those key, central landmarks have a higher impact on the recognition process and should be considered for inconspicuous privacy-preserving attacks. Nonetheless, there is still very much attention, always above 40\% of all pixels, paid to the general \textit{face} area, suggesting that perturbations should not be limited just to the scope of distinct landmarks, but extend to also affect areas such as cheeks or forehead.

Further, it can be clearly observed that the values obtained for vertically symmetric facial classes -- eyes, ears and eyebrows -- achieved comparable results for left and right classes, while the same cannot be written about horizontally symmetric \textit{lower lip} and \textit{upper lip}. Recognition models clearly pay more attention to the \textit{upper lip}, which is positioned closer to the center of the face than its counterpart.

\input{BigFigureOrTable/table_appendix_rq5}

\section{[RQ6] Transferability of individual-specific patterns across different models - numerical data}
\label{Appendix_RQ6}

Table \ref{tab:Similarity_drop_model}  presents cosine similarity decrease values when all datasets were perturbed with proof-of-concept black pixel occlusions generated for IR\_CASIA. It can be observed that LEAM-guided occlusions are particularly effective for low thresholds, which could be considered inconspicuous. As there is a low quantity of high-importance pixels, the random approach is statistically unlikely to select them and consequently unable to provide a significant similarity drop. At the same time, the targeted LEAM-guided approach can find key areas and consistently transfer them across all the tested models. The similarity decrease at the same occlusion percentages tend to be the highest for IR\_CASIA and IR\_VGG, which share a backbone. Nonetheless, the efficiency loss, which is expected in the case of the transferability of an attack, is small enough for other models to still consider all the attacks as successful.
\begin{table*}[t]
    \centering
    \small
    \caption{Cosine similarity drop with occlusion per model}
    \label{tab:Similarity_drop_model}
    \begin{tabular}{|c|c|c|c|c|c|c|c|c|c|c|}
    \hline
        Occlusion type & Occlusion (\%) & AF\_R18 & AF\_R50 & AF\_R100 & CF\_R18 & CF\_R50 & CF\_R100 & IR\_CASIA & IR\_VGG & VGG\_16 \\ \hline
        LEAM & 0.01 & 0.0036 & 0.0039 & 0.0042 & 0.0037 & 0.0040 & 0.0032 & 0.0063 & 0.0058 & 0.0036 \\ \hline
        ~ & 0.05 & 0.0147 & 0.0141 & 0.0130 & 0.0163 & 0.0145 & 0.0105 & 0.0206 & 0.0236 & 0.0181 \\ \hline
        ~ & 0.10 & 0.0263 & 0.0241 & 0.0217 & 0.0292 & 0.0251 & 0.0181 & 0.0367 & 0.0425 & 0.0297 \\ \hline
        ~ & 0.25 & 0.0496 & 0.0466 & 0.0405 & 0.0556 & 0.0452 & 0.0351 & 0.0703 & 0.0761 & 0.0495 \\ \hline
        ~ & 0.50 & 0.0757 & 0.0707 & 0.0621 & 0.0842 & 0.0673 & 0.0544 & 0.1099 & 0.1091 & 0.0711 \\ \hline
        ~ & 0.75 & 0.0952 & 0.0871 & 0.0784 & 0.1054 & 0.0831 & 0.0701 & 0.1403 & 0.1354 & 0.0849 \\ \hline
        ~ & 1.00 & 0.1100 & 0.1032 & 0.0934 & 0.1223 & 0.0991 & 0.0851 & 0.1655 & 0.1579 & 0.0969 \\ \hline
        ~ & 1.25 & 0.1227 & 0.1167 & 0.1082 & 0.1368 & 0.1127 & 0.0991 & 0.1843 & 0.1761 & 0.1074 \\ \hline
        ~ & 1.50 & 0.1343 & 0.1300 & 0.1227 & 0.1494 & 0.1267 & 0.1127 & 0.2006 & 0.1943 & 0.1158 \\ \hline
        ~ & 2.00 & 0.1531 & 0.1526 & 0.1494 & 0.1697 & 0.1510 & 0.1375 & 0.2284 & 0.2285 & 0.1315 \\ \hline
        ~ & 5.00 & 0.2225 & 0.2648 & 0.2804 & 0.2432 & 0.2562 & 0.2651 & 0.3170 & 0.3444 & 0.1827 \\ \hline
        ~ & 10.00 & 0.2725 & 0.3499 & 0.3866 & 0.2928 & 0.3405 & 0.3770 & 0.3752 & 0.4226 & 0.2166 \\ \hline \hline
        Random & 0.01 & 0.0004 & 0.0003 & 0.0006 & 0.0005 & 0.0002 & 0.0003 & 0.0003 & 0.0004 & 0.0000 \\ \hline
        ~ & 0.05 & 0.0014 & 0.0020 & 0.0021 & 0.0017 & 0.0012 & 0.0020 & 0.0022 & 0.0010 & 0.0004 \\ \hline
        ~ & 0.10 & 0.0030 & 0.0034 & 0.0040 & 0.0032 & 0.0028 & 0.0031 & 0.0040 & 0.0031 & 0.0009 \\ \hline
        ~ & 0.25 & 0.0071 & 0.0088 & 0.0103 & 0.0061 & 0.0061 & 0.0090 & 0.0081 & 0.0089 & 0.0026 \\ \hline
        ~ & 0.50 & 0.0138 & 0.0164 & 0.0189 & 0.0149 & 0.0155 & 0.0170 & 0.0180 & 0.0162 & 0.0066 \\ \hline
        ~ & 0.75 & 0.0226 & 0.0235 & 0.0280 & 0.0201 & 0.0218 & 0.0228 & 0.0267 & 0.0238 & 0.0113 \\ \hline
        ~ & 1.00 & 0.0287 & 0.0292 & 0.0342 & 0.0268 & 0.0273 & 0.0293 & 0.0355 & 0.0309 & 0.0157 \\ \hline
        ~ & 1.25 & 0.0336 & 0.0345 & 0.0395 & 0.0334 & 0.0339 & 0.0360 & 0.0425 & 0.0366 & 0.0186 \\ \hline
        ~ & 1.50 & 0.0391 & 0.0405 & 0.0475 & 0.0388 & 0.0408 & 0.0431 & 0.0508 & 0.0432 & 0.0223 \\ \hline
        ~ & 2.00 & 0.0497 & 0.0512 & 0.0587 & 0.0479 & 0.0495 & 0.0519 & 0.0678 & 0.0546 & 0.0287 \\ \hline
        ~ & 5.00 & 0.0990 & 0.1004 & 0.1115 & 0.0988 & 0.1005 & 0.1053 & 0.1286 & 0.1093 & 0.0611 \\ \hline
        ~ & 10.00 & 0.1531 & 0.1668 & 0.1748 & 0.1599 & 0.1538 & 0.1676 & 0.2224 & 0.1964 & 0.1113 \\ \hline
    \end{tabular}
\end{table*}

\section{[RQ7] Gender, ethnicity and age variation ablation study - numerical data}
\label{Appendix_RQ7}

\begin{table*}[t]
    \centering
    \small
    \caption{Cosine similarity drop with occlusion per model - Female}
    \label{tab:Similarity_drop_model_female}
    \begin{tabular}{|c|c|c|c|c|c|c|c|c|c|c|}
    \hline 
    Occlusion type & Occlusion (\%) & AF\_R18 & AF\_R50 & AF\_R100 & CF\_R18 & CF\_R50 & CF\_R100 & IR\_CASIA & IR\_VGG & VGG\_16 \\ \hline 
    LEAM & 0.01 & 0.0039 & 0.0047 & 0.0051 & 0.0045 & 0.0045 & 0.0035 & 0.0065 & 0.0081 & 0.0041 \\ \hline
        ~ & 0.05 & 0.0161 & 0.0159 & 0.0154 & 0.0189 & 0.0159 & 0.0111 & 0.0221 & 0.0277 & 0.0211 \\ \hline
        ~ & 0.10 & 0.0279 & 0.0272 & 0.0240 & 0.0326 & 0.0276 & 0.0191 & 0.0393 & 0.0492 & 0.0359 \\ \hline
        ~ & 0.25 & 0.0526 & 0.0499 & 0.0444 & 0.0607 & 0.0477 & 0.0370 & 0.0771 & 0.0881 & 0.0606 \\ \hline
        ~ & 0.50 & 0.0802 & 0.0763 & 0.0680 & 0.0921 & 0.0735 & 0.0596 & 0.1187 & 0.1230 & 0.0864 \\ \hline
        ~ & 0.75 & 0.0998 & 0.0934 & 0.0849 & 0.1134 & 0.0910 & 0.0768 & 0.1489 & 0.1497 & 0.1017 \\ \hline
        ~ & 1.00 & 0.1144 & 0.1099 & 0.1004 & 0.1298 & 0.1074 & 0.0928 & 0.1769 & 0.1759 & 0.1159 \\ \hline
        ~ & 1.25 & 0.1269 & 0.1237 & 0.1157 & 0.1442 & 0.1210 & 0.1079 & 0.1935 & 0.1941 & 0.1269 \\ \hline
        ~ & 1.50 & 0.1380 & 0.1374 & 0.1305 & 0.1574 & 0.1357 & 0.1209 & 0.2113 & 0.2124 & 0.1369 \\ \hline
        ~ & 2.00 & 0.1567 & 0.1620 & 0.1582 & 0.1786 & 0.1620 & 0.1484 & 0.2403 & 0.2477 & 0.1552 \\ \hline
        ~ & 5.00 & 0.2230 & 0.2838 & 0.2960 & 0.2488 & 0.2705 & 0.2877 & 0.3264 & 0.3616 & 0.2110 \\ \hline
        ~ & 10.00 & 0.2739 & 0.3684 & 0.3975 & 0.2970 & 0.3601 & 0.4002 & 0.3831 & 0.4294 & 0.2407 \\ \hline \hline
        Random & 0.01 & 0.0006 & 0.0004 & 0.0007 & 0.0005 & 0.0004 & 0.0007 & 0.0004 & 0.0001 & 0.0001 \\ \hline
        ~ & 0.05 & 0.0017 & 0.0019 & 0.0025 & 0.0016 & 0.0010 & 0.0022 & 0.0028 & 0.0009 & 0.0000 \\ \hline
        ~ & 0.10 & 0.0032 & 0.0034 & 0.0049 & 0.0039 & 0.0031 & 0.0037 & 0.0060 & 0.0043 & 0.0014 \\ \hline
        ~ & 0.25 & 0.0069 & 0.0100 & 0.0120 & 0.0069 & 0.0059 & 0.0097 & 0.0117 & 0.0112 & 0.0034 \\ \hline
        ~ & 0.50 & 0.0154 & 0.0205 & 0.0223 & 0.0177 & 0.0188 & 0.0194 & 0.0233 & 0.0198 & 0.0078 \\ \hline
        ~ & 0.75 & 0.0238 & 0.0255 & 0.0314 & 0.0233 & 0.0251 & 0.0253 & 0.0339 & 0.0262 & 0.0140 \\ \hline
        ~ & 1.00 & 0.0300 & 0.0310 & 0.0392 & 0.0303 & 0.0305 & 0.0330 & 0.0472 & 0.0391 & 0.0197 \\ \hline
        ~ & 1.25 & 0.0364 & 0.0385 & 0.0444 & 0.0382 & 0.0389 & 0.0400 & 0.0539 & 0.0447 & 0.0221 \\ \hline
        ~ & 1.50 & 0.0435 & 0.0452 & 0.0543 & 0.0443 & 0.0458 & 0.0487 & 0.0615 & 0.0509 & 0.0248 \\ \hline
        ~ & 2.00 & 0.0544 & 0.0572 & 0.0648 & 0.0542 & 0.0550 & 0.0583 & 0.0792 & 0.0637 & 0.0345 \\ \hline
        ~ & 5.00 & 0.1028 & 0.1067 & 0.1195 & 0.1092 & 0.1084 & 0.1140 & 0.1466 & 0.1238 & 0.0682 \\ \hline
        ~ & 10.00 & 0.1576 & 0.1751 & 0.1923 & 0.1728 & 0.1675 & 0.1835 & 0.2455 & 0.2232 & 0.1248 \\ \hline
    \end{tabular}
\end{table*}

\begin{table*}[t]
    \centering
    \small
    \caption{Cosine similarity drop with occlusion per model - Male}
    \label{tab:Similarity_drop_model_male}
    \begin{tabular}{|c|c|c|c|c|c|c|c|c|c|c|}
    \hline Occlusion type & Occlusion (\%) & AF\_R18 & AF\_R50 & AF\_R100 & CF\_R18 & CF\_R50 & CF\_R100 & IR\_CASIA & IR\_VGG & VGG\_16 \\ \hline
        LEAM & 0.01 & 0.0028 & 0.0031 & 0.0033 & 0.0023 & 0.0032 & 0.0026 & 0.0058 & 0.0036 & 0.0026 \\ \hline
        ~ & 0.05 & 0.0131 & 0.0121 & 0.0103 & 0.0128 & 0.0124 & 0.0090 & 0.0189 & 0.0199 & 0.0150 \\ \hline
        ~ & 0.10 & 0.0234 & 0.0205 & 0.0187 & 0.0239 & 0.0220 & 0.0165 & 0.0339 & 0.0359 & 0.0237 \\ \hline
        ~ & 0.25 & 0.0442 & 0.0426 & 0.0367 & 0.0475 & 0.0418 & 0.0330 & 0.0637 & 0.0646 & 0.0396 \\ \hline
        ~ & 0.50 & 0.0678 & 0.0645 & 0.0563 & 0.0733 & 0.0598 & 0.0493 & 0.1009 & 0.0949 & 0.0574 \\ \hline
        ~ & 0.75 & 0.0874 & 0.0811 & 0.0712 & 0.0947 & 0.0749 & 0.0634 & 0.1309 & 0.1207 & 0.0700 \\ \hline
        ~ & 1.00 & 0.1021 & 0.0958 & 0.0849 & 0.1108 & 0.0892 & 0.0769 & 0.1537 & 0.1414 & 0.0797 \\ \hline
        ~ & 1.25 & 0.1144 & 0.1087 & 0.0983 & 0.1255 & 0.1026 & 0.0894 & 0.1742 & 0.1599 & 0.0895 \\ \hline
        ~ & 1.50 & 0.1260 & 0.1209 & 0.1120 & 0.1377 & 0.1155 & 0.1032 & 0.1896 & 0.1780 & 0.0962 \\ \hline
        ~ & 2.00 & 0.1447 & 0.1423 & 0.1369 & 0.1564 & 0.1375 & 0.1267 & 0.2163 & 0.2103 & 0.1100 \\ \hline
        ~ & 5.00 & 0.2141 & 0.2453 & 0.2607 & 0.2293 & 0.2377 & 0.2433 & 0.3076 & 0.3267 & 0.1560 \\ \hline
        ~ & 10.00 & 0.2624 & 0.3292 & 0.3695 & 0.2800 & 0.3181 & 0.3546 & 0.3689 & 0.4114 & 0.1929 \\ \hline \hline
        Random & 0.01 & 0.0004 & 0.0003 & 0.0006 & 0.0004 & 0.0000 & -0.0001 & 0.0004 & 0.0005 & 0.0000 \\ \hline
        ~ & 0.05 & 0.0009 & 0.0021 & 0.0015 & 0.0014 & 0.0015 & 0.0017 & 0.0020 & 0.0016 & 0.0010 \\ \hline
        ~ & 0.10 & 0.0023 & 0.0029 & 0.0026 & 0.0020 & 0.0024 & 0.0024 & 0.0022 & 0.0020 & 0.0004 \\ \hline
        ~ & 0.25 & 0.0068 & 0.0070 & 0.0081 & 0.0055 & 0.0059 & 0.0079 & 0.0057 & 0.0074 & 0.0019 \\ \hline
        ~ & 0.50 & 0.0115 & 0.0125 & 0.0155 & 0.0119 & 0.0126 & 0.0144 & 0.0141 & 0.0126 & 0.0055 \\ \hline
        ~ & 0.75 & 0.0203 & 0.0204 & 0.0240 & 0.0159 & 0.0189 & 0.0196 & 0.0199 & 0.0199 & 0.0076 \\ \hline
        ~ & 1.00 & 0.0257 & 0.0259 & 0.0279 & 0.0221 & 0.0231 & 0.0254 & 0.0249 & 0.0238 & 0.0124 \\ \hline
        ~ & 1.25 & 0.0312 & 0.0297 & 0.0352 & 0.0288 & 0.0289 & 0.0315 & 0.0311 & 0.0305 & 0.0142 \\ \hline
        ~ & 1.50 & 0.0344 & 0.0346 & 0.0393 & 0.0329 & 0.0347 & 0.0366 & 0.0402 & 0.0353 & 0.0192 \\ \hline
        ~ & 2.00 & 0.0443 & 0.0458 & 0.0516 & 0.0414 & 0.0438 & 0.0459 & 0.0553 & 0.0447 & 0.0227 \\ \hline
        ~ & 5.00 & 0.0903 & 0.0908 & 0.0999 & 0.0851 & 0.0904 & 0.0944 & 0.1098 & 0.0958 & 0.0530 \\ \hline
        ~ & 10.00 & 0.1441 & 0.1527 & 0.1542 & 0.1428 & 0.1373 & 0.1494 & 0.1976 & 0.1659 & 0.0961 \\ \hline
    \end{tabular}
\end{table*}

Our occlusion analysis results were split by gender and presented in Tables \ref{tab:Similarity_drop_model_female} and \ref{tab:Similarity_drop_model_male}. 
One can notice that for almost all tested occlusion percentages, types, and used models, the similarity drop is greater among female pictures. 
Relatively high disparities can be observed for VGG\_16, where, for instance, for 1\% occlusion, one can observe a 0.0797 similarity decrease for males vs. 0.1159 for females.

\begin{table*}[!ht]
    \centering
    \small
    \caption{Cosine similarity vs. age variation per model - 1\% LEAM-guided occlusion}
    \label{tab:Similarity_drop_model_age}
    \begin{tabular}{|l|c|c|c|c|c|c|c|c|c|c|c|c|c|c|c|c|c|}
        \hline
        Age difference & 1 & 2 & 3 & 4 & 5 & 6 & 7 & 8 & 9 & 10 & 11 & 12 & 13 & 14 \\ \hline
        AF\_18 & 0.3827 & 0.3326 & 0.3167 & 0.3153 & 0.2858 & 0.2573 & 0.2904 & 0.2257 & 0.2253 & 0.2086 & 0.2084 & 0.2133 & 0.1969 & 0.1765 \\ \hline
        AF\_50 & 0.3931 & 0.3273 & 0.3054 & 0.3198 & 0.3028 & 0.2405 & 0.2735 & 0.2309 & 0.2076 & 0.2133 & 0.2148 & 0.2123 & 0.1960 & 0.1705 \\ \hline
        AF\_100 & 0.4652 & 0.3656 & 0.3300 & 0.3624 & 0.3522 & 0.2758 & 0.2963 & 0.2786 & 0.2229 & 0.2440 & 0.2329 & 0.2175 & 0.2148 & 0.1787 \\ \hline
        CF\_18 & 0.3345 & 0.2948 & 0.2621 & 0.2726 & 0.2571 & 0.2127 & 0.2501 & 0.1926 & 0.1879 & 0.1861 & 0.1737 & 0.2018 & 0.1613 & 0.1251 \\ \hline
        CF\_100 & 0.4611 & 0.3711 & 0.3377 & 0.3658 & 0.3495 & 0.2883 & 0.2955 & 0.2560 & 0.2267 & 0.2529 & 0.2222 & 0.2323 & 0.2122 & 0.1823 \\ \hline
        CF\_50 & 0.4001 & 0.3301 & 0.3162 & 0.3207 & 0.2987 & 0.2575 & 0.2912 & 0.2373 & 0.1971 & 0.2200 & 0.2074 & 0.2132 & 0.1826 & 0.1569 \\ \hline
        IR\_CASIA & 0.4615 & 0.4115 & 0.3407 & 0.3645 & 0.3454 & 0.3328 & 0.3447 & 0.2925 & 0.2603 & 0.2847 & 0.2465 & 0.2791 & 0.2415 & 0.1879 \\ \hline
        IR\_VGG & 0.4858 & 0.4283 & 0.3869 & 0.3803 & 0.4096 & 0.3704 & 0.3849 & 0.3335 & 0.2882 & 0.3179 & 0.2850 & 0.2996 & 0.2734 & 0.2511 \\ \hline
        VGG\_16 & 0.5672 & 0.5334 & 0.4979 & 0.5296 & 0.5368 & 0.5030 & 0.5227 & 0.4991 & 0.4548 & 0.4808 & 0.4794 & 0.4927 & 0.4674 & 0.4649 \\ \hline
        Age difference & 15 & 16 & 17 & 18 & 19 & 20 & 21 & 22 & 23 & 24 & 25 & 26 & 27 & 28 \\ \hline
        AF\_18 & 0.1650 & 0.1331 & 0.1642 & 0.1469 & 0.1433 & 0.1653 & 0.1505 & 0.1899 & 0.1587 & 0.1748 & 0.1190 & 0.1548 & 0.2046 & 0.0971 \\ \hline
        AF\_50 & 0.1332 & 0.1389 & 0.1399 & 0.1371 & 0.1306 & 0.1691 & 0.1418 & 0.1548 & 0.1288 & 0.1801 & 0.1219 & 0.1121 & 0.1819 & 0.0810 \\ \hline
        AF\_100 & 0.1287 & 0.1405 & 0.1568 & 0.1353 & 0.1288 & 0.1715 & 0.1761 & 0.1640 & 0.1290 & 0.1639 & 0.1188 & 0.1436 & 0.1687 & 0.1116 \\ \hline
        CF\_18 & 0.1225 & 0.1213 & 0.1132 & 0.1231 & 0.1274 & 0.1392 & 0.0887 & 0.1654 & 0.1251 & 0.1018 & 0.1004 & 0.0903 & 0.1748 & 0.0528 \\ \hline
        CF\_100 & 0.1545 & 0.1434 & 0.1662 & 0.1442 & 0.1344 & 0.1704 & 0.1682 & 0.1414 & 0.1356 & 0.1366 & 0.1193 & 0.1762 & 0.1656 & 0.0869 \\ \hline
        CF\_50 & 0.1316 & 0.1209 & 0.1454 & 0.1206 & 0.1336 & 0.1599 & 0.1343 & 0.1365 & 0.1258 & 0.1215 & 0.0907 & 0.1227 & 0.1670 & 0.0536 \\ \hline
        IR\_CASIA & 0.2239 & 0.1600 & 0.1747 & 0.1839 & 0.1623 & 0.1808 & 0.1597 & 0.1755 & 0.2104 & 0.2161 & 0.1220 & 0.1647 & 0.1785 & 0.1172 \\ \hline
        IR\_VGG & 0.2257 & 0.2041 & 0.2174 & 0.2024 & 0.2192 & 0.2501 & 0.2145 & 0.2342 & 0.2587 & 0.2405 & 0.1178 & 0.2036 & 0.3000 & 0.1150 \\ \hline
        VGG\_16 & 0.4546 & 0.4447 & 0.4456 & 0.4670 & 0.4309 & 0.4876 & 0.4481 & 0.4918 & 0.4596 & 0.4741 & 0.4193 & 0.3613 & 0.4992 & 0.3746 \\ \hline
        Age difference & 29 & 30 & 31 & 32 & 33 & 34 & 35 & 36 & 37 & 38 & 39 & 40 & 41 & 42 \\ \hline
        AF\_18 & 0.1438 & 0.1406 & 0.2794 & 0.2052 & 0.1012 & 0.1171 & 0.2293 & 0.1633 & 0.2279 & 0.1778 & 0.1133 & 0.1766 & 0.0274 & 0.1617 \\ \hline
        AF\_50 & 0.1638 & 0.0765 & 0.2687 & 0.1904 & 0.0987 & 0.1272 & 0.2041 & 0.1903 & 0.2191 & 0.1170 & 0.0881 & 0.1525 & 0.0473 & 0.1613 \\ \hline
        AF\_100 & 0.2035 & 0.1232 & 0.2953 & 0.1951 & 0.1211 & 0.1395 & 0.2004 & 0.1774 & 0.1678 & 0.1472 & 0.1325 & 0.1894 & 0.0345 & 0.1294 \\ \hline
        CF\_18 & 0.1058 & 0.1042 & 0.2439 & 0.1559 & 0.0949 & 0.1094 & 0.1376 & 0.0909 & 0.1830 & 0.0917 & 0.1167 & 0.1078 & 0.0178 & 0.1189 \\ \hline
        CF\_100 & 0.1771 & 0.1163 & 0.2884 & 0.1876 & 0.0900 & 0.1165 & 0.1839 & 0.1921 & 0.2006 & 0.1441 & 0.0999 & 0.1349 & 0.0557 & 0.1188 \\ \hline
        CF\_50 & 0.1718 & 0.0918 & 0.2790 & 0.1755 & 0.0659 & 0.1042 & 0.1565 & 0.1778 & 0.1643 & 0.1105 & 0.0615 & 0.1224 & 0.0163 & 0.1339 \\ \hline
        IR\_CASIA & 0.1294 & 0.1355 & 0.2617 & 0.2404 & 0.1134 & 0.1578 & 0.2203 & 0.0935 & 0.1401 & 0.1532 & 0.1081 & 0.1053 & -0.0300 & 0.1481 \\ \hline
        IR\_VGG & 0.1927 & 0.1892 & 0.3087 & 0.2890 & 0.0896 & 0.1641 & 0.2807 & -0.0262 & 0.1688 & 0.1892 & 0.2027 & 0.1898 & 0.0809 & 0.2067 \\ \hline
        VGG\_16 & 0.4320 & 0.4464 & 0.5430 & 0.4891 & 0.4079 & 0.4502 & 0.4145 & 0.4221 & 0.4410 & 0.3917 & 0.4209 & 0.4789 & 0.4324 & 0.4781 \\ \hline
    \end{tabular}
\end{table*}

To showcase non-uniform nature of cosine similarity decrease across assessed modes for age variance study, model-specific results for 1\% LEAM-guided occlusions were provided in Table \ref{tab:Similarity_drop_model_age}. It can be observed that the age variation resistance is a highly model-dependent characteristic. Therefore, any robust privacy-preserving attack should be tested against models that can withstand temporal changes in one's appearance.

%% file: BigFigureOrTable/table_appendix_rq5.tex
\begin{table*}[t]
    \centering
    \small
    \caption{Percentage distribution of facial areas positively activated for AF\_R18 per activation threshold}
    \label{tab:thresholds_af18}
    \begin{tabular}{|l|c|c|c|c|c|c|c|c|c|c|c|c|c|c|c|}
    \hline
        Threshold (\%) & 0.01 & 0.05 & 0.10 & 0.25 & 0.50 & 0.75 & 1.00 & 1.25 & 1.50 & 1.75 & 2.00 & 3.00 & 5.00 & 10.00 & 100.00 \\ \hline
        background & 2.37 & 2.70 & 3.03 & 3.58 & 4.14 & 4.47 & 3.78 & 4.89 & 5.04 & 5.19 & 5.32 & 5.76 & 6.42 & 7.44 & 7.47 \\ \hline
        clothes & 0.51 & 0.57 & 0.62 & 0.71 & 0.79 & 0.85 & 0.80 & 0.93 & 0.97 & 1.00 & 1.03 & 1.13 & 1.27 & 1.45 & 1.46 \\ \hline
        earrings & 0.01 & 0.01 & 0.01 & 0.01 & 0.02 & 0.02 & 0.02 & 0.02 & 0.03 & 0.03 & 0.03 & 0.03 & 0.04 & 0.05 & 0.05 \\ \hline
        eyeglasses & 1.45 & 1.41 & 1.38 & 1.36 & 1.35 & 1.34 & 1.25 & 1.34 & 1.33 & 1.33 & 1.32 & 1.30 & 1.25 & 1.17 & 0.99 \\ \hline
        face & 44.47 & 46.52 & 47.89 & 49.77 & 51.53 & 52.61 & 55.00 & 54.01 & 54.48 & 54.87 & 55.18 & 56.01 & 56.77 & 56.98 & 58.32 \\ \hline
        hair & 1.56 & 1.80 & 2.06 & 2.50 & 2.98 & 3.32 & 3.15 & 3.84 & 4.05 & 4.26 & 4.44 & 5.07 & 6.04 & 7.60 & 8.42 \\ \hline
        hat & 0.17 & 0.19 & 0.20 & 0.23 & 0.26 & 0.28 & 0.30 & 0.32 & 0.33 & 0.35 & 0.36 & 0.40 & 0.48 & 0.59 & 0.70 \\ \hline
        inner mouth & 0.69 & 0.70 & 0.73 & 0.78 & 0.82 & 0.84 & 0.82 & 0.86 & 0.86 & 0.87 & 0.87 & 0.86 & 0.84 & 0.79 & 0.68 \\ \hline
        left ear & 0.05 & 0.05 & 0.06 & 0.08 & 0.09 & 0.10 & 0.10 & 0.12 & 0.13 & 0.13 & 0.14 & 0.16 & 0.19 & 0.24 & 0.26 \\ \hline
        left eye & 4.96 & 4.46 & 4.03 & 3.49 & 3.02 & 2.76 & 2.20 & 2.45 & 2.35 & 2.26 & 2.19 & 1.98 & 1.74 & 1.51 & 0.92 \\ \hline
        left eyebrow & 2.47 & 2.45 & 2.46 & 2.47 & 2.46 & 2.42 & 2.51 & 2.36 & 2.33 & 2.30 & 2.27 & 2.17 & 2.02 & 1.82 & 1.75 \\ \hline
        lower lip & 1.14 & 1.22 & 1.31 & 1.45 & 1.60 & 1.69 & 1.69 & 1.81 & 1.85 & 1.88 & 1.90 & 1.95 & 1.96 & 1.88 & 1.63 \\ \hline
        neck & 0.64 & 0.71 & 0.79 & 0.90 & 1.00 & 1.08 & 1.05 & 1.20 & 1.25 & 1.29 & 1.33 & 1.48 & 1.69 & 1.99 & 2.10 \\ \hline
        necklace & 0.00 & 0.00 & 0.00 & 0.00 & 0.00 & 0.00 & 0.00 & 0.00 & 0.00 & 0.00 & 0.00 & 0.00 & 0.00 & 0.01 & 0.01 \\ \hline
        nose & 30.29 & 28.45 & 26.99 & 24.72 & 22.41 & 20.95 & 20.49 & 18.94 & 18.22 & 17.58 & 17.05 & 15.43 & 13.44 & 11.21 & 10.76 \\ \hline
        right ear & 0.06 & 0.08 & 0.10 & 0.12 & 0.14 & 0.15 & 0.13 & 0.18 & 0.18 & 0.19 & 0.20 & 0.22 & 0.26 & 0.31 & 0.30 \\ \hline
        right eyebrow & 2.36 & 2.37 & 2.39 & 2.39 & 2.38 & 2.35 & 2.45 & 2.30 & 2.27 & 2.25 & 2.22 & 2.13 & 2.00 & 1.80 & 1.73 \\ \hline
        right eye & 5.08 & 4.56 & 4.15 & 3.59 & 3.11 & 2.84 & 2.26 & 2.52 & 2.42 & 2.33 & 2.25 & 2.04 & 1.79 & 1.55 & 0.94 \\ \hline
        upper lip & 1.73 & 1.75 & 1.79 & 1.85 & 1.90 & 1.92 & 1.99 & 1.92 & 1.92 & 1.91 & 1.91 & 1.87 & 1.78 & 1.62 & 1.53 \\ \hline
    \end{tabular}
\end{table*}

\begin{table*}[t]
    \centering
    \small
    \caption{Percentage distribution of facial areas positively activated for AF\_R50 per activation threshold}
    \label{tab:thresholds_af50}
    \begin{tabular}{|l|c|c|c|c|c|c|c|c|c|c|c|c|c|c|c|}
    \hline        
        Threshold (\%) & 0.01 & 0.05 & 0.10 & 0.25 & 0.50 & 0.75 & 1.00 & 1.25 & 1.50 & 1.75 & 2.00 & 3.00 & 5.00 & 10.00 & 100.00 \\ \hline
        background & 2.93 & 3.18 & 3.41 & 3.85 & 4.42 & 4.85 & 5.49 & 5.55 & 5.82 & 6.06 & 6.28 & 6.96 & 7.84 & 8.88 & 8.74 \\ \hline
        clothes & 0.47 & 0.53 & 0.57 & 0.65 & 0.75 & 0.82 & 0.88 & 0.93 & 0.97 & 1.00 & 1.04 & 1.13 & 1.26 & 1.42 & 1.34 \\ \hline
        earrings & 0.02 & 0.02 & 0.02 & 0.02 & 0.03 & 0.03 & 0.03 & 0.03 & 0.03 & 0.03 & 0.03 & 0.04 & 0.04 & 0.05 & 0.03 \\ \hline
        eyeglasses & 1.32 & 1.32 & 1.31 & 1.30 & 1.29 & 1.27 & 1.19 & 1.26 & 1.24 & 1.23 & 1.22 & 1.18 & 1.13 & 1.07 & 0.90 \\ \hline
        face & 44.64 & 46.53 & 48.01 & 50.06 & 51.91 & 52.92 & 54.30 & 54.08 & 54.45 & 54.75 & 55.00 & 55.68 & 56.31 & 56.61 & 59.50 \\ \hline
        hair & 2.03 & 2.29 & 2.53 & 2.97 & 3.53 & 3.95 & 4.64 & 4.62 & 4.89 & 5.13 & 5.35 & 6.03 & 6.93 & 7.98 & 8.39 \\ \hline
        hat & 0.29 & 0.32 & 0.34 & 0.37 & 0.42 & 0.46 & 0.54 & 0.52 & 0.54 & 0.56 & 0.58 & 0.64 & 0.71 & 0.81 & 0.82 \\ \hline
        inner mouth & 1.08 & 1.13 & 1.14 & 1.14 & 1.13 & 1.11 & 1.00 & 1.07 & 1.06 & 1.05 & 1.04 & 1.00 & 0.95 & 0.90 & 0.65 \\ \hline
        left ear & 0.07 & 0.08 & 0.09 & 0.10 & 0.12 & 0.13 & 0.13 & 0.15 & 0.16 & 0.17 & 0.17 & 0.19 & 0.22 & 0.24 & 0.21 \\ \hline
        left eye & 3.47 & 3.21 & 2.96 & 2.61 & 2.28 & 2.09 & 1.78 & 1.85 & 1.78 & 1.72 & 1.66 & 1.51 & 1.34 & 1.19 & 0.81 \\ \hline
        left eyebrow & 1.70 & 1.74 & 1.78 & 1.81 & 1.82 & 1.82 & 1.71 & 1.79 & 1.77 & 1.75 & 1.73 & 1.67 & 1.56 & 1.41 & 1.22 \\ \hline
        lower lip & 2.33 & 2.45 & 2.50 & 2.52 & 2.48 & 2.43 & 2.16 & 2.34 & 2.31 & 2.28 & 2.25 & 2.16 & 2.05 & 1.94 & 1.56 \\ \hline
        neck & 0.28 & 0.33 & 0.37 & 0.46 & 0.57 & 0.66 & 0.71 & 0.79 & 0.84 & 0.88 & 0.92 & 1.06 & 1.24 & 1.47 & 1.57 \\ \hline
        necklace & 0.00 & 0.00 & 0.00 & 0.00 & 0.00 & 0.00 & 0.00 & 0.00 & 0.00 & 0.00 & 0.00 & 0.00 & 0.00 & 0.00 & 0.00 \\ \hline
        nose & 32.25 & 30.02 & 28.30 & 25.73 & 23.14 & 21.52 & 20.07 & 19.34 & 18.57 & 17.89 & 17.33 & 15.63 & 13.65 & 11.68 & 10.79 \\ \hline
        right ear & 0.08 & 0.10 & 0.11 & 0.13 & 0.15 & 0.17 & 0.16 & 0.19 & 0.19 & 0.20 & 0.21 & 0.23 & 0.26 & 0.29 & 0.24 \\ \hline
        right eyebrow & 1.99 & 2.00 & 2.01 & 2.02 & 2.00 & 1.98 & 1.85 & 1.93 & 1.91 & 1.88 & 1.86 & 1.78 & 1.65 & 1.49 & 1.28 \\ \hline
        right eye & 3.57 & 3.24 & 2.96 & 2.59 & 2.27 & 2.08 & 1.78 & 1.86 & 1.79 & 1.72 & 1.67 & 1.52 & 1.35 & 1.20 & 0.82 \\ \hline
        upper lip & 1.47 & 1.53 & 1.58 & 1.66 & 1.70 & 1.71 & 1.57 & 1.70 & 1.69 & 1.67 & 1.66 & 1.60 & 1.51 & 1.38 & 1.12 \\ \hline
    \end{tabular}
\end{table*}

\begin{table*}[t]
    \centering
    \small
    \caption{Percentage distribution of facial areas positively activated for AF\_R100 per activation threshold}
    \label{tab:thresholds_af100}
    \begin{tabular}{|l|c|c|c|c|c|c|c|c|c|c|c|c|c|c|c|}
    \hline
        Threshold (\%) & 0.01 & 0.05 & 0.10 & 0.25 & 0.50 & 0.75 & 1.00 & 1.25 & 1.50 & 1.75 & 2.00 & 3.00 & 5.00 & 10.00 & 100.00 \\ \hline
        background & 3.52 & 3.56 & 3.82 & 4.23 & 4.72 & 5.08 & 5.36 & 5.69 & 5.92 & 6.04 & 6.27 & 6.85 & 7.63 & 8.58 & 9.05 \\ \hline
        clothes & 0.36 & 0.40 & 0.44 & 0.51 & 0.58 & 0.64 & 0.66 & 0.73 & 0.77 & 0.79 & 0.83 & 0.92 & 1.03 & 1.17 & 1.07 \\ \hline
        earrings & 0.01 & 0.01 & 0.02 & 0.02 & 0.02 & 0.02 & 0.02 & 0.03 & 0.03 & 0.03 & 0.03 & 0.03 & 0.04 & 0.04 & 0.03 \\ \hline
        eyeglasses & 1.39 & 1.38 & 1.38 & 1.38 & 1.37 & 1.35 & 1.24 & 1.31 & 1.29 & 1.27 & 1.29 & 1.26 & 1.19 & 1.12 & 0.95 \\ \hline
        face & 44.59 & 46.99 & 48.68 & 51.00 & 52.99 & 54.07 & 56.57 & 55.30 & 55.68 & 56.04 & 56.22 & 56.81 & 57.20 & 57.07 & 59.61 \\ \hline
        hair & 2.24 & 2.51 & 2.75 & 3.19 & 3.72 & 4.10 & 4.11 & 4.65 & 4.89 & 5.12 & 5.31 & 5.89 & 6.65 & 7.49 & 7.50 \\ \hline
        hat & 0.42 & 0.45 & 0.46 & 0.50 & 0.54 & 0.57 & 0.59 & 0.63 & 0.64 & 0.66 & 0.68 & 0.74 & 0.81 & 0.90 & 0.91 \\ \hline
        inner mouth & 0.88 & 0.90 & 0.91 & 0.94 & 0.95 & 0.95 & 0.88 & 0.93 & 0.94 & 0.93 & 0.93 & 0.91 & 0.88 & 0.84 & 0.68 \\ \hline
        left ear & 0.03 & 0.04 & 0.05 & 0.06 & 0.07 & 0.09 & 0.09 & 0.10 & 0.11 & 0.11 & 0.12 & 0.13 & 0.15 & 0.17 & 0.16 \\ \hline
        left eye & 2.66 & 2.45 & 2.31 & 2.10 & 1.88 & 1.74 & 1.47 & 1.57 & 1.51 & 1.46 & 1.42 & 1.29 & 1.17 & 1.07 & 0.78 \\ \hline
        left eyebrow & 2.32 & 2.35 & 2.33 & 2.29 & 2.22 & 2.17 & 1.93 & 2.07 & 2.03 & 2.00 & 1.97 & 1.87 & 1.75 & 1.63 & 1.29 \\ \hline
        lower lip & 0.90 & 0.97 & 1.04 & 1.17 & 1.30 & 1.37 & 1.35 & 1.46 & 1.48 & 1.51 & 1.51 & 1.53 & 1.53 & 1.52 & 1.36 \\ \hline
        neck & 0.28 & 0.29 & 0.34 & 0.41 & 0.51 & 0.58 & 0.59 & 0.69 & 0.73 & 0.77 & 0.81 & 0.91 & 1.06 & 1.22 & 1.15 \\ \hline
        necklace & 0.00 & 0.00 & 0.00 & 0.00 & 0.00 & 0.00 & 0.00 & 0.00 & 0.00 & 0.00 & 0.00 & 0.00 & 0.00 & 0.00 & 0.00 \\ \hline
        nose & 33.60 & 31.05 & 28.91 & 25.89 & 23.07 & 21.40 & 20.00 & 19.25 & 18.51 & 17.91 & 17.37 & 15.85 & 14.19 & 12.74 & 12.02 \\ \hline
        right ear & 0.05 & 0.06 & 0.07 & 0.09 & 0.11 & 0.12 & 0.12 & 0.14 & 0.15 & 0.15 & 0.16 & 0.18 & 0.20 & 0.23 & 0.20 \\ \hline
        right eyebrow & 2.45 & 2.46 & 2.45 & 2.39 & 2.30 & 2.24 & 1.99 & 2.14 & 2.10 & 2.07 & 2.02 & 1.92 & 1.80 & 1.68 & 1.35 \\ \hline
        right eye & 2.84 & 2.62 & 2.46 & 2.21 & 1.96 & 1.81 & 1.52 & 1.62 & 1.56 & 1.50 & 1.45 & 1.33 & 1.20 & 1.09 & 0.79 \\ \hline
        upper lip & 1.45 & 1.51 & 1.57 & 1.64 & 1.68 & 1.69 & 1.51 & 1.67 & 1.66 & 1.65 & 1.63 & 1.57 & 1.51 & 1.44 & 1.10 \\ \hline
    \end{tabular}
\end{table*} 

\begin{table*}[t]
    \centering
    \small
    \caption{Percentage distribution of facial areas positively activated for CF\_R18 per activation threshold}
    \label{tab:thresholds_cf18}
    \begin{tabular}{|l|c|c|c|c|c|c|c|c|c|c|c|c|c|c|c|}
    \hline
        Threshold (\%) & 0.01 & 0.05 & 0.10 & 0.25 & 0.50 & 0.75 & 1.00 & 1.25 & 1.50 & 1.75 & 2.00 & 3.00 & 5.00 & 10.00 & 100.00 \\ \hline
        background & 2.92 & 3.22 & 3.50 & 3.98 & 4.52 & 4.87 & 5.34 & 5.38 & 5.57 & 5.75 & 5.90 & 6.42 & 7.15 & 8.17 & 9.03 \\ \hline
        clothes & 0.64 & 0.69 & 0.75 & 0.85 & 0.96 & 1.04 & 1.16 & 1.15 & 1.19 & 1.23 & 1.26 & 1.36 & 1.50 & 1.68 & 1.86 \\ \hline
        earrings & 0.03 & 0.03 & 0.03 & 0.03 & 0.03 & 0.04 & 0.05 & 0.04 & 0.04 & 0.04 & 0.04 & 0.05 & 0.05 & 0.06 & 0.07 \\ \hline
        eyeglasses & 1.35 & 1.33 & 1.32 & 1.31 & 1.30 & 1.30 & 1.29 & 1.29 & 1.28 & 1.27 & 1.27 & 1.25 & 1.21 & 1.16 & 1.12 \\ \hline
        face & 44.87 & 46.34 & 47.34 & 48.71 & 49.98 & 50.74 & 50.45 & 51.70 & 52.02 & 52.29 & 52.50 & 53.04 & 53.42 & 53.26 & 51.43 \\ \hline
        hair & 2.99 & 3.34 & 3.66 & 4.21 & 4.86 & 5.30 & 6.89 & 5.95 & 6.21 & 6.44 & 6.65 & 7.32 & 8.29 & 9.59 & 11.84 \\ \hline
        hat & 0.29 & 0.32 & 0.35 & 0.39 & 0.43 & 0.46 & 0.59 & 0.50 & 0.52 & 0.54 & 0.55 & 0.60 & 0.67 & 0.77 & 0.96 \\ \hline
        inner mouth & 0.72 & 0.76 & 0.79 & 0.84 & 0.88 & 0.90 & 0.91 & 0.92 & 0.92 & 0.92 & 0.92 & 0.91 & 0.89 & 0.86 & 0.83 \\ \hline
        left ear & 0.07 & 0.08 & 0.09 & 0.11 & 0.13 & 0.14 & 0.15 & 0.15 & 0.16 & 0.17 & 0.17 & 0.19 & 0.21 & 0.24 & 0.28 \\ \hline
        left eye & 3.88 & 3.55 & 3.27 & 2.90 & 2.58 & 2.40 & 2.11 & 2.19 & 2.11 & 2.05 & 2.00 & 1.85 & 1.69 & 1.53 & 1.24 \\ \hline
        left eyebrow & 2.22 & 2.26 & 2.29 & 2.31 & 2.29 & 2.27 & 2.30 & 2.21 & 2.18 & 2.16 & 2.13 & 2.05 & 1.94 & 1.80 & 1.78 \\ \hline
        lower lip & 1.04 & 1.12 & 1.22 & 1.36 & 1.49 & 1.57 & 1.55 & 1.68 & 1.72 & 1.74 & 1.77 & 1.82 & 1.86 & 1.84 & 1.67 \\ \hline
        neck & 0.83 & 0.93 & 1.05 & 1.22 & 1.40 & 1.52 & 1.53 & 1.67 & 1.73 & 1.78 & 1.83 & 1.97 & 2.16 & 2.40 & 2.42 \\ \hline
        necklace & 0.00 & 0.00 & 0.00 & 0.00 & 0.00 & 0.00 & 0.00 & 0.00 & 0.00 & 0.00 & 0.00 & 0.00 & 0.00 & 0.01 & 0.01 \\ \hline
        nose & 29.07 & 27.39 & 25.97 & 23.76 & 21.49 & 20.06 & 18.57 & 18.13 & 17.43 & 16.82 & 16.31 & 14.77 & 12.95 & 11.07 & 10.24 \\ \hline
        right ear & 0.07 & 0.09 & 0.10 & 0.13 & 0.15 & 0.17 & 0.18 & 0.19 & 0.20 & 0.21 & 0.21 & 0.24 & 0.27 & 0.31 & 0.32 \\ \hline
        right eyebrow & 2.94 & 2.93 & 2.92 & 2.86 & 2.76 & 2.68 & 2.71 & 2.57 & 2.53 & 2.49 & 2.45 & 2.34 & 2.19 & 2.02 & 2.03 \\ \hline
        right eye & 4.28 & 3.83 & 3.50 & 3.09 & 2.73 & 2.53 & 2.27 & 2.28 & 2.19 & 2.12 & 2.06 & 1.89 & 1.70 & 1.52 & 1.31 \\ \hline
        upper lip & 1.79 & 1.82 & 1.87 & 1.94 & 2.00 & 2.01 & 1.94 & 2.01 & 2.00 & 1.99 & 1.98 & 1.93 & 1.84 & 1.72 & 1.57 \\ \hline
    \end{tabular}
\end{table*}

\begin{table*}[t]
    \centering
    \small
    \caption{Percentage distribution of facial areas positively activated for CF\_R50 per activation threshold}
    \label{tab:thresholds_cf50}
    \begin{tabular}{|l|c|c|c|c|c|c|c|c|c|c|c|c|c|c|c|}
    \hline
        Threshold (\%) & 0.01 & 0.05 & 0.10 & 0.25 & 0.50 & 0.75 & 1.00 & 1.25 & 1.50 & 1.75 & 2.00 & 3.00 & 5.00 & 10.00 & 100.00 \\ \hline
        background & 2.74 & 2.88 & 3.03 & 3.30 & 3.64 & 3.91 & 4.67 & 4.35 & 4.53 & 4.67 & 4.85 & 5.35 & 6.08 & 7.07 & 8.77 \\ \hline
        clothes & 0.38 & 0.40 & 0.42 & 0.46 & 0.51 & 0.55 & 0.65 & 0.62 & 0.65 & 0.67 & 0.70 & 0.77 & 0.88 & 1.03 & 1.28 \\ \hline
        earrings & 0.02 & 0.02 & 0.02 & 0.02 & 0.03 & 0.03 & 0.03 & 0.03 & 0.03 & 0.03 & 0.03 & 0.03 & 0.04 & 0.04 & 0.04 \\ \hline
        eyeglasses & 1.28 & 1.28 & 1.28 & 1.29 & 1.29 & 1.29 & 1.23 & 1.28 & 1.27 & 1.28 & 1.25 & 1.22 & 1.18 & 1.11 & 1.02 \\ \hline
        face & 45.03 & 47.07 & 48.72 & 51.10 & 53.28 & 54.50 & 55.68 & 55.97 & 56.46 & 56.84 & 57.20 & 58.12 & 58.97 & 59.30 & 57.79 \\ \hline
        hair & 1.72 & 1.90 & 2.09 & 2.45 & 2.91 & 3.26 & 4.08 & 3.82 & 4.04 & 4.29 & 4.41 & 4.98 & 5.74 & 6.76 & 8.76 \\ \hline
        hat & 0.28 & 0.30 & 0.31 & 0.34 & 0.37 & 0.39 & 0.48 & 0.43 & 0.44 & 0.47 & 0.47 & 0.51 & 0.58 & 0.66 & 0.82 \\ \hline
        inner mouth & 0.84 & 0.85 & 0.87 & 0.89 & 0.92 & 0.93 & 0.89 & 0.93 & 0.93 & 0.93 & 0.93 & 0.91 & 0.87 & 0.82 & 0.71 \\ \hline
        left ear & 0.08 & 0.08 & 0.09 & 0.10 & 0.11 & 0.12 & 0.13 & 0.13 & 0.14 & 0.14 & 0.15 & 0.17 & 0.19 & 0.22 & 0.26 \\ \hline
        left eye & 3.48 & 3.26 & 3.01 & 2.66 & 2.33 & 2.14 & 1.87 & 1.89 & 1.81 & 1.73 & 1.68 & 1.50 & 1.31 & 1.14 & 0.99 \\ \hline
        left eyebrow & 2.66 & 2.68 & 2.70 & 2.69 & 2.64 & 2.59 & 2.38 & 2.50 & 2.46 & 2.42 & 2.39 & 2.27 & 2.11 & 1.91 & 1.62 \\ \hline
        lower lip & 1.00 & 1.08 & 1.16 & 1.31 & 1.46 & 1.56 & 1.51 & 1.67 & 1.71 & 1.74 & 1.75 & 1.80 & 1.80 & 1.75 & 1.54 \\ \hline
        neck & 0.15 & 0.17 & 0.19 & 0.24 & 0.30 & 0.35 & 0.41 & 0.44 & 0.47 & 0.50 & 0.54 & 0.65 & 0.83 & 1.10 & 1.38 \\ \hline
        necklace & 0.00 & 0.00 & 0.00 & 0.00 & 0.00 & 0.00 & 0.00 & 0.00 & 0.00 & 0.00 & 0.00 & 0.00 & 0.00 & 0.00 & 0.00 \\ \hline
        nose & 33.24 & 31.00 & 29.16 & 26.40 & 23.68 & 22.02 & 20.17 & 19.82 & 19.04 & 18.37 & 17.80 & 16.13 & 14.22 & 12.34 & 10.95 \\ \hline
        right ear & 0.09 & 0.09 & 0.10 & 0.11 & 0.12 & 0.13 & 0.14 & 0.14 & 0.15 & 0.15 & 0.16 & 0.18 & 0.20 & 0.23 & 0.25 \\ \hline
        right eyebrow & 2.18 & 2.23 & 2.26 & 2.28 & 2.28 & 2.26 & 2.09 & 2.21 & 2.19 & 2.17 & 2.14 & 2.06 & 1.94 & 1.78 & 1.50 \\ \hline
        right eye & 3.14 & 2.95 & 2.76 & 2.46 & 2.19 & 2.02 & 1.78 & 1.80 & 1.73 & 1.66 & 1.61 & 1.45 & 1.26 & 1.10 & 0.96 \\ \hline
        upper lip & 1.69 & 1.76 & 1.81 & 1.89 & 1.95 & 1.97 & 1.81 & 1.97 & 1.96 & 1.94 & 1.93 & 1.88 & 1.78 & 1.64 & 1.34 \\ \hline
    \end{tabular}
\end{table*}

\begin{table*}[t]
    \centering
    \small
    \caption{Percentage distribution of facial areas positively activated for CF\_R100 per activation threshold}
    \label{tab:thresholds_cf100}
    \begin{tabular}{|l|c|c|c|c|c|c|c|c|c|c|c|c|c|c|c|}
    \hline
        Threshold (\%) & 0.01 & 0.05 & 0.10 & 0.25 & 0.50 & 0.75 & 1.00 & 1.25 & 1.50 & 1.75 & 2.00 & 3.00 & 5.00 & 10.00 & 100.00 \\ \hline
        background & 4.06 & 4.26 & 4.34 & 4.47 & 4.63 & 4.76 & 5.48 & 4.89 & 5.11 & 5.17 & 5.24 & 5.51 & 5.90 & 6.49 & 7.74 \\ \hline
        clothes & 0.65 & 0.68 & 0.68 & 0.68 & 0.68 & 0.69 & 0.83 & 0.71 & 0.73 & 0.73 & 0.74 & 0.77 & 0.82 & 0.91 & 1.12 \\ \hline
        earrings & 0.03 & 0.03 & 0.03 & 0.03 & 0.03 & 0.03 & 0.04 & 0.04 & 0.03 & 0.03 & 0.04 & 0.04 & 0.04 & 0.04 & 0.05 \\ \hline
        eyeglasses & 1.29 & 1.27 & 1.26 & 1.23 & 1.21 & 1.20 & 1.24 & 1.16 & 1.16 & 1.17 & 1.16 & 1.14 & 1.12 & 1.09 & 1.05 \\ \hline
        face & 46.73 & 48.15 & 49.46 & 51.42 & 53.38 & 54.57 & 53.63 & 56.13 & 56.69 & 57.16 & 57.55 & 58.66 & 59.77 & 60.38 & 58.42 \\ \hline
        hair & 5.63 & 5.79 & 5.83 & 5.93 & 6.09 & 6.23 & 7.71 & 6.54 & 6.53 & 6.64 & 6.72 & 7.00 & 7.42 & 8.12 & 10.21 \\ \hline
        hat & 0.66 & 0.64 & 0.65 & 0.65 & 0.65 & 0.65 & 0.75 & 0.66 & 0.66 & 0.66 & 0.67 & 0.69 & 0.72 & 0.78 & 0.91 \\ \hline
        inner mouth & 0.71 & 0.74 & 0.74 & 0.75 & 0.75 & 0.75 & 0.76 & 0.76 & 0.75 & 0.76 & 0.76 & 0.75 & 0.75 & 0.74 & 0.68 \\ \hline
        left ear & 0.14 & 0.14 & 0.14 & 0.14 & 0.14 & 0.15 & 0.15 & 0.15 & 0.16 & 0.16 & 0.16 & 0.17 & 0.18 & 0.20 & 0.21 \\ \hline
        left eye & 1.73 & 1.58 & 1.48 & 1.38 & 1.29 & 1.23 & 1.23 & 1.15 & 1.13 & 1.10 & 1.09 & 1.02 & 0.96 & 0.89 & 0.84 \\ \hline
        left eyebrow & 2.41 & 2.39 & 2.37 & 2.32 & 2.26 & 2.21 & 2.27 & 2.13 & 2.11 & 2.08 & 2.05 & 1.96 & 1.85 & 1.72 & 1.66 \\ \hline
        lower lip & 0.87 & 0.92 & 0.96 & 1.04 & 1.14 & 1.20 & 1.23 & 1.28 & 1.31 & 1.34 & 1.36 & 1.43 & 1.49 & 1.54 & 1.47 \\ \hline
        neck & 0.23 & 0.25 & 0.27 & 0.30 & 0.33 & 0.36 & 0.43 & 0.39 & 0.41 & 0.42 & 0.43 & 0.48 & 0.55 & 0.69 & 0.96 \\ \hline
        necklace & 0.00 & 0.00 & 0.00 & 0.00 & 0.00 & 0.00 & 0.00 & 0.00 & 0.00 & 0.00 & 0.00 & 0.00 & 0.00 & 0.00 & 0.00 \\ \hline
        nose & 28.62 & 27.11 & 25.89 & 23.95 & 21.85 & 20.50 & 18.87 & 18.73 & 17.98 & 17.41 & 16.93 & 15.47 & 13.76 & 12.01 & 10.58 \\ \hline
        right ear & 0.16 & 0.16 & 0.16 & 0.17 & 0.17 & 0.18 & 0.19 & 0.19 & 0.19 & 0.20 & 0.20 & 0.21 & 0.23 & 0.25 & 0.27 \\ \hline
        right eyebrow & 2.95 & 2.88 & 2.82 & 2.71 & 2.60 & 2.53 & 2.47 & 2.43 & 2.39 & 2.34 & 2.31 & 2.19 & 2.04 & 1.86 & 1.71 \\ \hline
        right eye & 1.81 & 1.66 & 1.56 & 1.45 & 1.35 & 1.29 & 1.26 & 1.19 & 1.18 & 1.15 & 1.13 & 1.07 & 0.99 & 0.92 & 0.84 \\ \hline
        upper lip & 1.31 & 1.34 & 1.36 & 1.41 & 1.44 & 1.45 & 1.46 & 1.46 & 1.46 & 1.46 & 1.46 & 1.45 & 1.41 & 1.37 & 1.27 \\ \hline
    \end{tabular}
\end{table*}

\begin{table*}[t]
    \centering
    \small
    \caption{Percentage distribution of facial areas positively activated for IR\_CASIA per activation threshold}
    \label{tab:thresholds_ircasia}
    \begin{tabular}{|l|c|c|c|c|c|c|c|c|c|c|c|c|c|c|c|}
    \hline
        Threshold (\%) & 0.01 & 0.05 & 0.10 & 0.25 & 0.50 & 0.75 & 1.00 & 1.25 & 1.50 & 1.75 & 2.00 & 3.00 & 5.00 & 10.00 & 100.00 \\ \hline 
        background & 4.27 & 4.03 & 3.99 & 3.99 & 4.05 & 4.11 & 4.17 & 4.22 & 4.27 & 4.32 & 4.37 & 4.56 & 4.89 & 5.60 & 10.95 \\ \hline
        clothes & 0.78 & 0.68 & 0.65 & 0.64 & 0.64 & 0.64 & 0.65 & 0.66 & 0.66 & 0.67 & 0.67 & 0.70 & 0.75 & 0.86 & 1.74 \\ \hline
        earrings & 0.02 & 0.02 & 0.02 & 0.02 & 0.02 & 0.02 & 0.02 & 0.02 & 0.02 & 0.02 & 0.02 & 0.02 & 0.03 & 0.03 & 0.06 \\ \hline
        eyeglasses & 0.70 & 0.69 & 0.69 & 0.70 & 0.72 & 0.73 & 0.75 & 0.76 & 0.77 & 0.78 & 0.80 & 0.83 & 0.89 & 0.94 & 0.85 \\ \hline
        face & 48.05 & 47.81 & 47.77 & 48.12 & 48.95 & 49.77 & 50.52 & 51.19 & 51.82 & 52.40 & 52.96 & 54.86 & 57.47 & 60.40 & 55.32 \\ \hline
        hair & 4.20 & 4.18 & 4.18 & 4.21 & 4.29 & 4.37 & 4.43 & 4.50 & 4.56 & 4.61 & 4.67 & 4.88 & 5.26 & 6.12 & 12.09 \\ \hline
        hat & 0.28 & 0.28 & 0.28 & 0.28 & 0.29 & 0.29 & 0.30 & 0.30 & 0.31 & 0.31 & 0.32 & 0.33 & 0.36 & 0.43 & 0.95 \\ \hline
        inner mouth & 0.52 & 0.51 & 0.50 & 0.51 & 0.52 & 0.54 & 0.56 & 0.57 & 0.58 & 0.60 & 0.61 & 0.65 & 0.71 & 0.79 & 0.58 \\ \hline
        left ear & 0.04 & 0.03 & 0.03 & 0.04 & 0.04 & 0.04 & 0.04 & 0.05 & 0.05 & 0.05 & 0.05 & 0.06 & 0.07 & 0.10 & 0.29 \\ \hline
        left eye & 0.54 & 0.57 & 0.58 & 0.60 & 0.63 & 0.65 & 0.67 & 0.68 & 0.69 & 0.70 & 0.71 & 0.74 & 0.79 & 0.80 & 0.56 \\ \hline
        left eyebrow & 2.59 & 2.47 & 2.39 & 2.28 & 2.19 & 2.13 & 2.09 & 2.05 & 2.03 & 2.00 & 1.99 & 1.93 & 1.86 & 1.75 & 1.32 \\ \hline
        lower lip & 0.82 & 0.86 & 0.89 & 0.94 & 1.00 & 1.04 & 1.08 & 1.11 & 1.14 & 1.16 & 1.19 & 1.28 & 1.44 & 1.74 & 1.45 \\ \hline
        neck & 0.84 & 0.59 & 0.55 & 0.55 & 0.56 & 0.58 & 0.59 & 0.60 & 0.62 & 0.63 & 0.64 & 0.68 & 0.75 & 0.95 & 2.24 \\ \hline
        necklace & 0.00 & 0.00 & 0.00 & 0.00 & 0.00 & 0.00 & 0.00 & 0.00 & 0.00 & 0.00 & 0.00 & 0.00 & 0.00 & 0.00 & 0.01 \\ \hline
        nose & 31.34 & 32.48 & 32.77 & 32.57 & 31.64 & 30.65 & 29.72 & 28.86 & 28.05 & 27.29 & 26.56 & 24.00 & 20.23 & 15.04 & 8.28 \\ \hline
        right ear & 0.04 & 0.04 & 0.04 & 0.05 & 0.05 & 0.05 & 0.05 & 0.06 & 0.06 & 0.06 & 0.06 & 0.07 & 0.08 & 0.11 & 0.31 \\ \hline
        right eyebrow & 3.13 & 2.92 & 2.80 & 2.62 & 2.47 & 2.38 & 2.32 & 2.27 & 2.23 & 2.20 & 2.17 & 2.09 & 1.99 & 1.84 & 1.35 \\ \hline
        right eye & 0.61 & 0.65 & 0.67 & 0.69 & 0.71 & 0.74 & 0.75 & 0.77 & 0.78 & 0.79 & 0.80 & 0.82 & 0.84 & 0.82 & 0.56 \\ \hline
        upper lip & 1.23 & 1.21 & 1.20 & 1.21 & 1.24 & 1.27 & 1.31 & 1.34 & 1.36 & 1.39 & 1.41 & 1.49 & 1.59 & 1.65 & 1.11 \\ \hline
    \end{tabular}
\end{table*}

\begin{table*}[t]
    \centering
    \small
    \caption{Percentage distribution of facial areas positively activated for IR\_VGG per activation threshold}
    \label{tab:thresholds_irvgg}
    \begin{tabular}{|l|c|c|c|c|c|c|c|c|c|c|c|c|c|c|c|}
    \hline
        Threshold (\%) & 0.01 & 0.05 & 0.10 & 0.25 & 0.50 & 0.75 & 1.00 & 1.25 & 1.50 & 1.75 & 2.00 & 3.00 & 5.00 & 10.00 & 100.00 \\ \hline
        background & 4.74 & 4.35 & 4.27 & 4.25 & 4.29 & 4.34 & 4.38 & 4.43 & 4.47 & 4.51 & 4.55 & 4.71 & 4.99 & 5.62 & 10.76 \\ \hline
        clothes & 1.11 & 0.89 & 0.84 & 0.81 & 0.81 & 0.81 & 0.81 & 0.82 & 0.82 & 0.82 & 0.83 & 0.85 & 0.89 & 1.02 & 2.08 \\ \hline
        earrings & 0.02 & 0.01 & 0.01 & 0.01 & 0.01 & 0.01 & 0.02 & 0.02 & 0.02 & 0.02 & 0.02 & 0.02 & 0.02 & 0.03 & 0.07 \\ \hline
        eyeglasses & 0.66 & 0.66 & 0.67 & 0.68 & 0.70 & 0.72 & 0.73 & 0.74 & 0.76 & 0.77 & 0.78 & 0.82 & 0.88 & 0.95 & 0.88 \\ \hline
        face & 51.05 & 51.60 & 51.77 & 52.15 & 52.72 & 53.22 & 53.66 & 54.07 & 54.46 & 54.83 & 55.19 & 56.46 & 58.34 & 60.87 & 55.46 \\ \hline
        hair & 3.35 & 3.31 & 3.29 & 3.29 & 3.33 & 3.37 & 3.42 & 3.46 & 3.50 & 3.55 & 3.59 & 3.76 & 4.08 & 4.86 & 11.50 \\ \hline
        hat & 0.34 & 0.34 & 0.34 & 0.34 & 0.34 & 0.35 & 0.35 & 0.35 & 0.35 & 0.36 & 0.36 & 0.37 & 0.38 & 0.43 & 0.90 \\ \hline
        inner mouth & 0.68 & 0.67 & 0.67 & 0.68 & 0.70 & 0.72 & 0.74 & 0.76 & 0.78 & 0.79 & 0.80 & 0.83 & 0.86 & 0.87 & 0.59 \\ \hline
        left ear & 0.07 & 0.07 & 0.07 & 0.07 & 0.07 & 0.07 & 0.07 & 0.07 & 0.07 & 0.07 & 0.07 & 0.08 & 0.09 & 0.11 & 0.32 \\ \hline
        left eye & 0.50 & 0.52 & 0.54 & 0.57 & 0.61 & 0.64 & 0.66 & 0.67 & 0.69 & 0.70 & 0.71 & 0.75 & 0.80 & 0.82 & 0.59 \\ \hline
        left eyebrow & 2.60 & 2.40 & 2.29 & 2.14 & 2.03 & 1.97 & 1.93 & 1.90 & 1.88 & 1.86 & 1.85 & 1.81 & 1.78 & 1.72 & 1.32 \\ \hline
        lower lip & 0.89 & 0.96 & 1.02 & 1.12 & 1.22 & 1.29 & 1.34 & 1.38 & 1.42 & 1.45 & 1.48 & 1.57 & 1.70 & 1.88 & 1.45 \\ \hline
        neck & 1.58 & 1.06 & 0.98 & 0.96 & 0.99 & 1.01 & 1.02 & 1.03 & 1.04 & 1.05 & 1.06 & 1.08 & 1.15 & 1.35 & 2.70 \\ \hline
        necklace & 0.00 & 0.00 & 0.00 & 0.00 & 0.00 & 0.00 & 0.00 & 0.00 & 0.00 & 0.00 & 0.00 & 0.00 & 0.00 & 0.00 & 0.01 \\ \hline
        nose & 27.34 & 28.28 & 28.47 & 28.25 & 27.54 & 26.84 & 26.19 & 25.60 & 25.03 & 24.50 & 23.99 & 22.16 & 19.33 & 14.93 & 7.95 \\ \hline
        right ear & 0.08 & 0.07 & 0.07 & 0.07 & 0.07 & 0.07 & 0.07 & 0.08 & 0.08 & 0.08 & 0.08 & 0.08 & 0.10 & 0.12 & 0.36 \\ \hline
        right eyebrow & 2.79 & 2.56 & 2.44 & 2.28 & 2.16 & 2.09 & 2.05 & 2.02 & 2.00 & 1.98 & 1.96 & 1.92 & 1.87 & 1.78 & 1.34 \\ \hline
        right eye & 0.61 & 0.63 & 0.64 & 0.66 & 0.68 & 0.71 & 0.73 & 0.75 & 0.76 & 0.78 & 0.79 & 0.83 & 0.87 & 0.87 & 0.59 \\ \hline
        upper lip & 1.61 & 1.61 & 1.62 & 1.66 & 1.73 & 1.78 & 1.83 & 1.85 & 1.88 & 1.89 & 1.90 & 1.91 & 1.89 & 1.78 & 1.13 \\ \hline
    \end{tabular}
\end{table*}

\begin{table*}[t]
    \centering
    \small
    \caption{Percentage distribution of facial areas positively activated for VGG\_16 per activation threshold}
    \label{tab:thresholds_vgg16}
    \begin{tabular}{|l|c|c|c|c|c|c|c|c|c|c|c|c|c|c|c|}
    \hline
        Threshold (\%) & 0.01 & 0.05 & 0.10 & 0.25 & 0.50 & 0.75 & 1.00 & 1.25 & 1.50 & 1.75 & 2.00 & 3.00 & 5.00 & 10.00 & 100.00 \\ \hline
        background & 5.94 & 6.31 & 6.60 & 7.19 & 7.70 & 8.09 & 8.39 & 8.64 & 8.82 & 8.98 & 9.18 & 9.54 & 10.09 & 10.95 & 32.90 \\ \hline
        clothes & 0.40 & 0.46 & 0.50 & 0.57 & 0.64 & 0.68 & 0.72 & 0.74 & 0.78 & 0.80 & 0.81 & 0.87 & 0.93 & 1.07 & 3.89 \\ \hline
        earrings & 0.01 & 0.01 & 0.01 & 0.01 & 0.02 & 0.02 & 0.02 & 0.02 & 0.02 & 0.02 & 0.02 & 0.02 & 0.02 & 0.03 & 0.09 \\ \hline
        eyeglasses & 1.53 & 1.47 & 1.46 & 1.46 & 1.42 & 1.41 & 1.40 & 1.38 & 1.39 & 1.39 & 1.40 & 1.37 & 1.39 & 1.30 & 0.38 \\ \hline
        face & 40.38 & 42.50 & 43.83 & 45.80 & 47.23 & 47.99 & 48.51 & 48.92 & 49.18 & 49.43 & 49.61 & 50.28 & 50.87 & 51.61 & 27.14 \\ \hline
        hair & 5.63 & 5.96 & 6.31 & 6.92 & 7.54 & 7.95 & 8.25 & 8.48 & 8.66 & 8.82 & 8.95 & 9.37 & 10.03 & 10.93 & 24.67 \\ \hline
        hat & 0.88 & 0.91 & 0.94 & 1.02 & 1.06 & 1.10 & 1.12 & 1.14 & 1.16 & 1.17 & 1.19 & 1.22 & 1.34 & 1.36 & 2.23 \\ \hline
        inner mouth & 0.59 & 0.59 & 0.59 & 0.61 & 0.63 & 0.64 & 0.65 & 0.65 & 0.66 & 0.67 & 0.67 & 0.68 & 0.68 & 0.69 & 0.27 \\ \hline
        left ear & 0.08 & 0.08 & 0.09 & 0.09 & 0.11 & 0.11 & 0.12 & 0.12 & 0.13 & 0.14 & 0.14 & 0.15 & 0.17 & 0.20 & 0.44 \\ \hline
        left eye & 1.72 & 1.63 & 1.56 & 1.45 & 1.38 & 1.34 & 1.31 & 1.27 & 1.26 & 1.24 & 1.22 & 1.18 & 1.09 & 0.97 & 0.20 \\ \hline
        left eyebrow & 3.08 & 3.04 & 2.96 & 2.80 & 2.69 & 2.61 & 2.55 & 2.48 & 2.45 & 2.41 & 2.36 & 2.25 & 2.05 & 1.82 & 0.40 \\ \hline
        lower lip & 0.49 & 0.51 & 0.54 & 0.61 & 0.67 & 0.73 & 0.77 & 0.81 & 0.83 & 0.86 & 0.88 & 0.96 & 1.07 & 1.18 & 0.62 \\ \hline
        neck & 0.04 & 0.06 & 0.07 & 0.09 & 0.12 & 0.14 & 0.16 & 0.18 & 0.19 & 0.20 & 0.20 & 0.23 & 0.27 & 0.33 & 2.82 \\ \hline
        necklace & 0.00 & 0.00 & 0.00 & 0.00 & 0.00 & 0.00 & 0.00 & 0.00 & 0.00 & 0.00 & 0.00 & 0.00 & 0.00 & 0.00 & 0.01 \\ \hline
        nose & 33.32 & 30.78 & 28.99 & 26.01 & 23.56 & 22.04 & 20.95 & 20.13 & 19.49 & 18.94 & 18.45 & 17.10 & 15.40 & 13.35 & 2.36 \\ \hline
        right ear & 0.07 & 0.08 & 0.08 & 0.09 & 0.11 & 0.12 & 0.13 & 0.13 & 0.14 & 0.14 & 0.15 & 0.16 & 0.19 & 0.22 & 0.52 \\ \hline
        right eyebrow & 3.27 & 3.17 & 3.09 & 2.93 & 2.82 & 2.73 & 2.66 & 2.60 & 2.56 & 2.52 & 2.48 & 2.35 & 2.17 & 1.88 & 0.42 \\ \hline
        right eye & 1.82 & 1.67 & 1.59 & 1.48 & 1.40 & 1.35 & 1.32 & 1.29 & 1.27 & 1.25 & 1.24 & 1.18 & 1.11 & 0.98 & 0.21 \\ \hline
        upper lip & 0.76 & 0.77 & 0.80 & 0.85 & 0.90 & 0.94 & 0.97 & 1.00 & 1.01 & 1.03 & 1.04 & 1.08 & 1.12 & 1.14 & 0.43 \\ \hline
    \end{tabular}
\end{table*}

%% file: submission-template/main.bbl
\begin{thebibliography}{46}


\ifx \showCODEN    \undefined \def \showCODEN     #1{\unskip}     \fi
\ifx \showDOI      \undefined \def \showDOI       #1{#1}\fi
\ifx \showISBNx    \undefined \def \showISBNx     #1{\unskip}     \fi
\ifx \showISBNxiii \undefined \def \showISBNxiii  #1{\unskip}     \fi
\ifx \showISSN     \undefined \def \showISSN      #1{\unskip}     \fi
\ifx \showLCCN     \undefined \def \showLCCN      #1{\unskip}     \fi
\ifx \shownote     \undefined \def \shownote      #1{#1}          \fi
\ifx \showarticletitle \undefined \def \showarticletitle #1{#1}   \fi
\ifx \showURL      \undefined \def \showURL       {\relax}        \fi
\providecommand\bibfield[2]{#2}
\providecommand\bibinfo[2]{#2}
\providecommand\natexlab[1]{#1}
\providecommand\showeprint[2][]{arXiv:#2}

\bibitem[Albanie(2015)]%
        {vggweights}
\bibfield{author}{\bibinfo{person}{Samuel Albanie}.} \bibinfo{year}{2015}\natexlab{}.
\newblock \bibinfo{title}{Vgg Face model}.
\newblock
\newblock
\urldef\tempurl%
\url{https://www.robots.ox.ac.uk/~albanie/pytorch-models.html}
\showURL{%
\tempurl}
\newblock
\shownote{Accessed: 2025-02-27}.


\bibitem[An et~al\mbox{.}(2021)]%
        {an2021partialfctraining10}
\bibfield{author}{\bibinfo{person}{Xiang An}, \bibinfo{person}{Xuhan Zhu}, \bibinfo{person}{Yang Xiao}, \bibinfo{person}{Lan Wu}, \bibinfo{person}{Ming Zhang}, \bibinfo{person}{Yuan Gao}, \bibinfo{person}{Bin Qin}, \bibinfo{person}{Debing Zhang}, {and} \bibinfo{person}{Ying Fu}.} \bibinfo{year}{2021}\natexlab{}.
\newblock \bibinfo{title}{Partial FC: Training 10 Million Identities on a Single Machine}.
\newblock
\newblock
\showeprint[arxiv]{2010.05222}~[cs.CV]
\urldef\tempurl%
\url{https://arxiv.org/abs/2010.05222}
\showURL{%
\tempurl}


\bibitem[Bachhawat(2024)]%
        {bachhawat2024}
\bibfield{author}{\bibinfo{person}{Mudit Bachhawat}.} \bibinfo{year}{2024}\natexlab{}.
\newblock \bibinfo{title}{Generalizing GradCAM for Embedding Networks}.
\newblock
\newblock
\showeprint[arxiv]{2402.00909}~[cs.CV]
\urldef\tempurl%
\url{https://arxiv.org/abs/2402.00909}
\showURL{%
\tempurl}


\bibitem[Behrmann et~al\mbox{.}(2019)]%
        {behrmann2019invertible}
\bibfield{author}{\bibinfo{person}{Jens Behrmann}, \bibinfo{person}{Will Grathwohl}, \bibinfo{person}{Ricky~TQ Chen}, \bibinfo{person}{David Duvenaud}, {and} \bibinfo{person}{J{\"o}rn-Henrik Jacobsen}.} \bibinfo{year}{2019}\natexlab{}.
\newblock \showarticletitle{Invertible residual networks}. In \bibinfo{booktitle}{\emph{International conference on machine learning}}. PMLR, \bibinfo{pages}{573--582}.
\newblock


\bibitem[Bhattacharyya(1946)]%
        {Bhattacharyya}
\bibfield{author}{\bibinfo{person}{A. Bhattacharyya}.} \bibinfo{year}{1946}\natexlab{}.
\newblock \showarticletitle{On a Measure of Divergence between Two Multinomial Populations}.
\newblock \bibinfo{journal}{\emph{Sankhyā: The Indian Journal of Statistics (1933-1960)}} \bibinfo{volume}{7}, \bibinfo{number}{4} (\bibinfo{year}{1946}), \bibinfo{pages}{401--406}.
\newblock
\showISSN{00364452}
\urldef\tempurl%
\url{http://www.jstor.org/stable/25047882}
\showURL{%
\tempurl}


\bibitem[Boutros et~al\mbox{.}(2023)]%
        {BoutrosEtAl2023}
\bibfield{author}{\bibinfo{person}{Fadi Boutros}, \bibinfo{person}{Vitomir Struc}, \bibinfo{person}{Julian Fierrez}, {and} \bibinfo{person}{Naser Damer}.} \bibinfo{year}{2023}\natexlab{}.
\newblock \showarticletitle{Synthetic data for face recognition: Current state and future prospects}.
\newblock \bibinfo{journal}{\emph{Image and Vision Computing}}  \bibinfo{volume}{135} (\bibinfo{year}{2023}), \bibinfo{pages}{104688}.
\newblock
\showISSN{0262-8856}
\urldef\tempurl%
\url{https://doi.org/10.1016/j.imavis.2023.104688}
\showDOI{\tempurl}


\bibitem[Cao et~al\mbox{.}(2018)]%
        {CaoEtAl2018}
\bibfield{author}{\bibinfo{person}{Qiong Cao}, \bibinfo{person}{Li Shen}, \bibinfo{person}{Weidi Xie}, \bibinfo{person}{Omkar~M. Parkhi}, {and} \bibinfo{person}{Andrew Zisserman}.} \bibinfo{year}{2018}\natexlab{}.
\newblock \bibinfo{title}{VGGFace2: A dataset for recognising faces across pose and age}.
\newblock
\newblock
\showeprint[arxiv]{1710.08092}~[cs.CV]
\urldef\tempurl%
\url{https://arxiv.org/abs/1710.08092}
\showURL{%
\tempurl}


\bibitem[Chattopadhyay et~al\mbox{.}(2017)]%
        {ChattopadhyayEtAl2017}
\bibfield{author}{\bibinfo{person}{Aditya Chattopadhyay}, \bibinfo{person}{Anirban Sarkar}, \bibinfo{person}{Prantik Howlader}, {and} \bibinfo{person}{Vineeth~N. Balasubramanian}.} \bibinfo{year}{2017}\natexlab{}.
\newblock \showarticletitle{Grad-CAM++: Generalized Gradient-based Visual Explanations for Deep Convolutional Networks}.
\newblock \bibinfo{journal}{\emph{CoRR}}  \bibinfo{volume}{abs/1710.11063} (\bibinfo{year}{2017}).
\newblock
\showeprint[arXiv]{1710.11063}
\urldef\tempurl%
\url{http://arxiv.org/abs/1710.11063}
\showURL{%
\tempurl}


\bibitem[Chen et~al\mbox{.}(2020)]%
        {ChenEtAl2020}
\bibfield{author}{\bibinfo{person}{Lei Chen}, \bibinfo{person}{Jianhui Chen}, \bibinfo{person}{Hossein Hajimirsadeghi}, {and} \bibinfo{person}{Greg Mori}.} \bibinfo{year}{2020}\natexlab{}.
\newblock \showarticletitle{Adapting Grad-CAM for Embedding Networks}.
\newblock \bibinfo{journal}{\emph{CoRR}}  \bibinfo{volume}{abs/2001.06538} (\bibinfo{year}{2020}).
\newblock
\showeprint[arXiv]{2001.06538}
\urldef\tempurl%
\url{https://arxiv.org/abs/2001.06538}
\showURL{%
\tempurl}


\bibitem[Cherepanova et~al\mbox{.}(2021)]%
        {cherepanova2021lowkey}
\bibfield{author}{\bibinfo{person}{Valeriia Cherepanova}, \bibinfo{person}{Micah Goldblum}, \bibinfo{person}{Harrison Foley}, \bibinfo{person}{Shiyuan Duan}, \bibinfo{person}{John Dickerson}, \bibinfo{person}{Gavin Taylor}, {and} \bibinfo{person}{Tom Goldstein}.} \bibinfo{year}{2021}\natexlab{}.
\newblock \showarticletitle{Lowkey: Leveraging adversarial attacks to protect social media users from facial recognition}.
\newblock \bibinfo{journal}{\emph{arXiv preprint arXiv:2101.07922}} (\bibinfo{year}{2021}).
\newblock


\bibitem[Deng et~al\mbox{.}(2019a)]%
        {deng2018arcface}
\bibfield{author}{\bibinfo{person}{Jiankang Deng}, \bibinfo{person}{Jia Guo}, \bibinfo{person}{Xue Niannan}, {and} \bibinfo{person}{Stefanos Zafeiriou}.} \bibinfo{year}{2019}\natexlab{a}.
\newblock \showarticletitle{ArcFace: Additive Angular Margin Loss for Deep Face Recognition}. In \bibinfo{booktitle}{\emph{CVPR}}.
\newblock


\bibitem[Deng et~al\mbox{.}(2019b)]%
        {DengEtAl2019}
\bibfield{author}{\bibinfo{person}{Jiankang Deng}, \bibinfo{person}{Jia Guo}, \bibinfo{person}{Debing Zhang}, \bibinfo{person}{Yafeng Deng}, \bibinfo{person}{Xiangju Lu}, {and} \bibinfo{person}{Song Shi}.} \bibinfo{year}{2019}\natexlab{b}.
\newblock \showarticletitle{Lightweight Face Recognition Challenge}. In \bibinfo{booktitle}{\emph{2019 IEEE/CVF International Conference on Computer Vision Workshop (ICCVW)}}. \bibinfo{pages}{2638--2646}.
\newblock
\urldef\tempurl%
\url{https://doi.org/10.1109/ICCVW.2019.00322}
\showDOI{\tempurl}


\bibitem[Desimone et~al\mbox{.}(1995)]%
        {desimone1995neural}
\bibfield{author}{\bibinfo{person}{Robert Desimone}, \bibinfo{person}{John Duncan}, {et~al\mbox{.}}} \bibinfo{year}{1995}\natexlab{}.
\newblock \showarticletitle{Neural mechanisms of selective visual attention}.
\newblock \bibinfo{journal}{\emph{Annual review of neuroscience}} \bibinfo{volume}{18}, \bibinfo{number}{1} (\bibinfo{year}{1995}), \bibinfo{pages}{193--222}.
\newblock


\bibitem[Dong et~al\mbox{.}(2019)]%
        {DongEtAl2019}
\bibfield{author}{\bibinfo{person}{Yinpeng Dong}, \bibinfo{person}{Hang Su}, \bibinfo{person}{Baoyuan Wu}, \bibinfo{person}{Zhifeng Li}, \bibinfo{person}{Wei Liu}, \bibinfo{person}{Tong Zhang}, {and} \bibinfo{person}{Jun Zhu}.} \bibinfo{year}{2019}\natexlab{}.
\newblock \showarticletitle{Efficient Decision-based Black-box Adversarial Attacks on Face Recognition}.
\newblock \bibinfo{journal}{\emph{CoRR}}  \bibinfo{volume}{abs/1904.04433} (\bibinfo{year}{2019}).
\newblock
\showeprint[arXiv]{1904.04433}
\urldef\tempurl%
\url{http://arxiv.org/abs/1904.04433}
\showURL{%
\tempurl}


\bibitem[Draganov et~al\mbox{.}(2024)]%
        {DraganovEtAl2024}
\bibfield{author}{\bibinfo{person}{Andrew Draganov}, \bibinfo{person}{Sharvaree Vadgama}, {and} \bibinfo{person}{Erik~J. Bekkers}.} \bibinfo{year}{2024}\natexlab{}.
\newblock \bibinfo{title}{The Hidden Pitfalls of the Cosine Similarity Loss}.
\newblock
\newblock
\showeprint[arxiv]{2406.16468}~[cs.LG]
\urldef\tempurl%
\url{https://arxiv.org/abs/2406.16468}
\showURL{%
\tempurl}


\bibitem[Elharrouss et~al\mbox{.}(2024)]%
        {Elharrouss_2024}
\bibfield{author}{\bibinfo{person}{Omar Elharrouss}, \bibinfo{person}{Younes Akbari}, \bibinfo{person}{Noor Almadeed}, {and} \bibinfo{person}{Somaya Al-Maadeed}.} \bibinfo{year}{2024}\natexlab{}.
\newblock \showarticletitle{Backbones-review: Feature extractor networks for deep learning and deep reinforcement learning approaches in computer vision}.
\newblock \bibinfo{journal}{\emph{Computer Science Review}}  \bibinfo{volume}{53} (\bibinfo{date}{Aug.} \bibinfo{year}{2024}), \bibinfo{pages}{100645}.
\newblock
\showISSN{1574-0137}
\urldef\tempurl%
\url{https://doi.org/10.1016/j.cosrev.2024.100645}
\showDOI{\tempurl}


\bibitem[FacePerceiver(2022)]%
        {facer}
\bibfield{author}{\bibinfo{person}{FacePerceiver}.} \bibinfo{year}{2022}\natexlab{}.
\newblock \bibinfo{title}{{GitHub - FacePerceiver/facer: Face analysis tools for modern research, equipped with state-of-the-art Face Parsing and Face Alignment}}.
\newblock
\newblock
\urldef\tempurl%
\url{https://github.com/FacePerceiver/facer}
\showURL{%
\tempurl}


\bibitem[Flamary et~al\mbox{.}(2021)]%
        {flamary2021pot}
\bibfield{author}{\bibinfo{person}{R{\'e}mi Flamary}, \bibinfo{person}{Nicolas Courty}, \bibinfo{person}{Alexandre Gramfort}, \bibinfo{person}{Mokhtar~Z. Alaya}, \bibinfo{person}{Aur{\'e}lie Boisbunon}, \bibinfo{person}{Stanislas Chambon}, \bibinfo{person}{Laetitia Chapel}, \bibinfo{person}{Adrien Corenflos}, \bibinfo{person}{Kilian Fatras}, \bibinfo{person}{Nemo Fournier}, \bibinfo{person}{L{\'e}o Gautheron}, \bibinfo{person}{Nathalie~T.H. Gayraud}, \bibinfo{person}{Hicham Janati}, \bibinfo{person}{Alain Rakotomamonjy}, \bibinfo{person}{Ievgen Redko}, \bibinfo{person}{Antoine Rolet}, \bibinfo{person}{Antony Schutz}, \bibinfo{person}{Vivien Seguy}, \bibinfo{person}{Danica~J. Sutherland}, \bibinfo{person}{Romain Tavenard}, \bibinfo{person}{Alexander Tong}, {and} \bibinfo{person}{Titouan Vayer}.} \bibinfo{year}{2021}\natexlab{}.
\newblock \showarticletitle{POT: Python Optimal Transport}.
\newblock \bibinfo{journal}{\emph{Journal of Machine Learning Research}} \bibinfo{volume}{22}, \bibinfo{number}{78} (\bibinfo{year}{2021}), \bibinfo{pages}{1--8}.
\newblock
\urldef\tempurl%
\url{http://jmlr.org/papers/v22/20-451.html}
\showURL{%
\tempurl}


\bibitem[Guetta et~al\mbox{.}(2021)]%
        {GuettaEtAl2021}
\bibfield{author}{\bibinfo{person}{Nitzan Guetta}, \bibinfo{person}{Asaf Shabtai}, \bibinfo{person}{Inderjeet Singh}, \bibinfo{person}{Satoru Momiyama}, {and} \bibinfo{person}{Yuval Elovici}.} \bibinfo{year}{2021}\natexlab{}.
\newblock \showarticletitle{Dodging Attack Using Carefully Crafted Natural Makeup}.
\newblock \bibinfo{journal}{\emph{CoRR}}  \bibinfo{volume}{abs/2109.06467} (\bibinfo{year}{2021}).
\newblock
\showeprint[arXiv]{2109.06467}
\urldef\tempurl%
\url{https://arxiv.org/abs/2109.06467}
\showURL{%
\tempurl}


\bibitem[InsightFace(2021)]%
        {insightface}
\bibfield{author}{\bibinfo{person}{InsightFace}.} \bibinfo{year}{2021}\natexlab{}.
\newblock \bibinfo{title}{InsightFace Model Zoo}.
\newblock
\newblock
\urldef\tempurl%
\url{https://github.com/deepinsight/insightface/tree/master}
\showURL{%
\tempurl}
\newblock
\shownote{Accessed: 2025-02-27}.


\bibitem[Jiang et~al\mbox{.}(2021)]%
        {JiangEtAL2021}
\bibfield{author}{\bibinfo{person}{Peng-Tao Jiang}, \bibinfo{person}{Chang-Bin Zhang}, \bibinfo{person}{Qibin Hou}, {and} \bibinfo{person}{Yunchao Wei}.} \bibinfo{year}{2021}\natexlab{}.
\newblock \showarticletitle{LayerCAM: Exploring Hierarchical Class Activation Maps}.
\newblock \bibinfo{journal}{\emph{IEEE Transactions on Image Processing}}  \bibinfo{volume}{PP} (\bibinfo{date}{06} \bibinfo{year}{2021}), \bibinfo{pages}{1--1}.
\newblock
\urldef\tempurl%
\url{https://doi.org/10.1109/TIP.2021.3089943}
\showDOI{\tempurl}


\bibitem[King(2015)]%
        {dlib}
\bibfield{author}{\bibinfo{person}{Davis~E. King}.} \bibinfo{year}{2015}\natexlab{}.
\newblock \bibinfo{title}{{GitHub - davisking/dlib: A toolkit for making real world machine learning and data analysis applications in C++}}.
\newblock \bibinfo{howpublished}{\url{https://github.com/davisking/dlib}}.
\newblock


\bibitem[Komkov and Petiushko(2019)]%
        {Komkov2019}
\bibfield{author}{\bibinfo{person}{Stepan~Alekseevich Komkov} {and} \bibinfo{person}{Aleksandr Petiushko}.} \bibinfo{year}{2019}\natexlab{}.
\newblock \showarticletitle{AdvHat: Real-World Adversarial Attack on ArcFace Face ID System}.
\newblock \bibinfo{journal}{\emph{2020 25th International Conference on Pattern Recognition (ICPR)}} (\bibinfo{year}{2019}), \bibinfo{pages}{819--826}.
\newblock
\urldef\tempurl%
\url{https://api.semanticscholar.org/CorpusID:201645162}
\showURL{%
\tempurl}


\bibitem[Panis et~al\mbox{.}(2015)]%
        {FG-NET2015}
\bibfield{author}{\bibinfo{person}{Gabriel Panis}, \bibinfo{person}{Andreas Lanitis}, \bibinfo{person}{Nicolas Tsapatsoulis}, {and} \bibinfo{person}{Timothy Cootes}.} \bibinfo{year}{2015}\natexlab{}.
\newblock \showarticletitle{An Overview of Research on Facial Aging using the FG-NET Aging Database}.
\newblock \bibinfo{journal}{\emph{IET Biometrics}}  \bibinfo{volume}{5} (\bibinfo{date}{05} \bibinfo{year}{2015}).
\newblock
\urldef\tempurl%
\url{https://doi.org/10.1049/iet-bmt.2014.0053}
\showDOI{\tempurl}


\bibitem[Provis et~al\mbox{.}(2013)]%
        {retina2013}
\bibfield{author}{\bibinfo{person}{Jan~M Provis}, \bibinfo{person}{Adam~M Dubis}, \bibinfo{person}{Ted Maddess}, {and} \bibinfo{person}{Joseph Carroll}.} \bibinfo{year}{2013}\natexlab{}.
\newblock \showarticletitle{Adaptation of the central retina for high acuity vision: cones, the fovea and the avascular zone}.
\newblock \bibinfo{journal}{\emph{Prog. Retin. Eye Res.}}  \bibinfo{volume}{35} (\bibinfo{date}{July} \bibinfo{year}{2013}), \bibinfo{pages}{63--81}.
\newblock


\bibitem[Ribeiro et~al\mbox{.}(2016)]%
        {RibeiroEtAl2016}
\bibfield{author}{\bibinfo{person}{Marco~Tulio Ribeiro}, \bibinfo{person}{Sameer Singh}, {and} \bibinfo{person}{Carlos Guestrin}.} \bibinfo{year}{2016}\natexlab{}.
\newblock \showarticletitle{"Why Should I Trust You?": Explaining the Predictions of Any Classifier}. In \bibinfo{booktitle}{\emph{Proceedings of the 22nd ACM SIGKDD International Conference on Knowledge Discovery and Data Mining}} (San Francisco, California, USA) \emph{(\bibinfo{series}{KDD '16})}. \bibinfo{publisher}{Association for Computing Machinery}, \bibinfo{address}{New York, NY, USA}, \bibinfo{pages}{1135–1144}.
\newblock
\showISBNx{9781450342322}
\urldef\tempurl%
\url{https://doi.org/10.1145/2939672.2939778}
\showDOI{\tempurl}


\bibitem[Rubner et~al\mbox{.}(2000)]%
        {rubner2000earth}
\bibfield{author}{\bibinfo{person}{Yossi Rubner}, \bibinfo{person}{Carlo Tomasi}, {and} \bibinfo{person}{Leonidas~J Guibas}.} \bibinfo{year}{2000}\natexlab{}.
\newblock \showarticletitle{The earth mover's distance as a metric for image retrieval}.
\newblock \bibinfo{journal}{\emph{International journal of computer vision}}  \bibinfo{volume}{40} (\bibinfo{year}{2000}), \bibinfo{pages}{99--121}.
\newblock


\bibitem[Schroff et~al\mbox{.}(2015)]%
        {SchroffEtAl15}
\bibfield{author}{\bibinfo{person}{Florian Schroff}, \bibinfo{person}{Dmitry Kalenichenko}, {and} \bibinfo{person}{James Philbin}.} \bibinfo{year}{2015}\natexlab{}.
\newblock \showarticletitle{FaceNet: {A} Unified Embedding for Face Recognition and Clustering}.
\newblock \bibinfo{journal}{\emph{CoRR}}  \bibinfo{volume}{abs/1503.03832} (\bibinfo{year}{2015}).
\newblock
\showeprint[arXiv]{1503.03832}
\urldef\tempurl%
\url{http://arxiv.org/abs/1503.03832}
\showURL{%
\tempurl}


\bibitem[Selvaraju et~al\mbox{.}(2017)]%
        {SelvarajuEtAl2016}
\bibfield{author}{\bibinfo{person}{Ramprasaath~R. Selvaraju}, \bibinfo{person}{Michael Cogswell}, \bibinfo{person}{Abhishek Das}, \bibinfo{person}{Ramakrishna Vedantam}, \bibinfo{person}{Devi Parikh}, {and} \bibinfo{person}{Dhruv Batra}.} \bibinfo{year}{2017}\natexlab{}.
\newblock \showarticletitle{Grad-CAM: Visual Explanations from Deep Networks via Gradient-Based Localization}. In \bibinfo{booktitle}{\emph{2017 IEEE International Conference on Computer Vision (ICCV)}}. \bibinfo{pages}{618--626}.
\newblock
\urldef\tempurl%
\url{https://doi.org/10.1109/ICCV.2017.74}
\showDOI{\tempurl}


\bibitem[Shan et~al\mbox{.}(2020)]%
        {shan2020fawkes}
\bibfield{author}{\bibinfo{person}{Shawn Shan}, \bibinfo{person}{Emily Wenger}, \bibinfo{person}{Jiayun Zhang}, \bibinfo{person}{Huiying Li}, \bibinfo{person}{Haitao Zheng}, {and} \bibinfo{person}{Ben~Y Zhao}.} \bibinfo{year}{2020}\natexlab{}.
\newblock \showarticletitle{Fawkes: Protecting privacy against unauthorized deep learning models}. In \bibinfo{booktitle}{\emph{29th USENIX security symposium (USENIX Security 20)}}. \bibinfo{pages}{1589--1604}.
\newblock


\bibitem[Sharif et~al\mbox{.}(2019)]%
        {sharif2019general}
\bibfield{author}{\bibinfo{person}{Mahmood Sharif}, \bibinfo{person}{Sruti Bhagavatula}, \bibinfo{person}{Lujo Bauer}, {and} \bibinfo{person}{Michael~K Reiter}.} \bibinfo{year}{2019}\natexlab{}.
\newblock \showarticletitle{A general framework for adversarial examples with objectives}.
\newblock \bibinfo{journal}{\emph{ACM Transactions on Privacy and Security (TOPS)}} \bibinfo{volume}{22}, \bibinfo{number}{3} (\bibinfo{year}{2019}), \bibinfo{pages}{1--30}.
\newblock


\bibitem[Shen et~al\mbox{.}(2019)]%
        {shen2019vla}
\bibfield{author}{\bibinfo{person}{Meng Shen}, \bibinfo{person}{Zelin Liao}, \bibinfo{person}{Liehuang Zhu}, \bibinfo{person}{Ke Xu}, {and} \bibinfo{person}{Xiaojiang Du}.} \bibinfo{year}{2019}\natexlab{}.
\newblock \showarticletitle{Vla: A practical visible light-based attack on face recognition systems in physical world}.
\newblock \bibinfo{journal}{\emph{Proceedings of the ACM on Interactive, Mobile, Wearable and Ubiquitous Technologies}} \bibinfo{volume}{3}, \bibinfo{number}{3} (\bibinfo{year}{2019}), \bibinfo{pages}{1--19}.
\newblock


\bibitem[Soydaner(2022)]%
        {soydaner2022attention}
\bibfield{author}{\bibinfo{person}{Derya Soydaner}.} \bibinfo{year}{2022}\natexlab{}.
\newblock \showarticletitle{Attention mechanism in neural networks: where it comes and where it goes}.
\newblock \bibinfo{journal}{\emph{Neural Computing and Applications}} \bibinfo{volume}{34}, \bibinfo{number}{16} (\bibinfo{year}{2022}), \bibinfo{pages}{13371--13385}.
\newblock


\bibitem[Steck et~al\mbox{.}(2024)]%
        {SteckEtAl2024}
\bibfield{author}{\bibinfo{person}{Harald Steck}, \bibinfo{person}{Chaitanya Ekanadham}, {and} \bibinfo{person}{Nathan Kallus}.} \bibinfo{year}{2024}\natexlab{}.
\newblock \showarticletitle{Is Cosine-Similarity of Embeddings Really About Similarity?}. In \bibinfo{booktitle}{\emph{Companion Proceedings of the ACM Web Conference 2024}} \emph{(\bibinfo{series}{WWW ’24})}. \bibinfo{publisher}{ACM}, \bibinfo{pages}{887–890}.
\newblock
\urldef\tempurl%
\url{https://doi.org/10.1145/3589335.3651526}
\showDOI{\tempurl}


\bibitem[Szegedy et~al\mbox{.}(2015)]%
        {SzegedyEtAl2015}
\bibfield{author}{\bibinfo{person}{Christian Szegedy}, \bibinfo{person}{Wei Liu}, \bibinfo{person}{Yangqing Jia}, \bibinfo{person}{Pierre Sermanet}, \bibinfo{person}{Scott Reed}, \bibinfo{person}{Dragomir Anguelov}, \bibinfo{person}{Dumitru Erhan}, \bibinfo{person}{Vincent Vanhoucke}, {and} \bibinfo{person}{Andrew Rabinovich}.} \bibinfo{year}{2015}\natexlab{}.
\newblock \showarticletitle{Going deeper with convolutions}. In \bibinfo{booktitle}{\emph{2015 IEEE Conference on Computer Vision and Pattern Recognition (CVPR)}}. \bibinfo{pages}{1--9}.
\newblock
\urldef\tempurl%
\url{https://doi.org/10.1109/CVPR.2015.7298594}
\showDOI{\tempurl}


\bibitem[Timesler(2018)]%
        {facenetpytorch}
\bibfield{author}{\bibinfo{person}{Timesler}.} \bibinfo{year}{2018}\natexlab{}.
\newblock \bibinfo{title}{facenet-pytorch}.
\newblock
\newblock
\urldef\tempurl%
\url{https://github.com/timesler/facenet-pytorch}
\showURL{%
\tempurl}
\newblock
\shownote{Accessed: 2025-01-30}.


\bibitem[Vakhshiteh et~al\mbox{.}(2021)]%
        {VakhshitehEtAl2021}
\bibfield{author}{\bibinfo{person}{Fatemeh Vakhshiteh}, \bibinfo{person}{Ahmad Nickabadi}, {and} \bibinfo{person}{Raghavendra Ramachandra}.} \bibinfo{year}{2021}\natexlab{}.
\newblock \showarticletitle{Adversarial Attacks Against Face Recognition: A Comprehensive Study}.
\newblock \bibinfo{journal}{\emph{IEEE Access}}  \bibinfo{volume}{9} (\bibinfo{year}{2021}), \bibinfo{pages}{92735--92756}.
\newblock
\urldef\tempurl%
\url{https://doi.org/10.1109/ACCESS.2021.3092646}
\showDOI{\tempurl}


\bibitem[Wang et~al\mbox{.}(2018a)]%
        {wang2018}
\bibfield{author}{\bibinfo{person}{Fei Wang}, \bibinfo{person}{Liren Chen}, \bibinfo{person}{Cheng Li}, \bibinfo{person}{Shiyao Huang}, \bibinfo{person}{Yanjie Chen}, \bibinfo{person}{Chen Qian}, {and} \bibinfo{person}{Chen~Change Loy}.} \bibinfo{year}{2018}\natexlab{a}.
\newblock \bibinfo{title}{The Devil of Face Recognition is in the Noise}.
\newblock
\newblock
\showeprint[arxiv]{1807.11649}~[cs.CV]
\urldef\tempurl%
\url{https://arxiv.org/abs/1807.11649}
\showURL{%
\tempurl}


\bibitem[Wang et~al\mbox{.}(2018b)]%
        {wang2018cosfacelargemargincosine}
\bibfield{author}{\bibinfo{person}{Hao Wang}, \bibinfo{person}{Yitong Wang}, \bibinfo{person}{Zheng Zhou}, \bibinfo{person}{Xing Ji}, \bibinfo{person}{Dihong Gong}, \bibinfo{person}{Jingchao Zhou}, \bibinfo{person}{Zhifeng Li}, {and} \bibinfo{person}{Wei Liu}.} \bibinfo{year}{2018}\natexlab{b}.
\newblock \bibinfo{title}{CosFace: Large Margin Cosine Loss for Deep Face Recognition}.
\newblock
\newblock
\showeprint[arxiv]{1801.09414}~[cs.CV]
\urldef\tempurl%
\url{https://arxiv.org/abs/1801.09414}
\showURL{%
\tempurl}


\bibitem[Xia et~al\mbox{.}(2015)]%
        {XiaEtAl2015}
\bibfield{author}{\bibinfo{person}{Peipei Xia}, \bibinfo{person}{Li Zhang}, {and} \bibinfo{person}{Fanzhang Li}.} \bibinfo{year}{2015}\natexlab{}.
\newblock \showarticletitle{Learning similarity with cosine similarity ensemble}.
\newblock \bibinfo{journal}{\emph{Information Sciences}}  \bibinfo{volume}{307} (\bibinfo{year}{2015}), \bibinfo{pages}{39--52}.
\newblock
\showISSN{0020-0255}
\urldef\tempurl%
\url{https://doi.org/10.1016/j.ins.2015.02.024}
\showDOI{\tempurl}


\bibitem[Yang et~al\mbox{.}(2022)]%
        {YangEtAl2022}
\bibfield{author}{\bibinfo{person}{Xiao Yang}, \bibinfo{person}{Yinpeng Dong}, \bibinfo{person}{Tianyu Pang}, \bibinfo{person}{Zihao Xiao}, \bibinfo{person}{Hang Su}, {and} \bibinfo{person}{Jun Zhu}.} \bibinfo{year}{2022}\natexlab{}.
\newblock \bibinfo{title}{Controllable Evaluation and Generation of Physical Adversarial Patch on Face Recognition}.
\newblock
\newblock
\showeprint[arxiv]{2203.04623}~[cs.CV]
\urldef\tempurl%
\url{https://arxiv.org/abs/2203.04623}
\showURL{%
\tempurl}


\bibitem[Yang et~al\mbox{.}(2023)]%
        {YangEtAl2023}
\bibfield{author}{\bibinfo{person}{Xiao Yang}, \bibinfo{person}{Chang Liu}, \bibinfo{person}{Longlong Xu}, \bibinfo{person}{Yikai Wang}, \bibinfo{person}{Yinpeng Dong}, \bibinfo{person}{Ning Chen}, \bibinfo{person}{Hang Su}, {and} \bibinfo{person}{Jun Zhu}.} \bibinfo{year}{2023}\natexlab{}.
\newblock \showarticletitle{{ Towards Effective Adversarial Textured 3D Meshes on Physical Face Recognition }}. In \bibinfo{booktitle}{\emph{2023 IEEE/CVF Conference on Computer Vision and Pattern Recognition (CVPR)}}. \bibinfo{publisher}{IEEE Computer Society}, \bibinfo{address}{Los Alamitos, CA, USA}, \bibinfo{pages}{4119--4128}.
\newblock
\urldef\tempurl%
\url{https://doi.org/10.1109/CVPR52729.2023.00401}
\showDOI{\tempurl}


\bibitem[Yi et~al\mbox{.}(2014)]%
        {YiEtAl2014}
\bibfield{author}{\bibinfo{person}{Dong Yi}, \bibinfo{person}{Zhen Lei}, \bibinfo{person}{Shengcai Liao}, {and} \bibinfo{person}{Stan~Z. Li}.} \bibinfo{year}{2014}\natexlab{}.
\newblock \bibinfo{title}{Learning Face Representation from Scratch}.
\newblock
\newblock
\showeprint[arxiv]{1411.7923}~[cs.CV]
\urldef\tempurl%
\url{https://arxiv.org/abs/1411.7923}
\showURL{%
\tempurl}


\bibitem[Zhao et~al\mbox{.}(2003)]%
        {ZhaoEtAl2003}
\bibfield{author}{\bibinfo{person}{W. Zhao}, \bibinfo{person}{R. Chellappa}, \bibinfo{person}{P.~J. Phillips}, {and} \bibinfo{person}{A. Rosenfeld}.} \bibinfo{year}{2003}\natexlab{}.
\newblock \showarticletitle{Face recognition: A literature survey}.
\newblock \bibinfo{journal}{\emph{ACM Comput. Surv.}} \bibinfo{volume}{35}, \bibinfo{number}{4} (\bibinfo{year}{2003}), \bibinfo{pages}{399–458}.
\newblock
\showISSN{0360-0300}
\urldef\tempurl%
\url{https://doi.org/10.1145/954339.954342}
\showDOI{\tempurl}


\bibitem[Zheng et~al\mbox{.}(2022)]%
        {ZhengEtAl2022}
\bibfield{author}{\bibinfo{person}{Yinglin Zheng}, \bibinfo{person}{Hao Yang}, \bibinfo{person}{Ting Zhang}, \bibinfo{person}{Jianmin Bao}, \bibinfo{person}{Dongdong Chen}, \bibinfo{person}{Yangyu Huang}, \bibinfo{person}{Lu Yuan}, \bibinfo{person}{Dong Chen}, \bibinfo{person}{Ming Zeng}, {and} \bibinfo{person}{Fang Wen}.} \bibinfo{year}{2022}\natexlab{}.
\newblock \bibinfo{title}{General Facial Representation Learning in a Visual-Linguistic Manner}.
\newblock
\newblock
\showeprint[arxiv]{2112.03109}~[cs.CV]
\urldef\tempurl%
\url{https://arxiv.org/abs/2112.03109}
\showURL{%
\tempurl}


\bibitem[Zhou et~al\mbox{.}(2015)]%
        {ZhouEtAl2015}
\bibfield{author}{\bibinfo{person}{Bolei Zhou}, \bibinfo{person}{Aditya Khosla}, \bibinfo{person}{{\`{A}}gata Lapedriza}, \bibinfo{person}{Aude Oliva}, {and} \bibinfo{person}{Antonio Torralba}.} \bibinfo{year}{2015}\natexlab{}.
\newblock \showarticletitle{Learning Deep Features for Discriminative Localization}.
\newblock \bibinfo{journal}{\emph{CoRR}}  \bibinfo{volume}{abs/1512.04150} (\bibinfo{year}{2015}).
\newblock
\showeprint[arXiv]{1512.04150}
\urldef\tempurl%
\url{http://arxiv.org/abs/1512.04150}
\showURL{%
\tempurl}


\end{thebibliography}
